\def\year{2022}\relax
\documentclass[letterpaper]{article} 
\usepackage{aaai22}  
\usepackage{times}  
\usepackage{helvet}  
\usepackage{courier}  
\usepackage[hyphens]{url}  
\usepackage{graphicx} 
\def\UrlFont{\rm}  
\usepackage{natbib}  
\usepackage{caption} 
\DeclareCaptionStyle{ruled}{labelfont=normalfont,labelsep=colon,strut=off} 
\usepackage{algorithm}
\usepackage{algorithmic}
\usepackage{subcaption}

\usepackage{newfloat}
\usepackage{listings}
\floatstyle{ruled}
\newfloat{listing}{tb}{lst}{}
\floatname{listing}{Listing}
\usepackage{ulem}
\newcommand\redsout{\bgroup\markoverwith{\textcolor{red}{\rule[0.5ex]{2pt}{0.4pt}}}\ULon} 

\title{Fast Sparse Decision Tree Optimization via Reference Ensembles}

\author{
    Hayden McTavish\textsuperscript{\rm 1,\rm 3}\equalcontrib,
    Chudi Zhong\textsuperscript{\rm 2}\equalcontrib,
    Reto Achermann\textsuperscript{\rm 1}, 
    Ilias Karimalis\textsuperscript{\rm 1}, 
    Jacques Chen\textsuperscript{\rm 1}, \\
    Cynthia Rudin\textsuperscript{\rm 2}, 
    Margo Seltzer\textsuperscript{\rm 1}
}
\affiliations{

    \textsuperscript{\rm 1} University of British Columbia\\
    \textsuperscript{\rm 2} Duke University\\
    \textsuperscript{\rm 3} University of California, San Diego\\
    hmctavish@ucsd.edu, \{chudi.zhong, cynthia.rudin\}@duke.edu, \{achreto, mseltzer\}@cs.ubc.ca, iliaskar@students.cs.ubc.ca, jacquesc@student.ubc.ca

}

\usepackage[utf8]{inputenc} 
\usepackage[T1]{fontenc}    
\usepackage{booktabs}       
\usepackage{amsfonts}       
\usepackage{nicefrac}       
\usepackage{microtype}      
\usepackage{xcolor}         
\usepackage{graphicx}
\usepackage{enumitem}
\usepackage{amsmath} 
\usepackage{natbib}
\usepackage{amsthm}
\newtheorem{theorem}{Theorem}[section]

\newtheorem{corollary}{Corollary}[theorem]
\usepackage{bm}
\usepackage{algorithm}
\usepackage{algorithmic}
\usepackage{tabularx}
\usepackage{multirow}
\usepackage{dsfont}
\usepackage[linguistics]{forest}

\usepackage{pgfplots}
\pgfplotsset{compat=newest}

\def\x{\mathbf{x}}
\def\y{\mathbf{y}}

\def\CC{s_{T,\textrm{\rm correct}}}
\def\MC{s_{T,\textrm{\rm incorrect}}}
\def\CCp{s_{T',\textrm{\rm correct}}}
\def\MCp{s_{T',\textrm{\rm incorrect}}}

\def\BestTree{\textit{BestTree}}
\def\initub{\textit{init\_ub}}

\begin{document}

\maketitle
\begin{abstract}
Sparse decision tree optimization has been one of the most fundamental problems in AI since its inception and is a challenge at the core of interpretable machine learning. Sparse decision tree optimization is computationally hard, and despite steady effort since the 1960's, breakthroughs have been made on the problem only within the past few years, primarily on the problem of finding optimal sparse decision trees. However, current state-of-the-art algorithms often require impractical amounts of computation time and memory to find optimal or near-optimal trees for some real-world datasets, particularly those having several continuous-valued features. Given that the search spaces of these decision tree optimization problems are massive, can we practically hope to find a sparse decision tree that competes in accuracy with a black box machine learning model? We address this problem via smart guessing strategies that can be applied to any optimal branch-and-bound-based decision tree algorithm. The guesses come from knowledge gleaned from black box models. We show that by using these guesses, we can reduce the run time by multiple orders of magnitude while providing bounds on how far the resulting trees can deviate from the black box's accuracy and expressive power. Our approach enables guesses about how to bin continuous features, the size of the tree, and lower bounds on the error for the optimal decision tree. Our experiments show that in many cases we can rapidly construct sparse decision trees that match the accuracy of black box models. To summarize:
\textit{when you are having trouble optimizing, just guess}.
\end{abstract}

\section{Introduction}






Decision trees are one of the leading forms of interpretable AI models. Since the development of the first decision tree algorithm \citep{morgan1963problems}, a huge number of algorithms have been proposed to improve both accuracy and run time. However, major approaches are based on decision tree induction, using heuristic splitting and pruning \citep[e.g.,][]{breiman1984classification, Quinlan93}. Growing a tree in a greedy way, though fast, leads to suboptimal models with no indication of how far away the solution is from optimality. The generated trees are usually much more complicated than they need to be, hindering interpretability. Optimizing sparse decision trees remains one of the most fundamental problems in machine learning (ML).

Full decision tree optimization is NP-hard \citep{laurent1976constructing}, leading to challenges in searching for optimal trees in a reasonable time, even for small datasets. Major advances have been made recently using either mathematical programming solvers \citep[e.g.,][]{verwer2019learning} or customized branch-and-bound with dynamic programming \citep{aglin2020learning, lin2020generalized, demirovic2020murtree}, showing us that there is hope. 
However, these methods are frequently unable to find the optimal tree within a reasonable amount of time, or even if they do find the optimal solution, it can take a long time to \textit{prove} that the tree is optimal or close-to-optimal.


Ideally, we would like an algorithm that, within a few minutes, produces a sparse decision tree that is as accurate as a black box machine learning model. Also, we wish to have a guarantee that the model will have performance close to that of the black box. We present a practical way to achieve this 
by introducing a set of smart guessing techniques that speed up sparse decision tree computations for branch-and-bound methods by orders of magnitude. 

The key is to guess in a way that prunes the search space without eliminating optimal and near-optimal solutions. We derive those smart guesses from a black box tree-ensemble reference model whose performance we aim to match. 
Our guesses come in three flavors. The first type of guess reduces the number of thresholds we consider as a possible split on a continuous feature. Here, we use splits generated by black box ensembles. The second type of guess concerns the maximum depth we might need for an optimal tree, where we relate the complexity of a tree ensemble to the depth of an equivalently complex class of individual trees. The third type of guess uses the accuracy of black box models on subsets of the data to guess lower bounds on the loss for each subproblem we encounter. Our guesses are guaranteed to predict as well or better than the black box tree-ensemble reference model: taking the sparsest decision tree that makes the same predictions as the black box, our method will find this tree, an equivalently good tree, or an even better one. Together, these guesses decrease the run time by several orders of magnitude, allowing fast production of sparse and interpretable trees that achieve black box predictive performance. 

\section{Related Work}
Our work relates to the field of decision tree optimization and thinning tree ensembles. 

\textbf{Decision tree optimization.} Optimization techniques have been used for decision trees from the 1990s until the present \citep{bennett1996optimal, dobkin1997induction, farhangfar2008fast, nijssen2007mining, nijssen2010optimal}. Recently, many works have directly optimized a performance metric (e.g., accuracy) with soft or hard sparsity constraints on the tree size. Such decision tree optimization problems can be formulated using mixed integer programming (MIP) \citep{bertsimas2017optimal, verwer2019learning,vilas2021optimal,gunluk2021optimal,ErtekinRu18,aghaei2020learning}. 
Other approaches use SAT solvers to find optimal decision trees \citep{narodytska2018learning, hu2020learning}, though these techniques require data to be perfectly separable, which is not typical for machine learning. \citet{carrizosa2021mathematical} provide an overview of mathematical programming for decision trees. 

Another branch of decision tree optimization has produced customized dynamic programming algorithms that incorporate branch-and-bound techniques. \citet{hu2019optimal,AngelinoEtAl2018,ChenRu2018} use analytical bounds combined with bit-vector based computation to efficiently reduce the search space and construct optimal sparse decision trees. \citet{lin2020generalized} extend this work to use dynamic programming. \citet{aglin2020learning} also use dynamic programming with bounds to find optimal trees of a given depth. \citet{demirovic2020murtree} additionally introduce constraints on both depth and number of nodes to improve the scalability of decision tree optimization. 

Our guessing strategies can further improve the scalability of all branch-and-bound based optimal decision tree algorithms without fundamentally changing their respective search strategies. 
Instead, we reduce time and memory costs by using a subset of thresholds and tighter lower bounds to prune the search space.


\textbf{``Born-again'' trees.} \citet{breiman1996born} proposed to replace a tree ensemble with a newly constructed single tree. The tree ensemble is used to generate additional observations that are then used to find the best split for a tree node in the new tree. 
\citet{zhou2016interpreting} later follow a similar strategy.   
Other recent work uses black box models to determine splitting and stopping criteria for growing a single tree inductively \citep{bai2019rectified} or exploit the class distributions predicted by an ensemble to determine splitting and stopping criteria \citep{van2007seeing}. 
\citet{vandewiele2016genesim} use a genetic approach to construct a large ensemble and combine models from different subsets of this ensemble to get a single model with high accuracy.

Another branch of work focuses on the inner structure of the tree ensemble. \citet{akiba1998turning} generate if-then rules from each of the ensemble classifiers and convert the rules into binary vectors. These vectors are then used as training data for learning a new decision tree. Recent work following this line of reasoning extracts, ranks, and prunes conjunctions from the tree ensemble and organizes the set of conjunctions into a rule set \citep{sirikulviriya2011integration}, an ordered rule list \citep{deng2019interpreting}, or a single tree \citep{sagi2020explainable}. \citet{hara2018making} adopt a probabilistic model representation of a tree ensemble and use Bayesian model selection for tree pruning.

The work of \citet{vidal2020born} produces a decision tree of minimal complexity that makes identical predictions to the reference model (a tree ensemble). Such a decision tree is called a born-again tree ensemble.


We use a reference model for two purposes: to help reduce the number of possible values one could split on and to determine an accuracy goal for a subset of points. 
By using a tree ensemble as our reference model, we can guarantee that our solution will have a regularized objective that matches or improves upon that of any born-again tree ensemble. 
%
%
%

\section{Notation and Objectives} \label{sec:notation}

We denote the training dataset as $\{(\x_i, y_i)\}_{i=1}^N$, where $\x_i$ are $M$-vectors of features, and $y_i \in \{0,1\}$ are labels. Let $\x$ be the $N\times M$ covariate matrix and $\y$ be the $N$-vector of labels, and let $x_{ij}$ denote the $j$-th feature of $\x_i$. We transform each continuous feature into binary features by creating a split point at the mean value between every ordered pair of unique values present in the training data. Let $k_j$ be the number of unique values realized by feature $j$, then the total number of features is $\tilde{M} = \sum_{j=1}^M (k_j-1)$. We denote the binarized covariate matrix as $\tilde{\x}$ where $\tilde{\x}_i \in \{0,1\}^{\tilde{M}}$ are binary features. 

Let $\mathcal{L}(t,\tilde{\x},\y) = \frac{1}{N}\sum_{i=1}^N \mathbf{1}[y_i \neq \hat{y}_i^t]$ be the loss of the tree $t$ on the training dataset, given predictions $\{\hat{y}_i^t\}_{i=1}^N$ from tree $t$. Most optimal decision tree algorithms minimize a misclassification loss constrained by a depth bound, i.e.,
\begin{equation}\label{eq:obj1}
    \underset{t}{\text{minimize}} \quad \mathcal{L}(t,\tilde{\x},\y), \text{ s.t. } \textrm{depth}(t) \leq d.
\end{equation}
Instead, \citet{hu2019optimal} and \citet{lin2020generalized} define the objective function $R(t,\tilde{\x},\y)$ as the combination of the misclassification loss and a sparsity penalty on the number of leaves. That is, $R(t,\tilde{\x},\y) = \mathcal{L}(t,\tilde{\x},\y) + \lambda H_t$, where $H_t$ is the number of leaves in the tree $t$ and $\lambda$ is a regularization parameter. They minimize the objective function, i.e., 
\begin{equation}\label{eq:obj2}
    \underset{t}{\text{minimize}} \quad \mathcal{L}(t,\tilde{\x},\y) + \lambda H_t.
\end{equation}
Ideally, we prefer to solve \eqref{eq:obj2} since we do not know the optimal depth in advance, and even at a given depth, we would prefer to minimize the number of leaves. But \eqref{eq:obj1} is much easier to solve. We discuss details in Section \ref{sec:depth_guess}.




\section{Methodology}
We present three guessing techniques. 
The first guesses how to transform continuous features into binary features, the second guesses tree depth for sparsity-regularized models, and the third guesses tighter bounds to allow faster time-to-completion. 
We use a boosted decision tree \citep{freund95, friedman2001greedy} as our reference model whose performance we want to match or exceed. We refer to Appendix \ref{app:proofs} for proofs of all theorems presented. 

\subsection{Guessing Thresholds via Column Elimination}\label{sec:thresh_guess}

Since most state-of-the-art optimal decision tree algorithms require binary inputs, continuous features require preprocessing. We can transform a continuous feature into a set of binary dummy variables.
Call each midpoint between two consecutive points a \textit{split point}.
Unfortunately, na\"ively creating split points at the mean values between each ordered pair of unique values present in the training data can dramatically increase the search space of decision trees \cite[][Theorem H.1]{lin2020generalized}, leading to the possibility of encountering either a time or memory limit. \citet{verwer2019learning} use a technique called bucketization to reduce the number of thresholds considered: instead of including split points between all realized values of each continuous feature, bucketization removes split points between realized feature values for which the labels are identical. This reduces computation, but for our purposes, it is still too conservative, and will lead to slow run times. Instead, we use a subset of thresholds from our reference boosted decision tree model, since we know that with these thresholds, it is possible to produce a model with the accuracy we are trying to match. Because computation of sparse trees has factorial time complexity, each feature we remove reduces run time substantially. 
%

\textbf{Column Elimination Algorithm}: 
Our algorithm works by iteratively pruning features (removing columns) of least importance until a predefined threshold is reached:
1) Starting with our reference model,
extract all thresholds for all features used in all of the trees in the boosting model. 2) Order them by variable importance (we use Gini importance), and remove the least important threshold (among all thresholds and all features). 3) Re-fit the boosted tree with the remaining features. 4) Continue this procedure until the training performance of the remaining tree drops below a predefined threshold. (In our experiments, we stop the procedure when there is any drop in performance at all, but one could use a different threshold such as 1\% if desired.)
After this procedure, the remaining data matrix is denoted by $\tilde{\x}_T^{\textrm{reduced}}$.

If, during this procedure, we eliminate too many thresholds, we will not be able to match black box accuracy. Theorem \ref{thm:threshold_guess} bounds the gap between the loss of the black-box tree ensemble and loss of the optimal tree after threshold guessing. 

Let $T'$ be the ensemble tree built with $(\tilde{\x}^{\textrm{reduced}}_T, \y)$. 
We consider $t'$ to be any decision tree that makes the same predictions as $T'$ for every observation in $\tilde{\x}_T^{\textrm{reduced}}$. 
We provide Figure \ref{fig:merge} in the appendix as an illustration of finding a $t'$ with the minimum number of leaves for a given ensemble.
Note that $H_{t'}$ may be relatively small even if $T'$ is a fairly large ensemble, because the predictions are binary (each leaf predicts either 0 or 1) and the outcomes tend to vary smoothly as continuous features change, with not many jumps. 

\begin{theorem}\label{thm:threshold_guess} (Guarantee for model on reduced data).
Define the following:
\begin{itemize}
\item Let $T$ be the ensemble tree built upon $(\x, \y)$. 
\item Let $T'$ be the ensemble tree built with $(\tilde{\x}_T^{\textrm{reduced}}, \y)$. 
\item Let $t'$ be any decision tree that makes the same predictions as $T'$ for every observation in $\tilde{\x}_T^{\textrm{reduced}}$. 
\item Let $t^* \in \arg\min_t \mathcal{L}(t, \tilde{\x}_T^{\textrm{reduced}}, \y) + \lambda H_t$ be an optimal tree on the reduced dataset. 
\end{itemize}
Then, $\mathcal{L}(t^*, \tilde{\x}_T^{\textrm{reduced}}, \y) - \mathcal{L}(T, \x, \y) \leq \lambda (H_{t'}- H_{t^*})$, or equivalently, 
$R(t^*, \tilde{\x}_T^{\textrm{reduced}}, \y) \leq \mathcal{L}(T, \x, \y) + \lambda H_{t'}$.
\end{theorem}

That is, for any tree $t'$ that matches the predictions of $T'$ for $\tilde{\x}_T^\textrm{reduced}$ (even the smallest such tree), the difference in loss between the black box tree ensemble $T$ and the optimal single tree $t^*$ (based on $\tilde{\x}_T^{\textrm{reduced}}$) is not worse than the regularization coefficient times the difference between the sizes of two trees: $t^*$ and $t'$.
Equivalently, $t^*$ will never be worse than $t'$ in terms of the regularized objective. 
If we pick $t'$ to be a born-again tree ensemble for $T'$ \citep{vidal2020born} (which we can do because born-again tree ensembles make the same predictions as $T'$ for all inputs, and therefore make the same predictions for $\tilde{\x}_T^{\textrm{reduced}}$), we can show that our thresholding approach guarantees that we will never do worse than the best born-again tree ensemble for our simplified reference model, in terms of the regularized objective. If we pick a $t'$ with the minimal number of leaves, this theorem shows we even match or beat the simplest tree that can exactly replicate the reference models' predictions on the training set.

\subsection{Guessing Depth} \label{sec:depth_guess}

As discussed in Section~\ref{sec:notation}, there are two different approaches to producing optimal decision trees: one uses a hard depth constraint and the other uses a per-leaf penalty.
Algorithms that use only depth constraints tend to run more quickly but can produce needlessly complicated trees, because they do not reward trees for being shallower than the maximum depth. Depth constraints assess trees only by the length of their longest path, not by any other measure of simplicity, such as the length of the average path or the number of different decision paths. In contrast, per-leaf penalty algorithms \citep{hu2019optimal,lin2020generalized} produce sparse models, but frequently have longer running times, as they search a much larger space, because they do not assume they know the depth of an optimal tree. 
We show that by adding a depth constraint (a guess on the maximum depth of an optimal tree) to per-leaf penalty algorithms, such as GOSDT~\citep{lin2020generalized}, we can achieve \textit{per-leaf} sparsity regularization at run times comparable to \textit{depth-constrained} algorithms, such as DL8.5~\citep{aglin2020learning}. In particular, we combine~\eqref{eq:obj1} and~\eqref{eq:obj2} to produce a new objective,
\begin{equation}\label{eq:obj5}
    \underset{t}{\text{minimize}} \quad \mathcal{L}(t,\tilde{\x},\y) + \lambda H_t  \text{ s.t. } \textrm{depth}(t) \leq d,
\end{equation}
where we aim to choose $d$ such that it reduces the search space without removing all optimal or near-optimal solutions. 
Most papers use depth guesses between 2 and 5 \citep{aglin2020learning}, which we have done in our experiments. However, Theorem \ref{thm:vcdepth} provides guidance on other ways to select a depth constraint to train accurate models quickly. 
Also, Theorem \ref{thm:depth_guess} bounds the gap between the objectives of optimal trees with a relatively smaller depth guess and with no depth guess. This gap depends on the trade-off between sparsity and accuracy. 

Interestingly, using a large depth constraint is often less efficient than removing the depth constraint entirely for GOSDT, because when we use a branch-and-bound approach with recursion, the ability to re-use solutions of recurring subproblems diminishes in the presence of a depth constraint. 

\begin{theorem}\label{thm:vcdepth} (Min depth needed to match complexity of ensemble).
Let $B$ be the base hypothesis class (e.g., decision stumps or shallow trees) that has VC dimension at least 3 and let $K \geq 3$ be the number of weak classifiers (members of $B$) combined in an ensemble model. Let $\mathcal{F}_{\textrm{ensemble}}$ be the set of weighted sums of weak classifiers, i.e., $T\in \mathcal{F}_{\textrm{ensemble}}$ has $T(x)=\textrm{sign}(\sum_{k=1}^K w_k h_k(x))$,  where $\forall k, w_k \in \mathbb{R}, h_k \in B$. Let $\mathcal{F}_{d,\textrm{tree}}$ be the class of single binary decision trees with depth at most: \[d = \left\lceil \log_2 \left(\left(K \!\cdot\! \textrm{VC}(B)\!+\!K\right)\cdot\left(3\ln(K \!\cdot\! \textrm{VC}(B)\!+\!K\right)\!+\!2)\right)\right\rceil.\] It is then true that $\textrm{VC}(\mathcal{F}_{d,\textrm{tree}}) \geq \textrm{VC}(\mathcal{F}_{\textrm{ensemble}})$. 
\end{theorem}

That is, the class of single trees with depth at most $d$ has complexity at least that of the ensemble.
The following results are special cases of Theorem \ref{thm:vcdepth}: 
\begin{itemize}[leftmargin=*]
    \item Suppose $B$ is the class of decision trees with depth at most 3. To match or exceed the complexity of an ensemble of $K=10$ trees from $B$ with a single tree, it is sufficient to use trees of depth 11.
    \item Suppose $B$ is the class of decision trees with depth at most 3. To match or exceed the complexity of an ensemble of $K=100$ trees from $B$ with a single tree, it is sufficient to use trees of depth 15. 
\end{itemize}
The bound is conservative, so we might choose a smaller depth than is calculated in the theorem; the theorem provides an upper bound on the depth we need to consider for matching the accuracy of the black box.  

\subsection{Guessing Tighter Lower Bounds} \label{sec:lbguess}


Branch-and-bound approaches to decision tree optimization, such as GOSDT and DL8.5, are limited by the inefficiency of their lower bound estimates. To remove a potential feature split from the search, the algorithm must prove that the best possible (lowest) objectives on the two resulting subproblems sum to a value that is worse (larger) than optimal. Calculating tight enough bounds to do this is often slow, requiring an algorithm to waste substantial time exploring suboptimal parts of the search space before it can prove the absence of an optimal solution in that space.

We use informed guesses to quickly tighten lower bounds. These guesses are based on a reference model -- another classifier 
that we believe will misclassify a similar set of training points to an optimal tree. Let $T$ be such a reference model and $\hat{y}_i^T$ be the predictions of that reference model on training observation $i$. 
Define $s_a$ as the subset of training observations that satisfy a boolean assertion $a$: 
\begin{align*}
    s_{a} &:= \left\{i:a(\tilde{\x}_i) = \textrm{True}, i \in \{1,...,N\}  \right\} \\
    \tilde{\x}({s_{a}}) &:= \left\{\tilde{\x}_i: i \in s_{a} \right\}\\
    \y({s_{a}}) &:= \left\{y_i: i \in s_{a} \right\}.
\end{align*}
We can then define our guessed lower bound as the disagreement of the reference and true labels for these observations (plus a penalty reflecting that at least 1 leaf will be used in solving this subproblem): 
\begin{equation}lb_\textrm{guess}(s_a):=\frac{1}{N}\sum_{i \in s_a} \mathbf{1}[y_i \neq \hat{y}_i^{T}] + \lambda.\end{equation}
We use this lower bound to prune the search space. In particular, we consider a subproblem to be solved if we find a subtree that achieves an objective less than or equal to its $lb_\textrm{guess}$ (even if we have not proved it is optimal); this is equivalent to assuming the subproblem is solved when we match the reference model's accuracy on that subproblem. We further let the algorithm omit any part of the space whose estimated lower bound is large enough to suggest that it does not contain an optimal solution. That is, for each subproblem, we use the reference model's accuracy as a guide for the best accuracy we hope to obtain.

Thus, we introduce modifications to a general branch-and-bound decision tree algorithm that allow it to use lower bound guessing informed by a reference model. We focus here on approaches to solve \eqref{eq:obj5}, noting that \eqref{eq:obj1} and \eqref{eq:obj2} 
are special cases of  \eqref{eq:obj5}. 
For a subset of observations $s_a$, let $t_{a}$ be the subtree used to classify those points, and let $H_{t_{a}}$ be the number of leaves in that subtree. We can then define: 
\begin{center}%
$R(t_a,\tilde{\x}(s_a), \y(s_a)) = \frac{1}{N}\sum_{i \in s_a} \mathbf{1}[y_i \neq \hat{y}_i^{t_{a}}] + \lambda H_{t_{a}}$.
\end{center}
For any dataset partition $A$, where $a\in A$ corresponds to the data handled by a given subtree of $t$:
\begin{center}%
$R(t, \tilde{\x}, \y) = \sum_{a \in A} R(t_a, \tilde{\x}(s_a), \y(s_a)) $. 
\end{center}

Given a set of observations $s_a$ and a depth constraint $d$ for which we want to find an optimal tree, consider the partition of $s_a$ resulting from ``splitting'' on a given boolean feature $j$ in $\tilde{\x}$, i.e., the sets given by:
\begin{align*}
s_a \cap s_j &= \left\{i:a(\tilde{\x}_i) \land (\tilde{\x}_{ij} = 1), i \in \{1,...,N\}  \right\}\\
s_a \cap s^c_{j} &= \left\{i:a(\tilde{\x}_i)\land (\tilde{\x}_{ij} = 0), i \in \{1,...,N\}  \right\}.
\end{align*}
Let $t^*_{a \cap j, d-1}$ be the optimal solution to the set of observations $s_a \cap s_j$ with depth constraint $d-1$, and let $t^*_{{a \cap j^c, d-1}}$ be similarly defined for the set of observations $s_a \cap s^c_j$. 
To find the optimal tree given the indices $s_a$ and the depth constraint $d$, one approach is to find the first split by solving
\begin{align} \label{eq:recursive formulation}
\underset{j}{\text{minimize}} & \left( R(t^*_{a \cap j, d-1}, \tilde{\x}(s_a \cap s_j), \y(s_a \cap s_j)) \nonumber \right. \\ & \left. + R(t^*_{a \cap j^c, d-1}, \tilde{\x}(s_a \cap s^c_j), \y(s_a \cap s_j^c)) \right). 
\end{align}
This leads to recursive subproblems representing different sets of observations and depth constraints. One can solve for all splits recursively according to the above equation. Our modification to use lower bound guessing informed by a reference model applies to all branch-and-bound optimization techniques that have this structure. 

Define $\text{majority}(s_a)$ as the majority label class in $s_a$,
and let $ub(s_a)$ be the objective of a single leaf predicting $\textrm{majority}(s_a)$ for each point in $s_a$:
\begin{center}%
$ub(s_a) = \frac{1}{N} \sum_{i \in s_a} \mathbf{1}[y_i \neq \text{majority}(s_a)] + \lambda,$
\end{center}
where (as usual) $\lambda$ is a fixed per-leaf penalty for the model. 
A recursive branch-and-bound algorithm using lower bound guessing finds a solution for observations $s_a$ with depth constraint $d$ with the following modifications to its ordinary approach (see Appendix \ref{app:lb-algs} for further details). Note that the subproblem is identified by $(s_a, d)$.

\subsubsection*{Lower-bound Guessing for Branch-and-Bound Search}
\begin{enumerate}\label{alg:lbg}
    \item \label{alg:lbg_basecase} \textit{(Use guess to initialize lower bound.)} If {$ ub(s_a) \leq lb_{\textrm{guess}}(s_a) + \lambda$} or $d=0$, we are done with the subproblem. We consider it solved with objective $ub(s_a)$, corresponding to a single leaf predicting $\text{majority}(s_a)$. 
    Otherwise, set the current lower bound $lb_\textrm{curr}(s_a, d) = lb_\textrm{guess}(s_a)$ and go to Step 2.
    \item Search the space of possible trees for subproblem $(s_a, d)$ by exploring the possible features on which the subproblem could be split to form two new subproblems, with depth constraint $d-1$, and solving those recursively. While searching:
    \begin{enumerate}
        \item \label{alg:lbg_earlyexit} \textit{(Additional termination condition if we match black box performance.)} If we find a subtree $t$ for subproblem $(s_a, d)$ with $R(t, \tilde{\x}(s_a), \y(s_a)) \leq lb_{\textrm{current}}(s_a, d)$, we are done with the subproblem. We consider it solved with objective $R(t, \tilde{\x}(s_a), \y(s_a))$, corresponding to subtree $t$.
        \item \label{alg:lbg_increase} \textit{(Modified lower bound update.)} At any time, we can opt to update the lower bound for the subproblem (with $d'$ as shorthand for $d-1$): 
        \[
        lb_{\textrm{spl}} \! \gets \! \min_{j \in \textrm{features}}(lb_\textrm{curr}(s_a \cap s_j, d') + lb_\textrm{curr}(s_a \cap s^c_j, d'))\]
        (This corresponds to splitting on the best feature.)
        \[
        lb_\textrm{curr}(s_a, d) \!\gets\! \max(lb_\textrm{curr}(s_a, d), \min (ub(s_a), lb_{\textrm{spl}}))
        \]
        (Here, $\min (ub(s_a), lb_{\textrm{spl}})$ is the better of not splitting on any feature and splitting on the best feature. The max ensures that our lower bounds never decrease. The $lb_\textrm{curr}$ term comes from either our initial lower bound guess or the lower bound we have so far.)
        If the lower bound increases to match $ub(s_a)$,
        we are done with the subproblem. We consider it solved with objective $ub(s_a)$, corresponding to a single leaf predicting $\text{majority}(s_a)$. If the lower bound increases but does not match $ub(s_a)$, we apply Case \ref{alg:lbg_earlyexit} with the new, increased $lb_\textrm{curr}(s_a, d)$. 
    \end{enumerate}
\end{enumerate}
This approach is provably close to optimal when the reference model makes errors similar to an optimal tree.
Let $\MC$ be the set of observations incorrectly classified by the reference model $T$, i.e., $\MC = \{i |y_i \neq \hat{y}_i^T\}$.

Let $t_{\textrm{guess}}$ be a tree returned from our lower-bound guessing algorithm. 
The following theorem holds: 

\begin{theorem} \label{thm:glb} (Guarantee on guessed model performance).
Let $R(t_{\textrm{guess}}, \tilde{\x}, \y)$ denote the objective of $t_\textrm{guess}$ on the full binarized dataset $(\tilde{\x}, \y)$ for some per-leaf penalty $\lambda$ (and with $t_\textrm{guess}$ subject to depth constraint d). Then for any decision tree $t$ that satisfies the same depth constraint d, we have: \begin{align*}R(t_{\textrm{guess}}, \tilde{\x}, \y) \leq &\frac{1}{N} \left(|\MC| + \sum_{i \in \CC} \mathbf{1}[y_i \neq \hat{y}_i^{t}]\right) \\ & + \lambda H_{t}.\end{align*}
That is, the objective of the guessing model is no worse than the union of errors made by the reference model and tree $t$.
\end{theorem}

The most significant consequence of this theorem is that 
\begin{eqnarray*}
\lefteqn{R(t_{\textrm{guess}}, \tilde{\x}, \y) - R(t^*,\tilde{\x},\y)} \\
&\leq& \frac{1}{N}\left(|\MC|- \sum_{i \in \MC} \mathbf{1}[y_i \neq \hat{y}_i^{t^*}]\right)
\end{eqnarray*} 
where we selected $t^*$, an optimal tree for the given $\lambda$ and depth constraint, as $t$, and we subtracted $R(t^*,\tilde{\x},\y)$ from both sides. This means that if the data points misclassified by the black box are a subset of the data points misclassified by the optimal decision tree, we are guaranteed not to lose any optimality. However much optimality we lose is a function of how many data points misclassified by the black box could have been correctly classified by the optimal tree.

We also prove that adding lower bound guessing after using threshold guessing does not change the worst case performance from Theorem \ref{thm:threshold_guess}.
\begin{corollary}\label{thm:glb-born-again}
 Let $T$, $T'$ and $t'$ be defined as in Theorem \ref{thm:threshold_guess}. 
Let $t_\textrm{guess}$ be the tree obtained using lower-bound guessing with $T'$ as the reference model, on $\tilde{\x}_T^{\textrm{reduced}}, \y$, with depth constraint matching or exceeding the depth of $t'$. Then 
$\mathcal{L}(t_\textrm{guess}, \tilde{\x}_T^{\textrm{reduced}}, \y) - \mathcal{L}(T, \x, \y) \leq \lambda (H_{t'} - H_{t_\textrm{guess}})$, or
equivalently, 
${R}(t_\textrm{guess}, \tilde{\x}_T^{\textrm{reduced}}, \y) \leq \mathcal{L}(T, \x, \y) + \lambda H_{t'}$.
\end{corollary}
The difference between this corollary and Theorem \ref{thm:threshold_guess} is that it uses $t_\textrm{guess}$ rather than $t^*$, without any weakening of the bound.
Because of this corollary, our experimental results focus on lower bound guessing in combination with threshold guessing, using the same reference model, rather than using lower bound guessing on its own.  

\section{Experiments}
\label{sec:eval}

Our evaluation addresses the following questions: 
\begin{enumerate}
 \item How accurate are interpretable models with guessing relative to black box models? (\S\ref{sec:guessing-behavior})
 \item How well does each guessing method perform? (\S\ref{sec:guessing-ablation}) 
 \item What happens if guesses are wrong? 
 (\S\ref{sec:bad-guess})
 \item What do sparse accurate trees look like? (\S\ref{sec:optimaltrees})
\end{enumerate}

We use seven datasets: one simulated 2D spiral pattern dataset, the Fair Isaac (FICO) credit risk dataset \citep{competition} for the Explainable ML Challenge, three recidivism datasets (COMPAS, \citealp{LarsonMaKiAn16}, Broward, \citealp{wang2020pursuit}, Netherlands, \citealt{tollenaar2013method}), and two coupon datasets (Takeaway and Restaurant), which were collected on Amazon Mechanical Turk via a survey \citep{wang2015or}. Table \ref{tab:data_small} summarizes all the datasets.

\begin{table}[ht]
    \small
    \centering
    \begin{tabular}{|c|c|c|c|}\hline
    Dataset & samples & features & binary features\\\hline
    Spiral & 100 & 2 & 180 \\\hline
    COMPAS & 6907 & 7 & 134 \\\hline
    Broward & 1954 & 38 & 588 \\\hline
    Netherlands & 20000 & 9 & 53890 \\\hline
    FICO & 10459 & 23 & 1917 \\\hline
    Takeaway & 2280 & 21 & 87 \\ \hline
    Restaurant & 2653 & 21 & 87 \\ \hline
    \end{tabular}
    \caption{Datasets}
    \label{tab:data_small}
\end{table}

Unless stated otherwise, all plots show the median value across 5 folds with error bars corresponding to the first and third quartile. We use GBDT as the reference model for guessing and run it using 
scikit-learn~\cite{scikit-learn}. 
We configure GBDT with default parameters, but select dataset-specific values for the depth and number of weak classifiers:  $(\textrm{n}\_\textrm{est}, \textrm{max}\_\textrm{depth}) = (20,3)$ for COMPAS and Spiral, $(40, 1)$ for Broward and FICO, $(50,2)$ for Takeaway and Restaurant, and $(30, 2)$ for Netherlands. Appendix~\ref{app:more-experiments} presents details about our hyper-parameter selection process, and Appendix~\ref{app:experiment-setup} presents the experimental setup and more details about the datasets.

\begin{figure*}[t]
    \centering
    \includegraphics[scale=0.45]{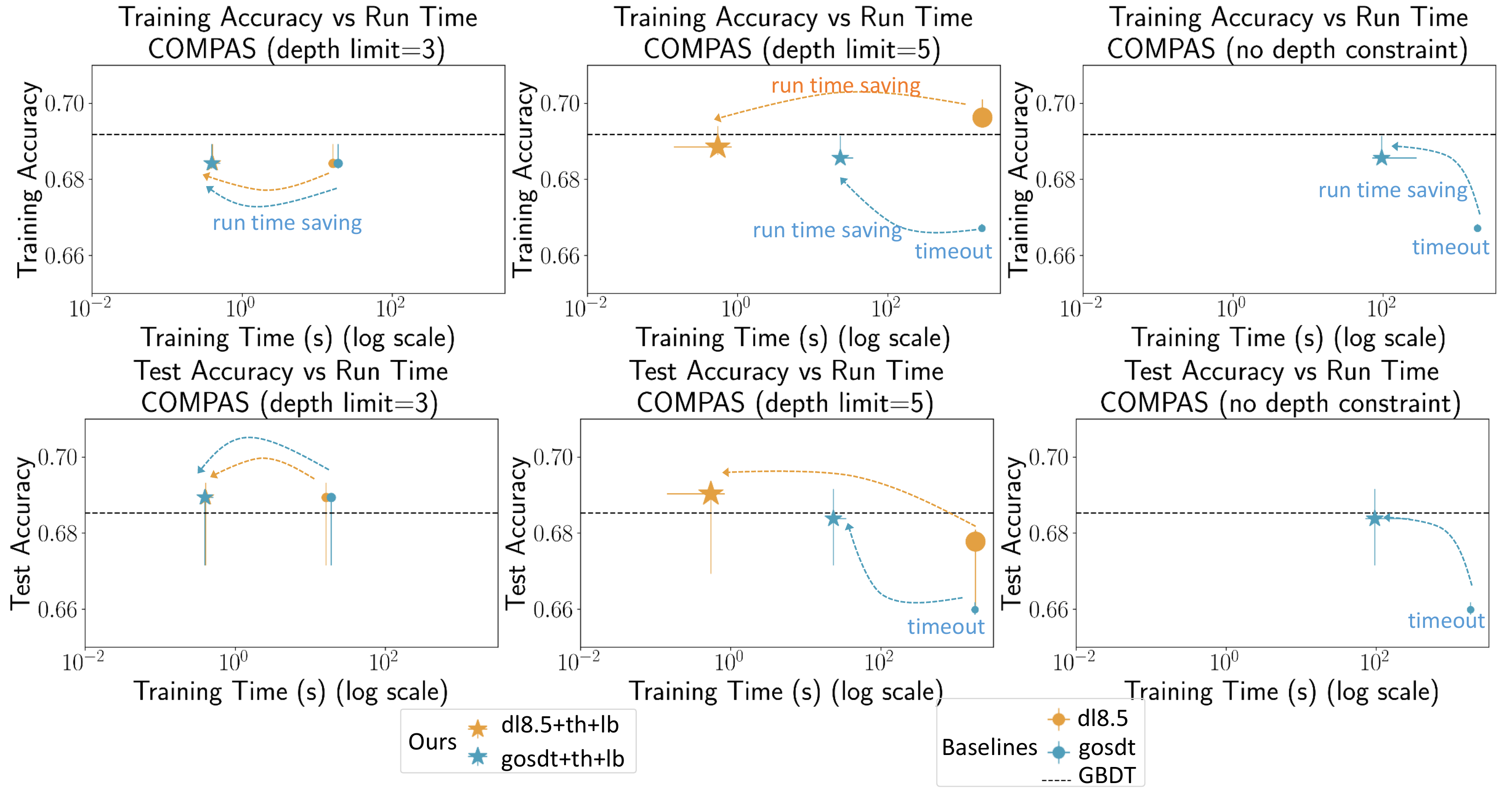}
    \caption{{(Train-Time Savings from Guessing)}. Training and test accuracy versus run time for GOSDT (blue) and DL8.5 (gold) on the COMPAS data set with regularization 0.001 and different depth constraints. The black line shows the accuracy of a GBDT model (100 max-depth 3 weak classifiers). 
    Circles show baseline performance (no guessing), stars show performance with all three guessing techniques, and marker size indicates the number of leaves. The displayed confidence bands come from 5-fold cross-validation. DL8.5 requires a depth constraint, so it does not appear in the right-most plots. }
    \label{fig:guessing-impact}
\end{figure*}

\subsection{Performance With and Without Guessing}\label{sec:guessing-behavior}
Our first experiments support our claim that using guessing strategies enables optimal decision tree algorithms to quickly find sparse trees whose accuracy is competitive with a GBDT that was trained using 100 max-depth 3 weak classifiers.

Figure~\ref{fig:guessing-impact} shows \textbf{training and testing results} on the COMPAS data set. (Results for the other data sets are in Appendix~\ref{app:overall-benefit}.)
The difference between the stars and the circles shows that \textit{our guessing strategies dramatically improve both DL8.5 and GOSDT run times, typically by 1-2 orders of magnitude.}
Furthermore, \textit{the trees we created have test accuracy that sometimes beats the black box}, because the sparsity of our trees acts as regularization. 

\begin{figure*}[ht]
    \centering
    \includegraphics[scale=0.43]{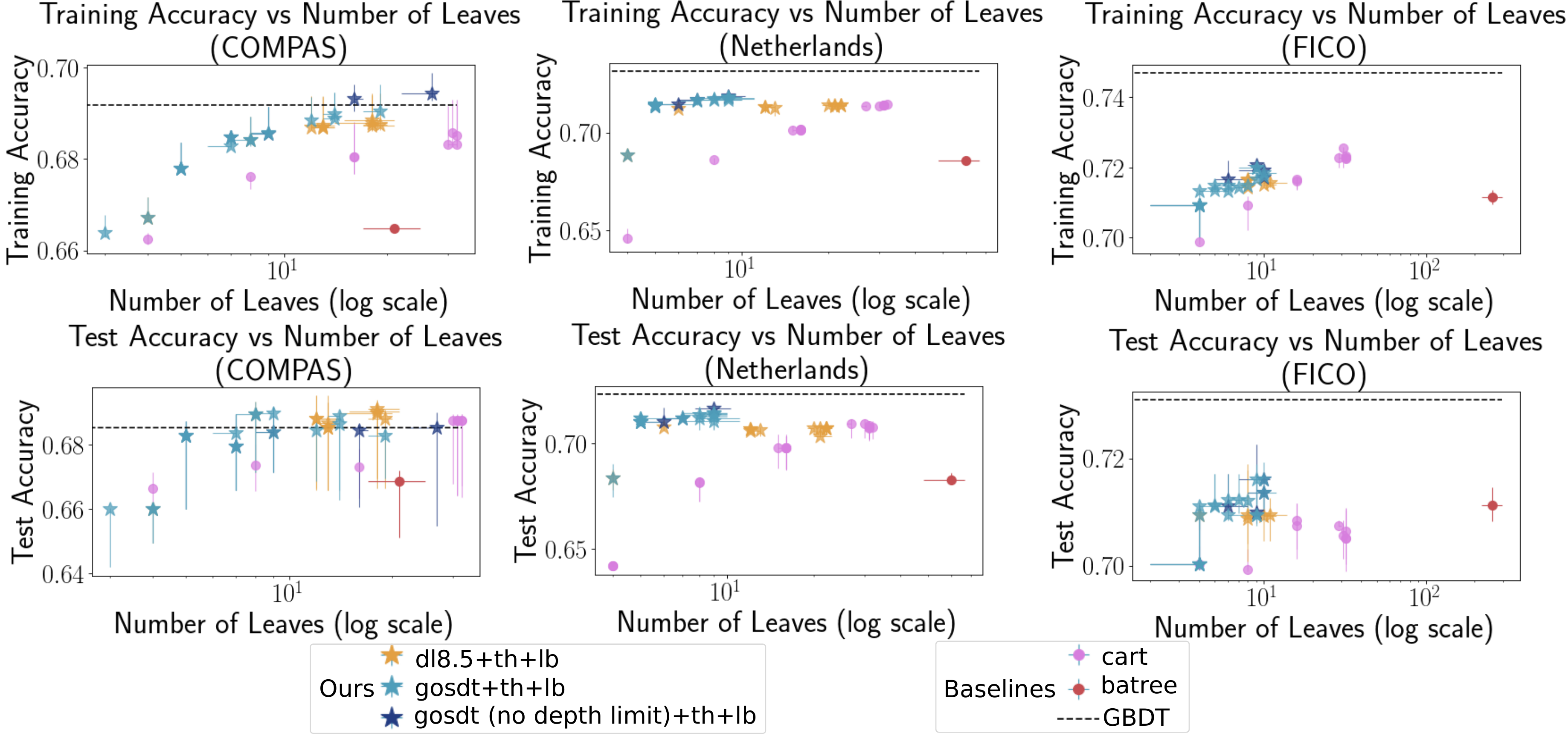}
    \caption{{(Sparsity vs$.$ Accuracy)}. DL8.5 and GOSDT use guessed thresholds and guessed lower bounds. CART is trained on the original dataset with no guess. This figure shows that our guessed trees define a frontier; we achieve the highest accuracy for almost every level of sparsity. The baseline methods (CART, batree) do not achieve results on the frontier.}
    \label{fig:all-the-models}
\end{figure*}

Figure~\ref{fig:all-the-models} shows the \textbf{accuracy-sparsity tradeoff} for different decision tree models (with GBDT accuracy indicated by the black line).
The \textit{batree} results are for the Born-Again tree ensembles of~\citet{vidal2020born}.
Results for the remaining datasets (which are similar) appear in Appendix~\ref{app:acc-v-sparsity}. The guessed models are both sparse and accurate, especially with respect to CART and batree.

We also compare GOSDT with all three guessing strategies to Interpretable AI's Optimal Decision Trees package, a proprietary commercial adaptation of Optimal Classification Tree (OCT) \citep{bertsimas2017optimal} 
in Appendix \ref{app:oct}. The results show the run time of the two methods is comparable despite the fact that GOSDT provides guarantees on accuracy while the adaptation of OCT does not. GOSDT with all three guesses tends to find sparser trees with comparable accuracy.

\subsection{Efficacy of Different Guessing Techniques}\label{sec:guessing-ablation}

Comparing the three graphs in Figure~\ref{fig:guessing-impact} shows how guessing different depths affects run time performance. For the next three experiments, we fix the maximum depth to 5 and examine the impact of the other guessing strategies.

\begin{figure}
    \centering 
    \includegraphics[scale=0.265]{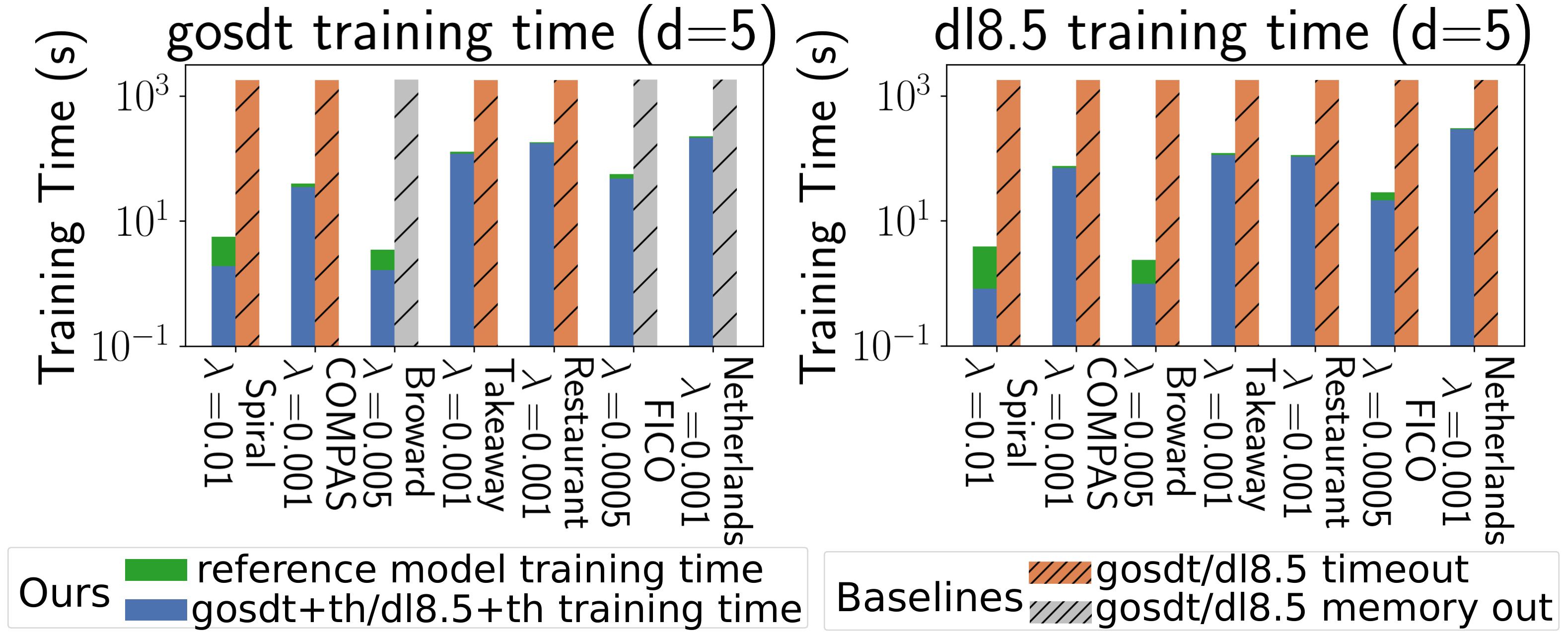}
    \caption{{(Value of Threshold Guessing)}.
    Training time (logscale) for GOSDT and DL8.5 with depth limit 5. All baselines (hatched) timed out (orange) or hit the memory limit (gray). Green parts indicate threshold guessing times.
    }
    \label{fig:abl_gt}
\end{figure}

\textbf{Guessing Thresholds:}
We found that guessing thresholds has the most dramatic impact, so we quantify that
impact first.
We compare the training time and accuracy with and without threshold guessing for GOSDT and DL8.5 on seven datasets. 
We limit run time to 30 minutes and run on a machine with 125GB memory.
Experiments that time out are shown with orange bars with hatches; Experiments that run out of memory are in gray with hatches.
If the experiment timed out, we analyze the best model found at the time of the timeout; if we run out of memory, we are unable to obtain a model.
Figure \ref{fig:abl_gt} shows that the \textit{training time after guessing thresholds (blue bars) is orders of magnitude faster than without thresholds guessing (orange bars)}. Orange bars are bounded by the time limit (1800 seconds) set for the experiments, but in reality, the training time is much longer. Moreover, threshold guessing leads to little change in both training and test accuracy (see Appendix \ref{app:thresholds}). 

\begin{figure}
    \centering 
    \includegraphics[scale=0.2645]{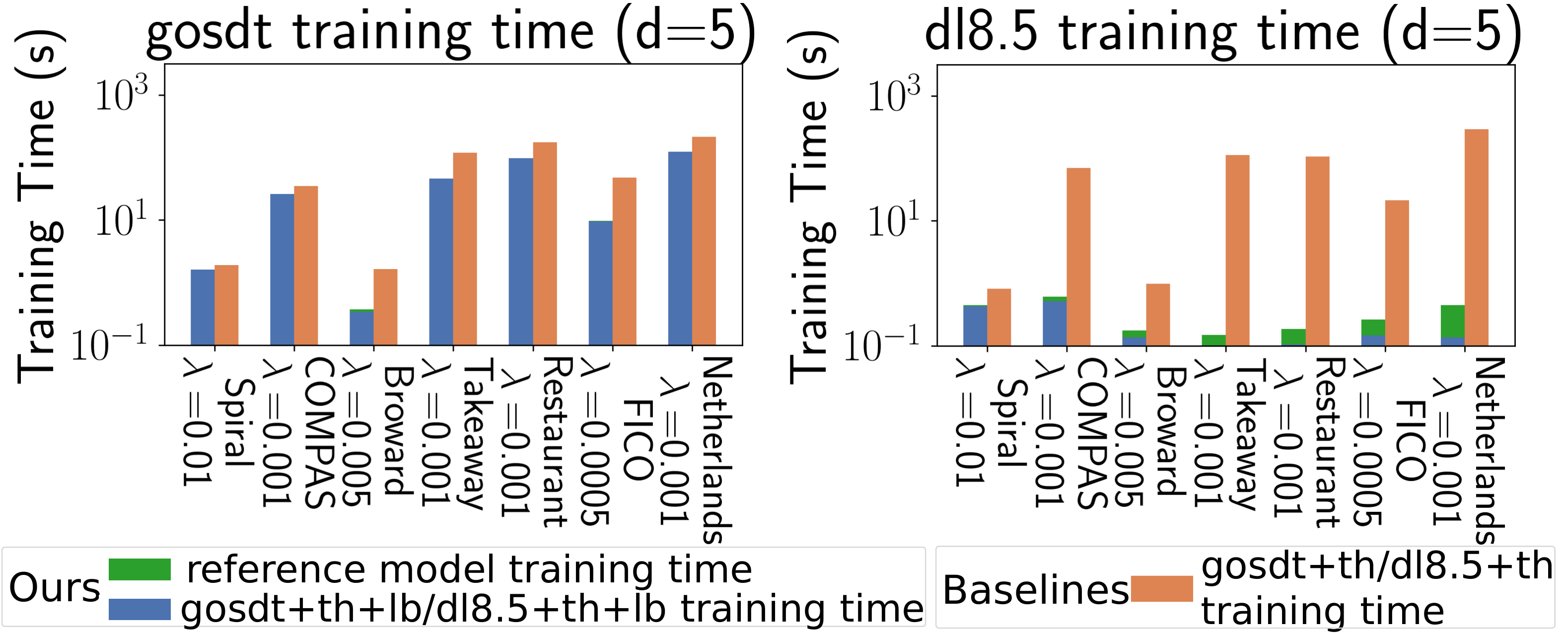}
    
    \caption{{(Value of Lower Bound Guessing)}. 
    Training time (logscale) for GOSDT and DL8.5 with and without lower-bound guessing, using depth limit 5 and threshold-guessing. 
    }
    \label{fig:guess_thresh_lb}
\end{figure}

\textbf{Guessing Lower Bounds:}
Next, we add lower-bound guesses to the threshold guesses to produce  Figure~\ref{fig:guess_thresh_lb}. The results are qualitatively similar to those in Figure \ref{fig:abl_gt}: using lower bound guesses produces an additional run time benefit. Appendix \ref{app:lowerbounds} shows more results and verifies that using lower-bound guesses often leads to a solution with the same training error as without lower-bound guesses or leads to finding a simpler solution with test error that is still comparable to the solution found without lower-bound guessing.

\begin{figure}
    \centering
    \includegraphics[scale=0.24]{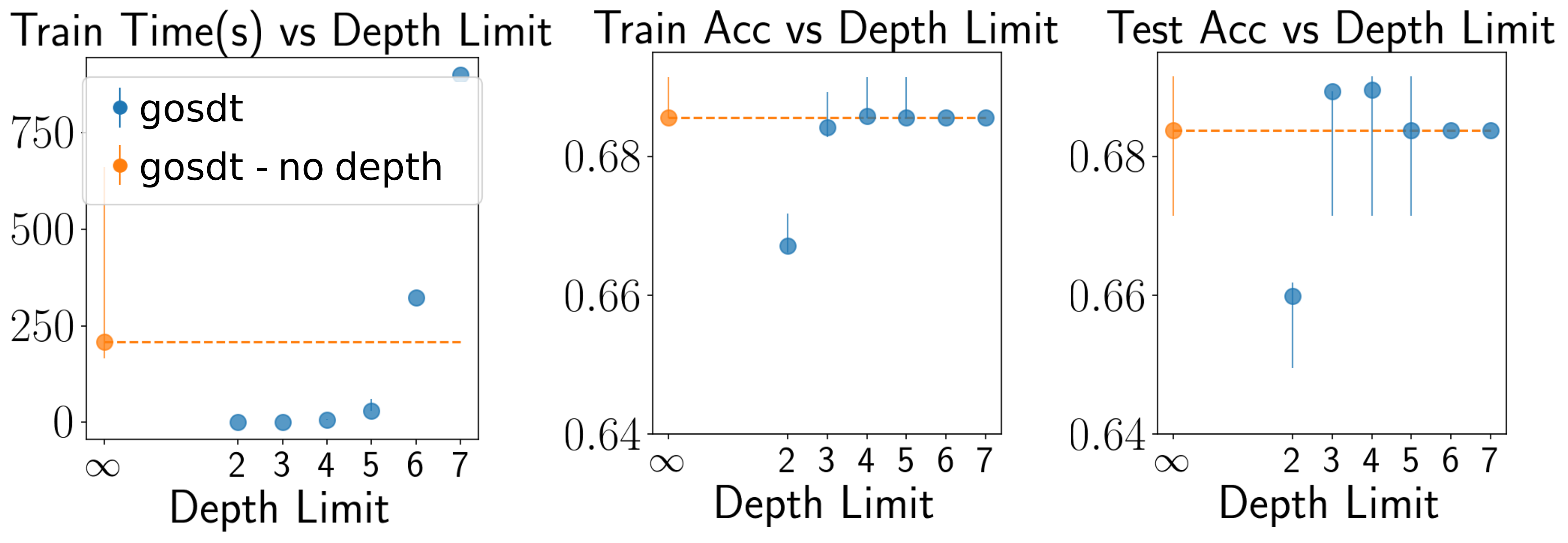}
    \caption{{(Value of Depth Constraints)}. Performance across depth constraints ($\infty$ means no constraint) for the COMPAS dataset with threshold guessing. Regularization 0.001.}
    \label{fig:depth-guess}
\end{figure}

\textbf{Value of Depth Constraints:}
We examine depth constraints' effect on GOSDT in Figure~\ref{fig:depth-guess}. We run GOSDT without a depth constraint, which produces an optimal tree for each of five folds using per-leaf regularization, and compare to GOSDT with a depth constraint. 
Above a certain threshold, depth constraints do not reduce training accuracy (since the constraint does not remove all optimal trees from the search space). Some depth constraints slightly improve test accuracy (see depths 4 and 5), since constraining GOSDT to find shallower trees can prevent or reduce overfitting. 

Usually, depth constraints allow GOSDT to run several orders of magnitude faster. For large depth constraints, however, the run time can be worse than that with no constraints. Using depth guesses reduces GOSDT's ability to share information between subproblems. When the constraint prunes enough of the search space, the search space reduction dominates; if the constraint does not prune enough, the inability to share subproblems dominates.


\subsection{When Guesses Are Wrong}\label{sec:bad-guess}
Although rare in our experiments, it is possible to lose accuracy if the reference model is too simple, i.e., only a few thresholds are in the reduced dataset or too many samples are misclassified by the reference model. This happens when we choose a reference model that has low accuracy to begin with. Thus, we should check the reference model's performance before using it and improve it before use if necessary. Appendix \ref{app:bad-guess} shows that if one uses a poor reference model, we do lose performance. Appendix \ref{app:more-experiments} shows that for a wide range of reasonable configurations, we do not suffer dramatic performance costs as a result of guessing. 

\subsection{Trees}\label{sec:optimaltrees}
\begin{figure}[ht]
    \centering
    \includegraphics[scale=0.33]{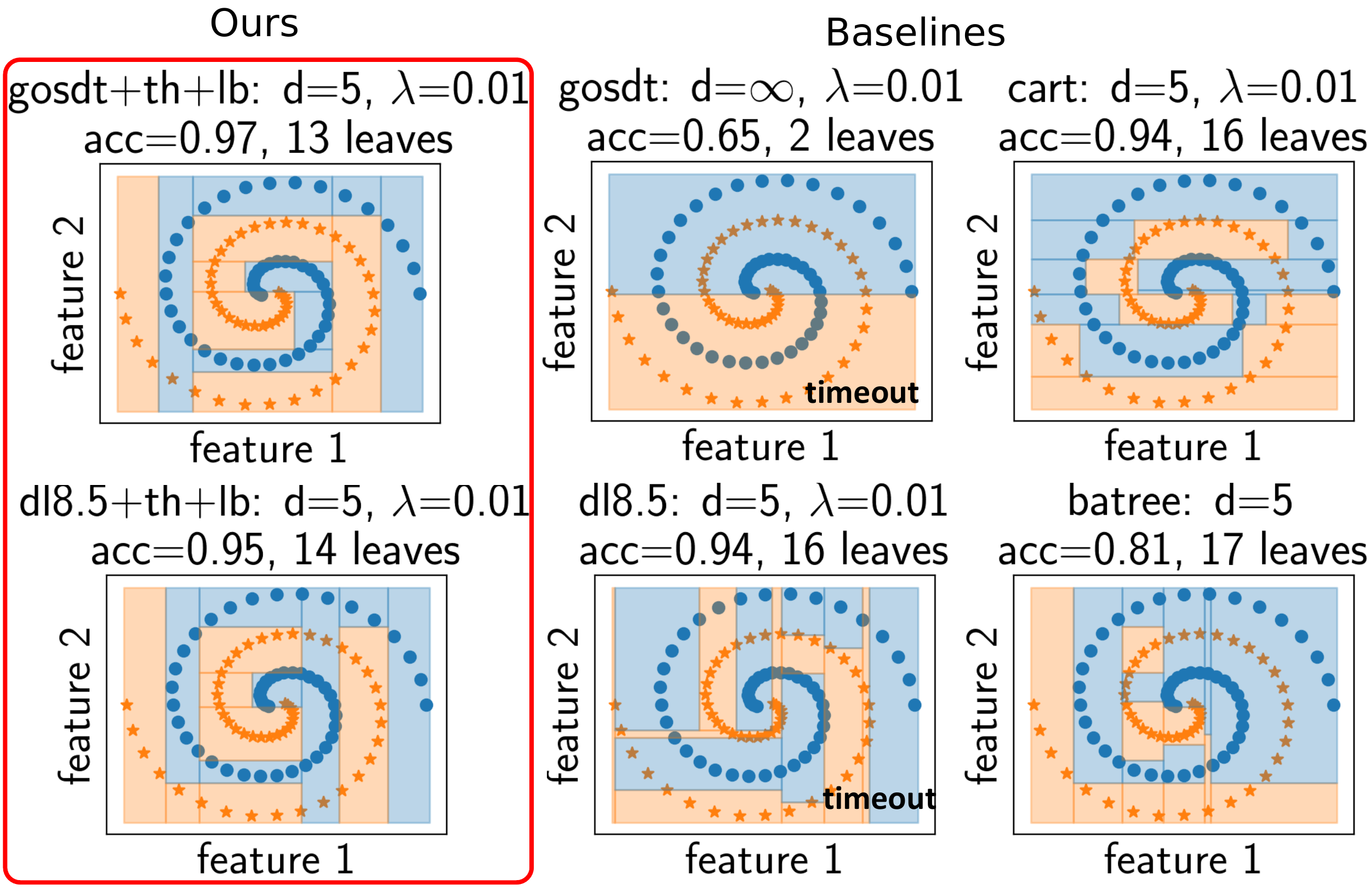}
    \caption{{(Tree Visualization)}. 
    Trees for GOSDT and DL8.5. 
    Dots indicate the data points and the shaded areas are the corresponding classification by the tree. 
    GOSDT trees were trained with and without all three guesses ($d=\infty$ means no depth constraint). DL8.5 trees were trained with and without threshold and lower bound guesses at depth 5. CART was trained with depth 5 and batree was trained using random forests with 10 max-depth 3 weak classifiers as the reference model. Training time is 1.606 seconds for GOSDT with all three guesses and 0.430 seconds for DL8.5 with all three guesses (left column). Both GOSDT without guesses and DL8.5 with only depth guesses timed out (middle column).
    }
    \label{fig:trees}
\end{figure}

\begin{figure}[t]
    \centering
    \begin{subfigure}[b]{0.48\textwidth}
    \centering
        \scalebox{0.6}{\begin{forest}
    [ $age \le 27.5$
    [ $age \le 21.5$, edge label={node[midway, above] {True}} [ $class$ [ $1$ ] ] [ $priors\_count \le 1.5$ [ $class$ [ $0$ ] ] [ $class$ [ $1$ ] ] ] ] [ $age \le 36.5$,edge label={node[midway, above] {False}} [ $priors\_count \le 2.5$ [ $class$ [ $0$ ] ] [ $class$ [ $1$ ] ] ] [ $priors\_count \le 7.5$ [ $class$ [ $0$ ] ] [ $class$ [ $1$ ] ] ] ] ]
    \end{forest}}
    \caption{GOSDT+th+lb (depth limit 3): training time=0.513 (sec), training accuracy=0.683, test accuracy=0.689, \#leaves=7 on fold 2.}
    \end{subfigure}
    \hfill
    \begin{subfigure}[b]{0.48\textwidth}
    \centering
\scalebox{0.6}{\begin{forest}
[ $age \le 19.5$ [ $class$, edge label={node[midway, above] {True}} [ $1$ ] ] [ $priors\_count \le 2.5$, edge label={node[midway, above] {False}} [ $age \le 22.5$ [ $sex\ not \ female$ [ $class$ [ $1$ ] ] [ $class$ [ $0$ ] ] ] [ $age \le 27.5$ [ $priors\_count \le 1.5$ [ $class$ [ $0$ ] ] [ $class$ [ $1$ ] ] ] [ $class$ [ $0$ ] ] ] ] [ $age \le 36.5$ [ $class$ [ $1$ ] ] [ $priors\_count \le 7.5$ [ $class$ [ $0$ ] ] [ $class$ [ $1$ ] ] ] ] ] ]
\end{forest}}
\caption{GOSDT+th+lb (depth limit 5):training time=34.804 (sec), training accuracy=0.685, test accuracy=0.696, \# leaves=9 on fold 2. }
    \end{subfigure}
    \caption{(Example Trees). GOSDT trees on the COMPAS dataset with guessed thresholds and guessed lower bounds at depth limit 3 and 5. The reference model was trained using 20 max-depth 3 weak classifiers. $\lambda=0.001$.}
    \label{fig:compas_tree_main}
\end{figure}
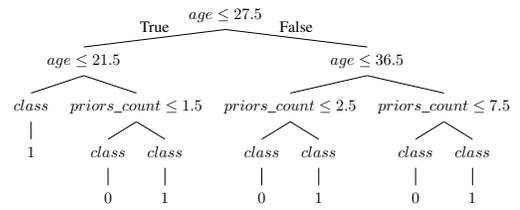
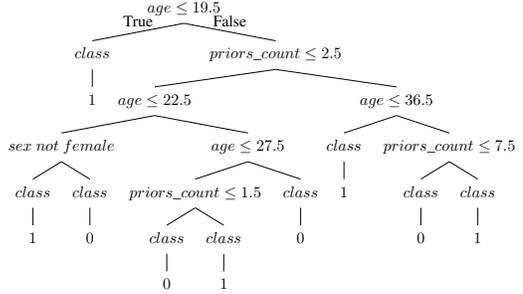

We qualitatively compare trees produced with our guessing technique to several baselines in Figure~\ref{fig:trees}. 
We observe that with all three guessing techniques, the resulting trees are not only more accurate overall, but they are also sparser than the trees produced by the baselines. Figure \ref{fig:compas_tree_main} shows example GOSDT trees trained using all three guesses on the COMPAS dataset. Appendix \ref{app:trees} shows more output trees and Appendix \ref{app:fico} compares decision sets transformed from our trees and trained by \citet{dash2018boolean} on the FICO dataset.  


\section{Conclusion}
We introduce smart guessing strategies to find sparse decision trees that compete in accuracy with a black box machine learning model while reducing training time by orders of magnitude relative to training a fully optimized tree. 
Our guessing strategies can be applied to several existing optimal decision tree algorithms with only minor modifications.
With these guessing strategies, powerful decision tree algorithms can be used on much larger datasets. 

\section*{Code Availability}
Implementations of GOSDT and DL8.5 with the guessing strategies discussed in this paper are available at https://github.com/ubc-systopia/gosdt-guesses and https://github.com/ubc-systopia/pydl8.5-lbguess. Our experiment code is available at https://github.com/ubc-systopia/tree-benchmark. 

\section*{Acknowledgements}
We acknowledge the following grant support: NIDA DA054994-01, DOE DE-SC0021358-01, NSF DGE-2022040, NSF FAIN-1934964. We acknowledge the support of the Natural Sciences and Engineering Research Council of Canada (NSERC). Nous remercions le Conseil de recherches en sciences naturelles et en génie du Canada (CRSNG) de son soutien.

\normalem
\bibstyle{aaai22}
\bibliography{refs}

\appendix
\onecolumn

\section{Theorems and Proofs}
\label{app:proofs}

\subsection{Proof of Theorem \ref{thm:threshold_guess}}\label{app:proof-4.1}
\textbf{Theorem \ref{thm:threshold_guess}}
\textit{(Guarantee for model on reduced data).
Define the following:
\begin{itemize}
\item Let $T$ be the ensemble tree built upon $(\x, \y)$. 
\item Let $T'$ be the ensemble tree built with $(\tilde{\x}_T^{\textrm{reduced}}, \y)$. 
\item Let $t'$ be any decision tree that makes the same predictions as $T'$ for every observation in $\tilde{\x}_T^{\textrm{reduced}}$. 
\item Let $t^* \in \arg\min_t \mathcal{L}(t, \tilde{\x}_T^{\textrm{reduced}}, \y) + \lambda H_t$ be an optimal tree on the reduced dataset. 
\end{itemize}
Then, $\mathcal{L}(t^*, \tilde{\x}_T^{\textrm{reduced}}, \y) - \mathcal{L}(T, \x, \y) \leq \lambda (H_{t'}- H_{t^*})$, or equivalently, 
$R(t^*, \tilde{\x}_T^{\textrm{reduced}}, \y) \leq \mathcal{L}(T, \x, \y) + \lambda H_{t'}$. }

\begin{figure}
    \centering
    \includegraphics[scale=0.4]{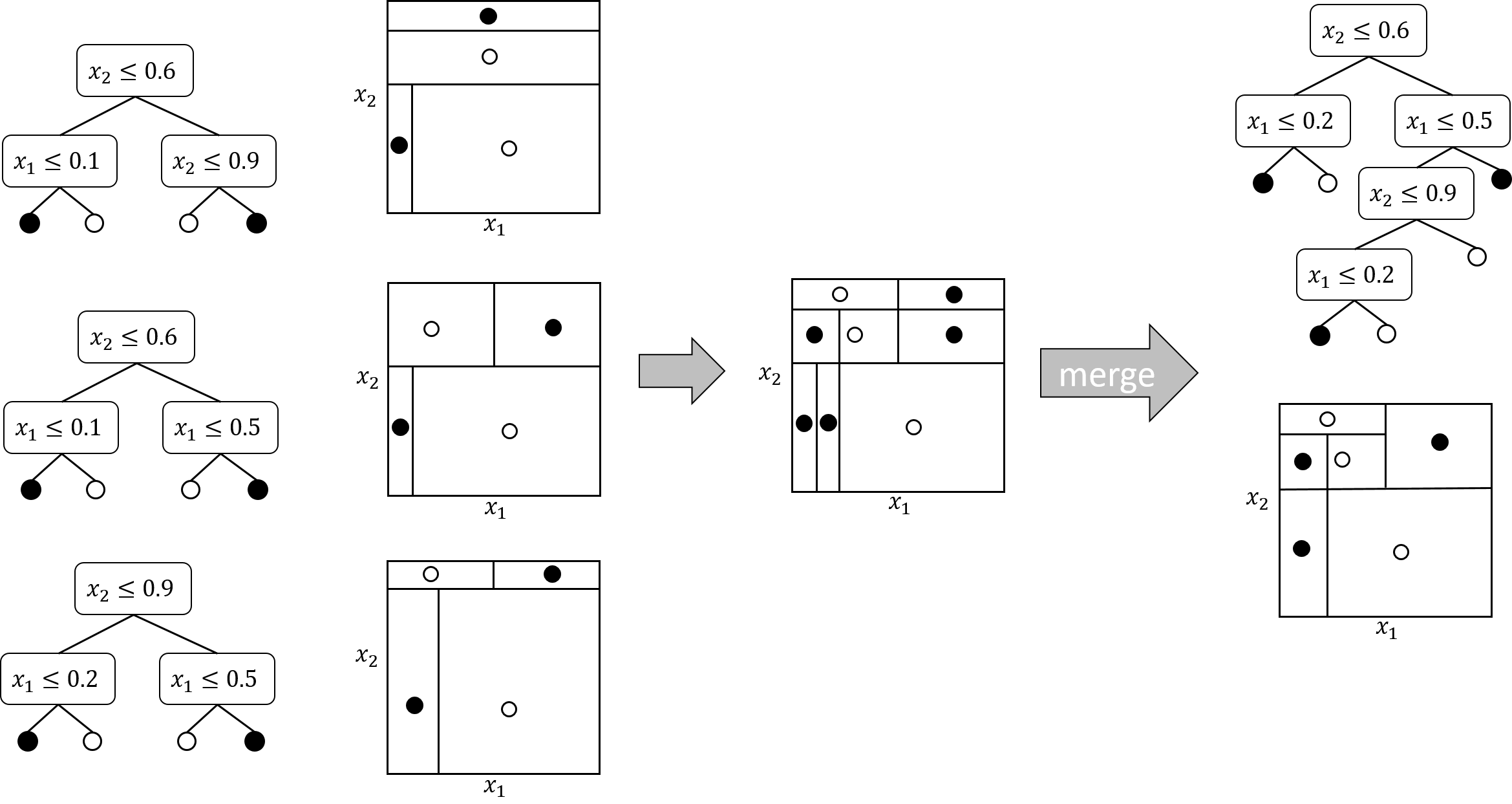}
    \caption{An example of getting a smallest single tree that exactly replicates the prediction of $T'$.}
    \label{fig:merge}
\end{figure}

\begin{proof}
When we are building $\tilde{\x}_T^{\textrm{reduced}}$ (the dataset used to construct $T'$), the stopping condition of column elimination is that accuracy decreases, which is when the loss increases: $\mathcal{L}(T, \x, \y) < \mathcal{L}(T', \tilde{\x}_T^{\textrm{reduced}}, \y)$. Thus, the opposite must always be true as we build $T'$: $\mathcal{L}(T', \tilde{\x}_T^{\textrm{reduced}}, \y) \leq \mathcal{L}(T, \x, \y)$. 
Since $H_{t'}$ is the number of leaves in the single tree that exactly replicates the predictions of $T'$ and since $t^*$ is an optimal tree, 
\begin{eqnarray*}
\mathcal{L}(t^*, \tilde{\x}_T^{\textrm{reduced}}, \y) + \lambda H_{t^*} 
&\leq& \mathcal{L}(T', \tilde{\x}_T^{\textrm{reduced}}, \y) + \lambda H_{t'}\\
&\leq&  \mathcal{L}(T, \x, \y) + \lambda H_{t'}.
\end{eqnarray*}
Then, $\mathcal{L}(t^*, \tilde{\x}_T^{\textrm{reduced}}, \y) - \mathcal{L}(T, \x, \y) 
\leq \lambda (H_{t'} - H_{t^*})$. 
\end{proof}

\subsection{Proof of Theorem \ref{thm:vcdepth}}
\textbf{Theorem \ref{thm:vcdepth}} \textit{Let $B$ be the base hypothesis class (e.g., decision stumps) that has VC dimension at least 3 and $K \geq 3$ be the number of weak classifiers. Let $\mathcal{F}_{\textrm{ensemble}}$ be the set of models that are a weighted sum of shallow trees for classification, i.e., $T\in \mathcal{F}_{\textrm{ensemble}}$ has $T(x)=\textrm{sign}(\sum_{k=1}^K w_k h_k(x))$,  where $\forall k, w_k \in \mathbb{R}, h_k \in B$. Let $\mathcal{F}_{d,\textrm{tree}}$ be the class of single binary decision trees with depth at most: \[d = \left\lceil \log_2 \left(\left(K \cdot \textrm{VC}(B)+K\right)\cdot\left(3\ln(K \cdot \textrm{VC}(B)+K\right)+2)\right)\right\rceil.\] It is then true that $\textrm{VC}(\mathcal{F}_{d,\textrm{tree}}) \geq \textrm{VC}(\mathcal{F}_{\textrm{ensemble}})$. }

\begin{proof}
Since $K$ and $VC(B)$ are both greater than or equal to 3 and since $T \in \mathcal{F}_{\textrm{ensemble}}$ has $T(x)=\textrm{sign}(\sum_{k=1}^K w_k h_k(x))$, where $\forall k, w_k \in \mathbb{R}, h_k \in B$, then according to \citet[Lemma 10.3]{shalev2014understanding},
$$VC(\mathcal{F}_{\textrm{ensemble}}) \leq \left(K\cdot VC(B)+K\right)\cdot\left(3\ln(K\cdot VC(B)+K))+2\right).$$
Since $\mathcal{F}_{\textrm{d, tree}}$ is the class of single binary decision trees with depth at most $d$, $2^d$ samples can be shattered by trees in $\mathcal{F}_{\textrm{d, tree}}$. That is, setting $d$ to the value in the statement of the theorem, $$VC(\mathcal{F}_{\textrm{d, tree}}) = 2^{\left\lceil \log_2 \left(\left(K \cdot \textrm{VC}(B)+K\right)\cdot\left(3\ln(K \cdot \textrm{VC}(B)+K\right)+2)\right)\right\rceil}.$$
Then, using that $2^{\lceil \log_2 \gamma \rceil} \geq \gamma$ for each $\gamma > 0$, we get $VC(\mathcal{F}_{\textrm{d, tree}}) \geq (K\cdot VC(B)+K)\cdot(3\ln(K\cdot VC(B)+K)+2) \geq VC(\mathcal{F}_{\textrm{ensemble}})$.
\end{proof}

\subsection{Bound for the gap between the objectives of optimal trees with a relatively smaller depth guess and with no depth guess}

Theorem \ref{thm:vcdepth} gives an upper bound of depth guess we can make to compete with a boosted tree. However, it is often too loose to be an effective guess. Ideally, we want to guess depth that matches the true depth, but it is likely that our guess is small. Theorem \ref{thm:depth_guess} bounds the gap between the objectives of optimal trees with a relatively smaller depth guess and with no depth guess. 

We first introduce some notation. Let $\Omega$ be a set of leaves. \textit{Capture} is an indicator function that equals 1 if $\tilde{\x}_i$ falls into one of the leaves in $\Omega$, and 0 otherwise, in which case we say that $\textrm{cap}(\tilde{\x}_i, \Omega) = 1$. For a dataset $\{(\tilde{\x}_i, y_i)\}_{i=1}^N$ 
we define a set of observations to be equivalent if they have exactly the same feature values, i.e., $\tilde{\x}_i = \tilde{\x}_j$ when $i\neq j$. 
Note that a dataset contains multiple sets of equivalent points (i.e., equivalence classes), denoted by $\{e_u\}_{u=1}^U$. Each observation belongs to an equivalence class $e_u$ (which has 1 element if the observation is unique). We denote $q_u$ as the minority label class among observations in $e_u$. In this case, the number of observations with the minority label in set $e_u$ is $\sum_{i=1}^N \mathbf{1}[\tilde{\x}_i \in e_u] \mathbf{1}[y_i = q_u]$. 

\begin{theorem}\label{thm:depth_guess}
Let $t^*$ be an optimal decision tree, i.e. $t^* \in \arg\min_t \mathcal{L}(t, \tilde{\x},\y) + \lambda H_t$ and $d^*$ be the depth of $t^*$. Define $d_{\textrm{guess}}$ as a guessed depth and assume that $d_{\textrm{guess}} < d^*$ (that is, the guessed depth is too small). Now, let $t_{\textrm{guess}}^* \in \arg\min_t \mathcal{L}(t, \tilde{\x},\y) + \lambda H_t$ s.t. $\textrm{depth}(t) \leq d_{\textrm{guess}}$ (which is the best tree we could create with the guessed depth). Let $t'_{\textrm{guess}}$ be the tree obtained by pruning all leaves below $d_{\textrm{guess}}$ in $t^*$ and let $(l_1, ..., l_{H_{t'_{\textrm{guess}}}})$ be leaves of $t'_{\textrm{guess}}$. Then, 
\begin{align*}
    R(t_{\textrm{guess}}^*,\tilde{\x},\y) - R(t^*,\tilde{\x},\y) &\leq R(t'_{\textrm{guess}},\tilde{\x},\y) - R(t^*,\tilde{\x},\y)\\
&\leq \frac{1}{N}\left( \sum_{i=1}^N \left(\sum_{h=1}^{H_{t'_{\textrm{guess}}}} \textrm{cap}(\tilde{\x}_i, l_h) \wedge \mathbf{1}[y_i \neq \hat{y}^{t'_{\textrm{guess}}}_i]- \sum_{u=1}^U \mathbf{1}[\tilde{\x}_i \in e_u] \mathbf{1}[y_i = q_u]\right)\right) \\
& \ \ \ \ \ - \lambda (H_{t^*} - H_{t'_{\textrm{guess}}}).
\end{align*}
\end{theorem}

\begin{figure}[ht]
    \centering
    \includegraphics[scale=0.32]{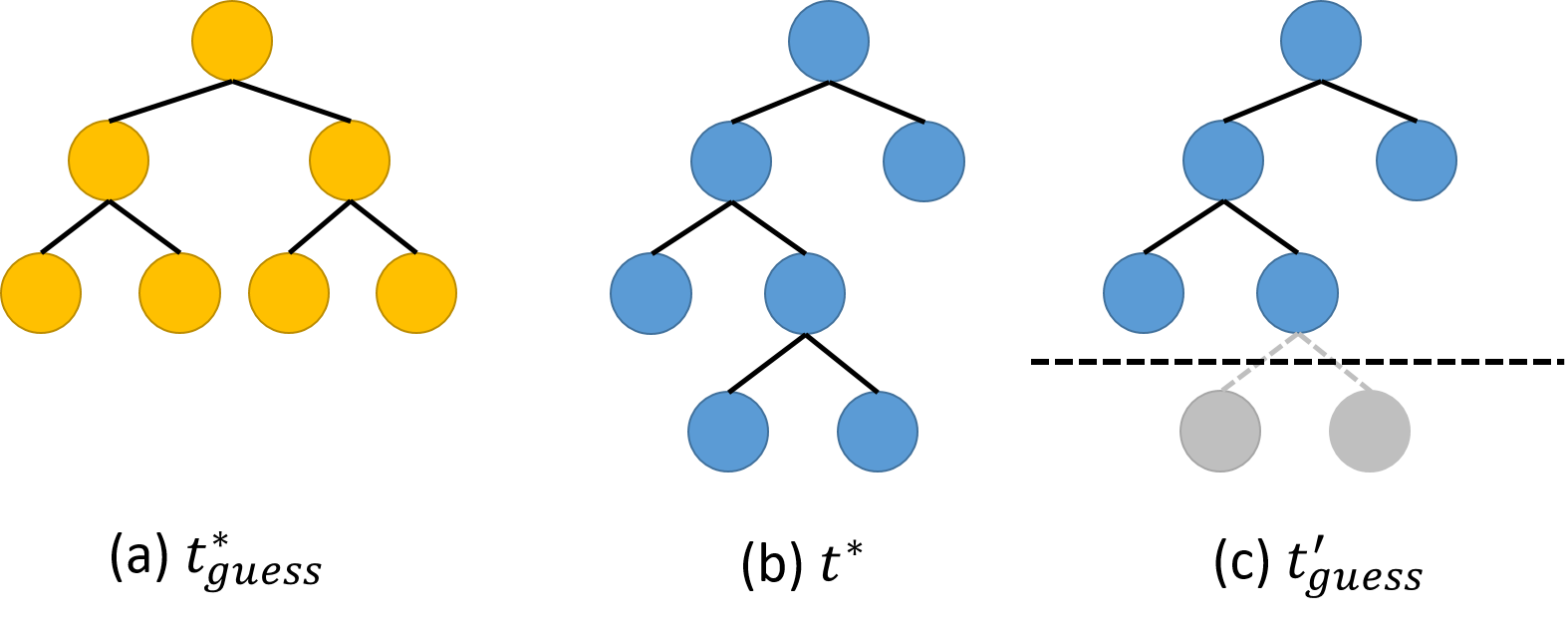}
    \caption{An example demonstrating $t^*_{\textrm{guess}}$, $t^*$, and $t'_{\textrm{guess}}$ in Theorem \ref{thm:depth_guess}. }
    \label{fig:depth_guess}
\end{figure}
Figure \ref{fig:depth_guess} shows an example of trees in Theorem \ref{thm:depth_guess}. Suppose the true depth of the optimal tree $t^*$ is 3 (subfigure b) and $d_{\textrm{guess}}$ is 2. In this toy example, we can remove the leaf pair at depth 3 in $t^*$ to get $t'_{\textrm{guess}}$ (subfigure c). And $t^*_{\textrm{guess}}$ shown in subfigure (a) is an optimal tree given depth constraint which can have different internal nodes compared with $t'_{\textrm{guess}}$.  

\begin{proof}
Since $t^*$ is the optimal tree with no depth guess and $t^*_{\textrm{guess}}$ is the optimal tree with depth guess where $d_{\textrm{guess}}$ is assumed to be smaller than $d^*$, $$R(t^*,  \tilde{\x}, \y) \leq R(t^*_{\textrm{guess}},  \tilde{\x}, \y).$$
Since $t'_{\textrm{guess}}$ is a tree obtained by pruning all leaves below $d_{\textrm{guess}}$ in $t^*$, $$R(t^*_{\textrm{guess}}, \tilde{\x}, \y) \leq R(t'_{\textrm{guess}}, \tilde{\x}, \y).$$ Therefore, 
\begin{equation*}
    \begin{aligned}
    R(t_{\textrm{guess}}^*, \tilde{\x}, \y) - R(t^*,  \tilde{\x}, \y) &\leq R(t'_{\textrm{guess}},  \tilde{\x}, \y) - R(t^*,  \tilde{\x}, \y) \\
    &= \frac{1}{N}\left(\sum_{i=1}^N \sum_{h=1}^{H_{t'_{\textrm{guess}}}} \textrm{cap}(\tilde{\x}_i, l_h) \wedge \mathbf{1}[y_i \neq \hat{y}^{t'_{\textrm{guess}}}_i]\right) + \lambda H_{t'_{\textrm{guess}}} - R(t^*, \tilde{\x}, \y).
    \end{aligned}
\end{equation*}

Moreover, since the loss of the optimal tree is always no less than the number of equivalent points over the sample size, i.e. $\frac{\textrm{\# of equivalent points}}{N}$, $$R(t^*, \tilde{\x}, \y)  = \mathcal{L}(t^*,\tilde{\x},\y) + \lambda H_{t^*} \geq \frac{1}{N} \left(\sum_{i=1}^N \sum_{u=1}^U \mathbf{1}[\tilde{\x}_i \in e_u] \mathbf{1}[y_i = q_u]\right) + \lambda H_{t^*}.$$ 
Hence, $R(t_{\textrm{guess}}^*, \tilde{\x}, \y) - R(t^*,  \tilde{\x}, \y) \leq \frac{1}{N} (\sum_{i=1}^N (\sum_{h=1}^{H_{t'_{\textrm{guess}}}} \textrm{cap}(\tilde{\x}_i, l_h) \wedge \mathbf{1}[y_i \neq \hat{y}^{t'_{\textrm{guess}}}_i] - \sum_{u=1}^U \mathbf{1}[\tilde{\x}_i \in e_u] \mathbf{1}[y_i = q_u])) - \lambda (H_{t^*} - H_{t'_{\textrm{guess}}}).$
\end{proof}

According to Theorem \ref{thm:depth_guess}, although it is possible to have a penalty to the objective when picking a depth constraint that is too small (because we no longer consider the optimal tree), this penalty is somewhat offset by the reduced complexity of the models in the new search space.

\subsection{Proof of Theorem \ref{thm:glb}}
\textbf{Theorem \ref{thm:glb}}
\textit{Let $R(t_{\textrm{guess}}, \tilde{\x}, \y)$ denote the objective of $t_\textrm{guess}$ on the full binarized dataset $(\tilde{\x}, \y)$ for some per-leaf penalty $\lambda$ (and with $t_\textrm{guess}$ subject to some depth constraint d). Then for any decision tree $t$ that satisfies the same depth constraint d, we have \begin{align*}R(t_{\textrm{guess}}, \tilde{\x}, \y) \leq &\frac{1}{N} \left(|\MC| + \sum_{i \in \CC} \mathbf{1}[y_i \neq \hat{y}_i^{t}]\right) + \lambda H_{t}.\end{align*}
That is, the objective of the guessing model is no worse than the union of errors made by the reference model and tree $t$.}

For this proof, we use the following notation to discuss lower bounds:
\begin{itemize} 

\item {$\lambda \geq 0$ is a regularizing term (a per-leaf penalty added to the risk function when evaluating each tree)} 

\item For some decision tree $t$, we calculate its risk as {${R}(t,\tilde{\x},\y) = \frac{1}{N}\sum_{i=1}^N \mathbf{1}[y_i \neq \hat{y}_i^t] + \lambda H_t$}, where $H_t$ refers to the number of leaves in tree $t$.

\item $T$ is a reference model we use to guess lower bounds.

\item $\CC$ and $\MC$ are, respectively, the indices of the set of observations in the training set correctly classified by $T$ and the set of observations incorrectly classified by $T$.

\item $s_a$ is the set of training set observations captured in our current subproblem.

\item $d$ represents the maximum allowed depth for solutions to our current subproblem. If $d=0$, no further splits are allowed. 


\item {$t_{\textrm{guess}}(s_a, d, \lambda)$} is the lower-bound-guessing-algorithm's solution for subproblem $s_a$, depth limit $d$, and regularizer $\lambda$. When we just specify $t_\textrm{guess}$ without arguments, we are referring to the lower-bound-guessing-algorithm's solution for the root subproblem (the whole dataset), with the depth limit argument provided for the tree as a whole. 

\item Consider a subproblem $s_a$ corresponding to the full set of points passing through a specific internal or leaf node of the optimal tree $t^*$ (call it node$_{t^*, s_a}$). Define $H_{t^*_a}$ as the number of leaves below node$_{s_a}$  (or 1 if node$_{s_a}$ is a leaf). Note that this is also the number of leaves needed in an optimal solution for subproblem $s_a$. Similarly define $H_{t_{\textrm{guess}, a}}$ as the number of leaves below node$_{t_\textrm{guess},s_a}$ in $t_\textrm{guess}$ (when $s_a$ corresponds to the full set of points passing through a node in $t_\textrm{guess}$). Note that this does not necessarily correspond to the number of leaves needed in an optimal solution for subproblem $s_a$ because $t_\textrm{guess}$ has not been fully optimized.

\item {$R_{\textrm{guess}}(s_a, d, \lambda)$} is the objective of the solution found for subproblem $s_a$, depth limit $d$, and regularizer $\lambda$ when we use lower bound guessing:
\begin{align*} R_{\textrm{guess}}(s_a, d, \lambda) := &\frac{1}{N}\sum_{i\in s_a} \mathbf{1}[y_i \neq \hat{y}_i^{t_{\textrm{guess}}(s_a, d, \lambda)}]  + \lambda H_{t_{\textrm{guess}, a}}.\end{align*} 

\item $lb_{\textrm{guess}}(s_a)$ was defined as {$$lb_{\textrm{guess}}(s_a) := \frac{1}{N}\sum_{i \in s_a} \mathbf{1}[y_i \neq \hat{y}_i^{T}] + \lambda,$$} which could be obtained at equality if we achieve the accuracy of $T$ in a single leaf. We add that this is equivalent (by definition) to {$$lb_{\textrm{guess}}(s_a) = \frac{1}{N}|\MC \cap s_a| + \lambda.$$}

\item We additionally define {$t^*(s_a, d, \lambda)$} as an optimal solution for subproblem $s_a$, depth limit $d$, and regularizer $\lambda$ (that is, the solution found when we do not use lower bound guessing). When we just specify $t^*$ without arguments, we are referring to a solution for the root subproblem (the whole dataset), with the depth limit argument provided for the tree as a whole.

\item We also define {$R(s_a, d, \lambda)$} as the objective of the optimal solution found for subproblem $s_a$, depth limit $d$, and regularizer $\lambda$ (that is, the objective of the solution found when we do not use lower bound guessing):  {$$R(s_a, d, \lambda) := \frac{1}{N}\sum_{i\in s_a} \mathbf{1}[y_i \neq \hat{y}_i^{t^*(s_a, d, \lambda)}] + \lambda H_{t^* \cap s_a}.$$}

\item {Define $lb_\textrm{max}(s_a, d, \lambda)$ as the highest lower bound estimate that occurs for a given subproblem $s_a$, depth budget $d$, and regularization $\lambda$, across the algorithm's whole execution when using lower bound guessing (that is, the highest value of $lb_\textrm{current}$ from the steps outlined in \ref{alg:lbg}). Note that $$R_{\textrm{guess}}(s_a, d, \lambda) \leq lb_\textrm{max}(s_a, d, \lambda)$$ because when a subproblem is solved, the current lower bound is made to match the objective of the solution returned for that subproblem. As a reminder, $R_{\textrm{guess}}$ is computed after the subproblem is solved. When using lower bound guesses, it is possible for intermediate lower bound estimates (and therefore $lb_\textrm{max}$) to exceed $R_{\textrm{guess}}(s_a, d, \lambda)$, and then the lower bound (but not $lb_\textrm{max}$) is decreased to match the objective of the best solution found when the subproblem is solved.
}
\end{itemize}



Before beginning the proof, recall the steps described in Section \ref{sec:lbguess}, 
which apply regardless of the particular search strategy used.

\begin{proof} 
Without loss of generality, select a tree $t$ that is within some depth constraint $d$. We wish to prove that the {risk} on the full dataset (with some regularization $\lambda$) is bounded as: 
\begin{equation} \label{eq:lbguesstoprove}
R(t_{\textrm{guess}}, \tilde{\x}, \y) \leq \frac{1}{N} \left(|\MC| + \sum_{i \in \CC} \mathbf{1}[y_i \neq \hat{y}_i^{t}]\right) + \lambda H_{t}.\end{equation}
or, equivalently, (defining $s_\textrm{all}$ as the set of all points in the dataset):
$$R_\textrm{guess}(s_\textrm{all}, d, \lambda) \leq \frac{1}{N} \left(|\MC| + \sum_{i \in \CC} \mathbf{1}[y_i \neq \hat{y}_i^{t}]\right) + \lambda H_{t}.$$

To show this, it is sufficient to show a result that is strictly more general. Specifically, we show that for any subproblem $s_a$ (with $d, \lambda \geq 0$) that occurs as an internal or leaf node in $t$ (including the root), we can bound $lb_\textrm{max}(s_a, d, \lambda)$ as follows. Equation \eqref{eq:lbguesstoprove} is a direct consequence, using $s_a$ as the full dataset $s_{\textrm{all}}$.
\begin{equation} \label{eq:lbguesslambda}
\begin{aligned}
lb_\textrm{max}(s_a, d, \lambda) \leq  \frac{1}{N} \left(|\MC \cap s_a|+\sum_{i \in \CC \cap s_a} \mathbf{1}[y_i \neq \hat{y}_i^{t}]\right) 
 + \lambda H_{t_a}. 
\end{aligned}
\end{equation}
What we originally wished to prove then follows from noting that, by definition of $lb_\textrm{max}$, we have
$R_\textrm{guess}(s_{\textrm{all}}, d, \lambda) \leq lb_\textrm{max}(s_\textrm{all}, d, \lambda)$. 
Here, $d, \lambda$ are the depth limit and regularization provided for $t_\textrm{guess}$ (where $d$ matches or exceeds the depth of $t$).

We prove this sufficient claim (hereafter referred to as Equation \ref{eq:lbguesslambda}) for all subproblems in $t$ using induction.
\\\\
\textbf{Base Case}: Let us take any subset of data $s_a$, depth constraint $d$, and regularization $\lambda$ which corresponds to a subproblem in tree $t$ whose solution in $t$ is a leaf node.

Because the solution to $s_a$ was a leaf in $t$, then its objective (without making further splits) is $ub = \frac{1}{N}(\sum_{i \in s_a} \mathbf{1}[y_i \neq \hat{y}_i^{t}]) + \lambda\cdot 1$. 
We want to show, in this case, that  \eqref{eq:lbguesslambda} holds.

Our initial lower bound guess is $lb_\textrm{guess}$. Either $lb_\textrm{guess} > ub$, or $lb_\textrm{guess} \leq ub$. If $lb_\textrm{guess} > ub$, we are done with the subproblem as per Step \ref{alg:lbg_basecase} in the Branch-and-Bound algorithm. Otherwise, when $lb_\textrm{guess} \leq ub$, we know from Step \ref{alg:lbg_increase} that the lower bound can never increase above $ub$. Therefore, the highest value of the lower bound during execution,  $lb_\textrm{max}(s_a, d, \lambda)$, obeys  $lb_\textrm{max}(s_a, d, \lambda)\leq \max (ub, lb_{\textrm{guess}})$. Then,

\begin{equation*}
    \begin{aligned}
    lb_\textrm{max}(s_a, d, \lambda) &\leq \max (ub, lb_{\textrm{guess}}) \;\;\;\;\textrm{(by argument just above)}\\
    &\leq \max \left(\frac{1}{N}\sum_{i \in s_a} (\mathbf{1}[y_i \neq \hat{y}_i^{t}]) + \lambda, \frac{1}{N}|\MC \cap s_a| + \lambda \right) \;\;\;\;\textrm{(by definition of $ub$ and $lb_\textrm{guess}$)}\\
    &\leq \frac{1}{N} \max \left(\sum_{i \in s_a} \mathbf{1}[y_i \neq \hat{y}_i^{t}],  |\MC \cap s_a|\right) + \lambda\\
    &\leq \frac{1}{N} \max \left(\sum_{i \in \CC \cap s_a} \mathbf{1}[y_i \neq \hat{y}_i^{t}]  + \sum_{i \in \MC \cap s_a} \mathbf{1}[y_i \neq \hat{y}_i^{t}], |\MC \cap s_a|\right) + \lambda.\\
\end{aligned}
\end{equation*}
Note that both terms inside the max are at most $\sum_{i \in \CC \cap s_a}\mathbf{1}[y_i \neq \hat{y}_i^{t}] + |\MC \cap s_a|$. Therefore, 
\begin{align*} 
lb_\textrm{max}(s_a, d, \lambda) \leq \frac{1}{N} \left( \sum_{i \in \CC \cap s_a}\mathbf{1}[y_i \neq \hat{y}_i^{t}] + |\MC \cap s_a|\right)  + \lambda.\end{align*}
Moreover, the number of leaves in the optimal tree for subproblem $s_a$ is 1, i.e., $H_{t_a} = 1$ (since $s_a$ corresponds to a leaf in $t$), so 
\begin{equation*}
    \begin{aligned}
    lb_\textrm{max}(s_a, d, \lambda) \leq  \frac{1}{N} \left(|\MC \cap s_a| +\sum_{i \in \CC \cap s_a} \mathbf{1}[y_i \neq \hat{y}_i^{t}]\right) + \lambda H_{t_a}. 
    \end{aligned}
\end{equation*}
And this matches Equation \eqref{eq:lbguesslambda}, as required. Thus, we have shown that the base case obeys the statement of the theorem.
\\\\
\textbf{Inductive Step}: Let the set of points $s_a$, depth constraint $d >0$, and regularization $\lambda \geq 0$ be a subproblem that corresponds to an internal node in $t$. Let $j$ indicate the feature that was split on in $t$ for this node, and define $s_j$ as the set of data points $\tilde{\x}_i$ such that $\tilde{x}_{ij} = 1$ and $s_j^c$ as the set of data points $\tilde{\x}_i$ such that $\tilde{x}_{ij} = 0$. We assume \eqref{eq:lbguesslambda} holds for both the left and right child subproblems and aim to show that it holds for their parent subproblem. The left subproblem is $s_a \cap s_j$ with depth $d-1$ and the right subproblem is $s_a \cap s_j^c$ with depth $d-1$. Thus, assuming (as per the inductive hypothesis) that \eqref{eq:lbguesslambda} holds, i.e., 
\begin{equation*}
    \begin{aligned}
    lb_\textrm{max}(s_a \cap s_j, d-1, \lambda) \leq & \frac{1}{N} \left(|\MC \cap s_a \cap s_j| + \sum_{i \in \CC \cap s_a \cap s_j} \mathbf{1}[y_i \neq \hat{y}_i^{t}]\right) + \lambda H_{t_a \cap s_j}
    \end{aligned}
\end{equation*}
and
\begin{equation*}
    \begin{aligned}
    lb_\textrm{max}(s_a \cap s_j^c, d-1, \lambda) \leq \frac{1}{N} \left(|\MC \cap s_a \cap s_j^c| + \sum_{i \in \CC \cap s_a \cap s_j^c} \mathbf{1}[y_i \neq \hat{y}_i^{t}]\right) + \lambda H_{t_{a \cap j^c}},
    \end{aligned}
\end{equation*}
it remains to show that Equation \eqref{eq:lbguesslambda} holds for $s_a$:
\begin{equation*}
    \begin{aligned}
    lb_\textrm{max}(s_a, d, \lambda) \leq & \frac{1}{N} \left(|\MC \cap s_a|   + \sum_{i \in \CC \cap s_a} \mathbf{1}[y_i \neq \hat{y}_i^{t}]\right) + \lambda H_{t_a}.
    \end{aligned}
\end{equation*}

We prove the inductive step by cases:
\begin{enumerate}
    \item If $ub \leq lb_\textrm{guess} + \lambda$, then 
    as per Step \ref{alg:lbg_basecase} in our branch-and-bound algorithm,
    $s_a$ corresponds to a leaf in $t_\textrm{guess}$, with a loss for this subproblem equal to $ub$. Since the algorithm returns immediately after changing the lower bound to $ub$, the maximum value the lower bound takes (that is, $lb_\textrm{max}$) is whichever of $lb_\textrm{guess}$ or $ub$ is higher. We have: 
    \begin{equation*}
        \begin{aligned}
        lb_\textrm{max}(s_a, d, \lambda) &\leq \max(lb_\textrm{guess}, ub) \\
        &\leq \max(lb_\textrm{guess}, lb_\textrm{guess} + \lambda) \; \; \; \; \text{(Since we conditioned on $ub \leq lb_\textrm{guess} + \lambda$)} \\
         &\leq lb_\textrm{guess} + \lambda \\
         &\leq \frac{1}{N}|\MC \cap s_a| + 2 \lambda \\& \leq \frac{1}{N}\left(|\MC \cap s_a| + \sum_{i \in \CC \cap s_a} \mathbf{1}[y_i \neq \hat{y}_i^{t}]\right) + 2 \lambda.
        \end{aligned}
    \end{equation*}
    Noting that because $s_a$ corresponds to an internal node in $t$, there are at least two leaves below it, so $H_{t_a} \geq 2$. Thus,
    \begin{equation*}
        \begin{aligned}
            lb_\textrm{max}(s_a, d, \lambda) \leq  \frac{1}{N}\left(|\MC \cap s_a| + \sum_{i \in \CC \cap s_a} \mathbf{1}[y_i \neq \hat{y}_i^{t}]\right) + \lambda H_{t_a}.
        \end{aligned}
    \end{equation*}
    This equation matches Equation \ref{eq:lbguesslambda}, as required.
    
    
    \item Else, as per Step \ref{alg:lbg_increase}, 
    the lower bound (and therefore $lb_\textrm{max}(s_a, d, \lambda)$) cannot exceed the combined lower bounds of the left and right subproblems from splitting on feature $j$. We know the split for feature $j$ will lead to a lower bound estimate no more than  $lb_\textrm{max}(s_a \cap s_j, d-1, \lambda) + lb_\textrm{max}(s_a \cap s^c_j, d-1, \lambda)$. Thus we have: 
    \begin{equation*}
            lb_\textrm{max}(s_a, d, \lambda) \leq  lb_\textrm{max}(s_a \cap s_j, d-1, \lambda) + lb_\textrm{max}(s_a \cap s_j^c, d-1, \lambda).
    \end{equation*}
    Using the inductive hypothesis, this reduces to 
    \begin{align*}
    lb_\textrm{max}(s_a, d, \lambda) &\leq  \frac{1}{N} \left(|\MC \cap s_a \cap s_j| + \sum_{i \in \CC \cap s_a \cap s_j} \mathbf{1}[y_i \neq \hat{y}_i^{t}]\right) + \lambda H_{t_{a \cap j}} \\ & \ \ \ \ \ + \frac{1}{N} \left(|\MC \cap s_a \cap s_j^c| + \sum_{i \in \CC \cap s_a \cap s_j^c} \mathbf{1}[y_i \neq \hat{y}_i^{t}]\right) +  \lambda H_{t_{a \cap j^c}}.
    \end{align*}
Noting that $s_a \cap s_j$ and $s_a \cap s_j^c$ partition $s_a$:
    \begin{align*}
    lb_\textrm{max}(s_a, d, \lambda) &\leq  \frac{1}{N} \left(|\MC \cap s_a| + \sum_{i \in \CC \cap s_a} \mathbf{1}[y_i \neq \hat{y}_i^{t}]\right) + \lambda H_{t_a}.
    \end{align*}
    This equation matches Equation \eqref{eq:lbguesslambda}, as required. (Although we also need to consider the case where $lb_\textrm{guess}$ never increases by the mechanism in \ref{alg:lbg_increase}, in this case it never increases above the initial $lb_\textrm{guess}$, and we know from the argument for case 1 that if $lb_\textrm{max}(s_a, d, \lambda) \leq lb_\textrm{guess}$ then $lb_\textrm{max}(s_a, d, \lambda)$ still satisfies equation \eqref{eq:lbguesslambda}.)
\end{enumerate}

Thus we have proved the inductive step holds for all possible cases. 

By induction, then, we have proved Equation \eqref{eq:lbguesslambda} holds for any internal or leaf node in $t$. 

Since the root node is an internal node of $t$, we have also proven Equation \eqref{eq:lbguesslambda} holds for the root problem. As per the justification given when claiming Equation \eqref{eq:lbguesslambda} was sufficient, that also means
$$R(t_{\textrm{guess}}, \tilde{\x}, \y) \leq \frac{1}{N} \left(|\MC| + \sum_{i \in \CC} \mathbf{1}[y_i \neq \hat{y}_i^{t}]\right) + \lambda H_{t}$$
which is what we wished to show.
\end{proof}

\subsection{Proof of Corollary \ref{thm:glb-born-again}}
\textbf{Corollary }\ref{thm:glb-born-again}
\textit{
 Let $T$, $T'$ and $t'$ be defined as in Theorem \ref{thm:threshold_guess}. 
Let $t_\textrm{guess}$ be the tree obtained using lower-bound guessing with $T'$ as the reference model, on $\tilde{\x}_T^{\textrm{reduced}}, \y$, with depth constraint matching or exceeding the depth of $t'$. Then 
$\mathcal{L}(t_\textrm{guess}, \tilde{\x}_T^{\textrm{reduced}}, \y) - \mathcal{L}(T, \x, \y) \leq \lambda (H_{t'} - H_{t_\textrm{guess}})$, or
equivalently, 
${R}(t_\textrm{guess}, \tilde{\x}_T^{\textrm{reduced}}, \y) \leq \mathcal{L}(T, \x, \y) + \lambda H_{t'}$.
}

\begin{proof}

Using Theorem \ref{thm:glb}, we know that for any tree $t$ within the depth constraint for our problem, and any reference model $T$ (this being a local variable not to be confused with $T$ as defined above), we have:
\begin{align*}R(t_{\textrm{guess}}, \tilde{\x}_T^\textrm{reduced}, \y) \leq &\frac{1}{N} \left(|\MC| + \sum_{i \in \CC} \mathbf{1}[y_i \neq \hat{y}_i^{t}]\right) + \lambda H_{t}.\end{align*}
Picking $t = t'$ (noting that $t'$ is, as required, within the depth constraint for our problem), and picking the reference model $T'$, we have:
\begin{align*}R(t_{\textrm{guess}}, \tilde{\x}_T^\textrm{reduced}, \y) \leq &\frac{1}{N} \left(|\MCp| + \sum_{i \in \CCp} \mathbf{1}[y_i \neq \hat{y}_i^{t'}]\right) + \lambda H_{t'}.\end{align*}
Since $t'$, by definition, makes the same classifications as $T'$, $\sum_{i \in \CCp} \mathbf{1}[y_i \neq \hat{y}_i^{t'}] = 0$. 
\begin{align*}R(t_{\textrm{guess}}, \tilde{\x}_T^\textrm{reduced}, \y) \leq &\frac{1}{N} \left(|\MCp|\right) + \lambda H_{t'}.\end{align*}
Recall by definition $\MCp = \{i |y_i \neq \hat{y}_i^{T'}\}$, 
\begin{align*}R(t_{\textrm{guess}}, \tilde{\x}_T^\textrm{reduced}, \y) \leq &\frac{1}{N} \left(\sum_{i=1}^N \mathbf{1}[y_i \neq \hat{y}_i^{T'}]\right) + \lambda H_{t'}.\end{align*}
Noting the definition of $\mathcal{L}(T', \tilde{\x}_T^\textrm{reduced}, \y)$
\begin{align*}R(t_{\textrm{guess}}, \tilde{\x}_T^\textrm{reduced}, \y) \leq \mathcal{L}(T', \tilde{\x}_T^\textrm{reduced}, \y) + \lambda H_{t'}.\end{align*}
Noting the definition of $R(t_\textrm{guess}, \tilde{\x}_T^\textrm{reduced}, \y)$
\begin{align*}
\mathcal{L}(t_{\textrm{guess}}, \tilde{\x}_T^\textrm{reduced}, \y) + \lambda H_{t_\textrm{guess}}\leq \mathcal{L}(T', \tilde{\x}_T^\textrm{reduced}, \y) + \lambda H_{t'}.\end{align*}
As in the proof provided in Appendix \ref{app:proof-4.1}, we note that when we are building $\tilde{\x}_T^{\textrm{reduced}}$ (the dataset used to construct $T'$), the stopping condition of column elimination is that accuracy decreases, which is when the loss increases: $\mathcal{L}(T, \x, \y) < \mathcal{L}(T', \tilde{\x}_T^{\textrm{reduced}}, \y)$. Thus, the opposite must always be true as we build $T'$: $\mathcal{L}(T', \tilde{\x}_T^{\textrm{reduced}}, \y) \leq \mathcal{L}(T, \x, \y)$.
\begin{align*}
\mathcal{L}(t_{\textrm{guess}}, \tilde{\x}_T^\textrm{reduced}, \y) + \lambda H_{t_\textrm{guess}}\leq \mathcal{L}(T, \x, \y) + \lambda H_{t'}.\end{align*}
Then, $\mathcal{L}(t_{\textrm{guess}}, \tilde{\x}_T^\textrm{reduced}, \y) - \mathcal{L}(T, \x, \y)  \leq \lambda (H_{t'} - \lambda H_{t_\textrm{guess}}).$



\end{proof}

\section{Experimental Details}
\label{app:experiment-setup}
In this section, we elaborate on the datasets used in our evaluation, preprocessing on those datasets, and the experimental setup. 

\subsection{Datasets}
We present results for seven datasets: one simulated 2-dimensional two spirals dataset, the Fair Isaac (FICO) credit risk dataset \citep{competition} used for the Explainable ML Challenge, three recidivism datasets (COMPAS, \citealp{LarsonMaKiAn16}, Broward, \citealp{wang2020pursuit}, Netherlands, \citealt{tollenaar2013method}), and two coupon datasets (Takeaway and Restaurant), which were collected on Amazon Mechanical Turk via a survey \citep{wang2015or}. We predict whether an individual will default on a loan for the FICO dataset, which individuals are arrested within two years of release on the Propublica recidivism dataset, whether defendants have any type of charge (that they were eventually convicted for) within two years from the current charge date/release date on the Broward recidivism dataset, whether defendants have any type of charge within four years on the Netherlands dataset, and whether a customer will accept a coupon for takeaway food or a cheap restaurant depending on their coupon usage history, current conditions while driving, and coupon expiration time on two coupon datasets. Table \ref{tab:data} summarizes all the datasets. 
All the datasets, except spiral, are publicly available, without license. The spiral dataset is a synthetic data set we created; the data are published with our benchmarking infrastructure (see Appendix~\ref{app:software}).

\subsection{Preprocessing}

\noindent\textbf{Spiral} and \textbf{FICO}: We use these two datasets directly, without preprocessing.

\noindent\textbf{COMPAS}: We selected features \textit{sex, age, juv\_fel\_count, juv\_misd\_count, juv\_other\_count, priors\_count, and c\_charge\_degree} and the label \textit{two\_year\_recid}. 

\noindent\textbf{Broward}: We selected features \textit{sex, age\_at\_current\_charge, age\_at\_first\_charge,
p\_charges, p\_incarceration, p\_probation, p\_juv\_fel\_count,p\_felprop\_viol, p\_murder, p\_felassault, p\_misdeassault, p\_famviol, p\_sex\_offense, p\_weapon, p\_fta\_two\_year, p\_fta\_two\_year\_plus, current\_violence, current\_violence20,p\_pending\_charge, p\_felony, p\_misdemeanor, p\_violence, total\_convictions,p\_arrest, p\_property, p\_traffic, p\_drug,p\_dui, p\_domestic, p\_stalking, p\_voyeurism, p\_fraud, p\_stealing, p\_trespass, six\_month, one\_year, three\_year, and five\_year} and the label \textit{general\_two\_year}. 

\noindent\textbf{Netherlands}: We translated the feature names from Dutch to English and then selected features \textit{sex, country of birth, log \# of previous penal cases, age in years, age at first penal case, offence type, 11-20 previous case, $>$20 previous case, and age squared} and the label recidivism\_in\_4y. 

\noindent\textbf{Takeaway} and \textbf{Restaurant}: We selected features \textit{destination, passanger, weather, temperature, time, expiration, gender, age, maritalStatus, Childrennumber, education, occupation, income, Bar, CoffeeHouse, CarryAway,RestaurantLessThan20, Restaurant20To50, toCoupon\_GEQ15min, toCoupon\_GEQ25min, direction\_same} and the label \textit{Y}, and removed observations with missing values. We then used one-hot encoding to transform these categorical features into binary features.

\begin{table}[ht]
    \small
    \centering
    \begin{tabularx}{\textwidth}{|X|c|c|c|X|}\hline
    Dataset & samples & features & binary features & classification problem\\\hline
    Spiral & 100 & 2 & 180 & whether a point belongs to the first spiral\\\hline
    COMPAS~\citep{LarsonMaKiAn16} & 6907 & 7 & 134 & whether individuals are arrested within two years of release \\\hline
    Broward~\citep{wang2020pursuit} & 1954 & 38 & 588 & whether defendants have any type of charge within two years from the current charge date/release data \\\hline
    Netherlands~\citep{tollenaar2013method} & 20000 & 9 & 53890 & whether defendants have any type of charge within four years\\\hline
    FICO~\citep{competition} & 10459 & 23 & 1917 & whether an individual will default on a loan\\\hline
    Takeaway food~\citep{wang2015or} & 2280 & 21 & 87 & whether a customer will accept a coupon\\ \hline
    Cheap Restaurant~\citep{wang2015or} & 2653 & 21 & 87 & whether a customer will accept a coupon\\ \hline
    \end{tabularx}
    \caption{Datasets}
    \label{tab:data}
\end{table}

\subsection{Evaluation Platform}

Unless otherwise noted, reported times are from a 32-core dual Intel E5-2683 v4 Broadwell processor running at 2.1 Ghz, with approximately 125 GB of available memory. We ran all tests single-threaded (i.e., we used only one of the 32 cores) on the Cedar cluster of Compute Canada.

\subsection{Software Packages Used and/or Modified}\label{app:software}
\noindent\textbf{GOSDT}: 
We implemented all three types of guesses in the implementation of GOSDT, using the publicly released code from~\citet{lin2020generalized}
(https://github.com/Jimmy-Lin/GeneralizedOptimalSparseDecisionTrees). That code bears no license; we will release our modified version with a 3-clause BSD license.

\noindent\textbf{DL8.5}
We also implemented guessing thresholds and lower bounds in the publicly available code from ~\citet{aglin2020learning} (https://github.com/aia-uclouvain/pydl8.5). That code bears an MIT license; we will release our modified version under the same license.

\noindent\textbf{CART}: We run CART using the Python implementation from Sci-Kit Learn. The depth and the minimum number of samples required per leaf node are constrained to adjust the resulting tree size.

\noindent\textbf{Born-again Tree Ensembles (batree)}: We run batree using the publicly available code from \citet{vidal2020born} (https://github.com/vidalt/BA-Trees). 

We based our benchmarking infrastructure off the tree-benchmark suite of Lin (https://github.com/Jimmy-Lin/TreeBenchmark). That code is currently unlicensed; we will release our modified tree-benchmark with a 3-clause BSD license.

\section{More Experimental Results}\label{app:more-experiments}
We elaborate on hyperparameters used for reference models and optimal decision tree algorithms. We then present extra experimental results that are omitted from the main paper due to space constraints. 

\textbf{Collection and Setup}: we ran the following experiment on the 7 datasets \textbf{COMPAS}, \textbf{Broward}, \textbf{Netherlands}, \textbf{FICO}, \textbf{Takeaway}, \textbf{Restaurant}, and \textbf{spiral}. For each dataset, we trained decision trees using varying configurations. For each configuration, we performed a 5-fold cross validation to measure training time, training accuracy and test accuracy for each fold (except the spiral dataset, which is not used in the tables). 

Table \ref{tab:tree_config} lists the configurations used for each dataset when training decision trees. For GOSDT, DL8.5, and CART, we set the depth limit from 2 to 5. For GOSDT, since it is possible, we also include the configuration with no depth constraint. For datasets except spiral, we set GOSDT regularizers to be 0.005, 0.002, 0.001, 0.0005, 0.0002, and 0.0001. If $\frac{1}{\textrm{sample size}} < \textrm{regularizer}$, we omit the regularizer (except for Broward and Takeaway, where we still include the largest regularization below this threshold, 0.0005). For spiral, we set GOSDT regularizers to be 0.1, 0.05, 0.02, and 0.01. DL8.5 and CART do not have a per-leaf regularization penalty but instead allow specification of the minimum support for each leaf. Hence, we set the minimum support of each leaf in these two algorithms to be  $\lceil \text{sample size} \times \text{regularizer} \rceil$. For batree, we used the default setting. 

For threshold guessing and lower bound guessing, we used a gradient boosted decision tree (GBDT) as the reference model. We trained GBDT using four different configurations: 40 decision stumps, 30 max-depth 2 weak classifiers, 50 max-depth 2 weak classifiers, and 20 max-depth 3 weak classifiers. There are no time or memory limitations imposed on configurations to train the reference models. We selected these hyperparameter pairs since they often generate GBDTs with good accuracy, and different reference models from these hyperparamers result in a different number of thresholds. This allows us to analyze how different reference models influence the performance of trees built upon them (see Appendix \ref{app:thresholds}, \ref{app:lowerbounds}, and \ref{app:bad-guess}). In practice, 
a user could tune the hyperparameters of their black box modeling algorithm to favor a specific tradeoff between the number of thresholds and the training performance (the former of these influences training-time performance of our methods). 
For that reason, if several hyperparameter pairs lead to similar performance, it could benefit computation time to use the hyperparameter that introduce the fewest thresholds. We fix the learning rate to 0.1. 

\begin{table*}
    \centering
    \begin{tabular}{|c|c|l|}\hline
    Dataset & Depth Limit & Regularizer ($\lambda$)\\\hline
    COMPAS & 2,3,4,5 ($\infty$ for GOSDT only)& 0.005, 0.002, 0.001, 0.0005, 0.0002\\\hline
    Broward & 2,3,4,5 ($\infty$ for GOSDT only)& 0.005, 0.002, 0.001, 0.0005 \\\hline
    Netherlands & 2,3,4,5 ($\infty$ for GOSDT only)& 0.005, 0.002, 0.001, 0.0005, 0.0002, 0.0001 \\\hline
    FICO & 2,3,4,5 ($\infty$ for GOSDT only)& 0.005, 0.002, 0.001, 0.0005, 0.0002\\\hline
    Takeaway & 2,3,4,5 ($\infty$ for GOSDT only)& 0.005, 0.002, 0.001, 0.0005 \\\hline
    Restaurant & 2,3,4,5 ($\infty$ for GOSDT only)& 0.005, 0.002, 0.001, 0.0005\\\hline
    Spiral & 2,3,4,5 ($\infty$ for GOSDT only) & 0.1, 0.05, 0.02, 0.01 \\ \hline 
    \end{tabular}
    \caption{Configurations used to train decision tree models}
    \label{tab:tree_config}
\end{table*}

\subsection{Train-time Savings from Guessing}
\label{app:overall-benefit}

\noindent\textbf{Calculations}: For each combination of dataset, algorithm, and configuration, we produce a set of up to 5 models, depending on how many runs exceeded the time limit or memory limit. We summarize the measurements (e.g., training time, training accuracy, and test accuracy) across the set of up to five models by plotting the median. We compute the $25^{th}$ and $75^{th}$ percentile and show them as lower and upper error values respectively. For the spiral dataset, since the training set is the whole dataset, we show the measurements without error bars. 

\begin{figure}
    \centering
    \begin{subfigure}[t]{0.48\textwidth}
         \centering
         \includegraphics[width=\textwidth]{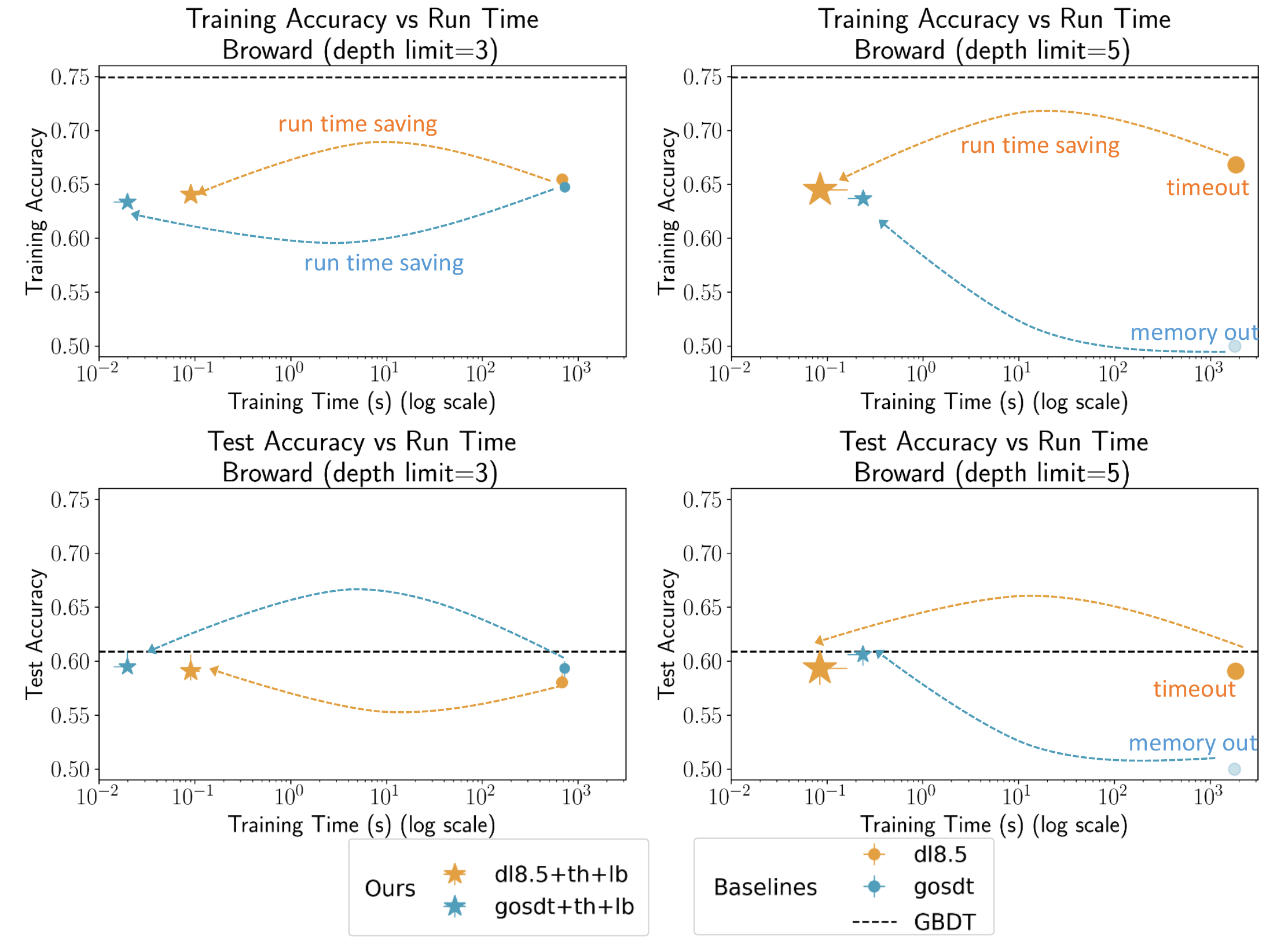}
         \caption{(\textbf{Train-time savings from guessing on the Broward dataset.}) The reference model was trained using 40 decision stumps. $\lambda=0.005$.}
         \label{fig:acc_vs_runtime_broward}
     \end{subfigure}
     \hfill
     \begin{subfigure}[t]{0.48\textwidth}
         \centering
         \includegraphics[width=\textwidth]{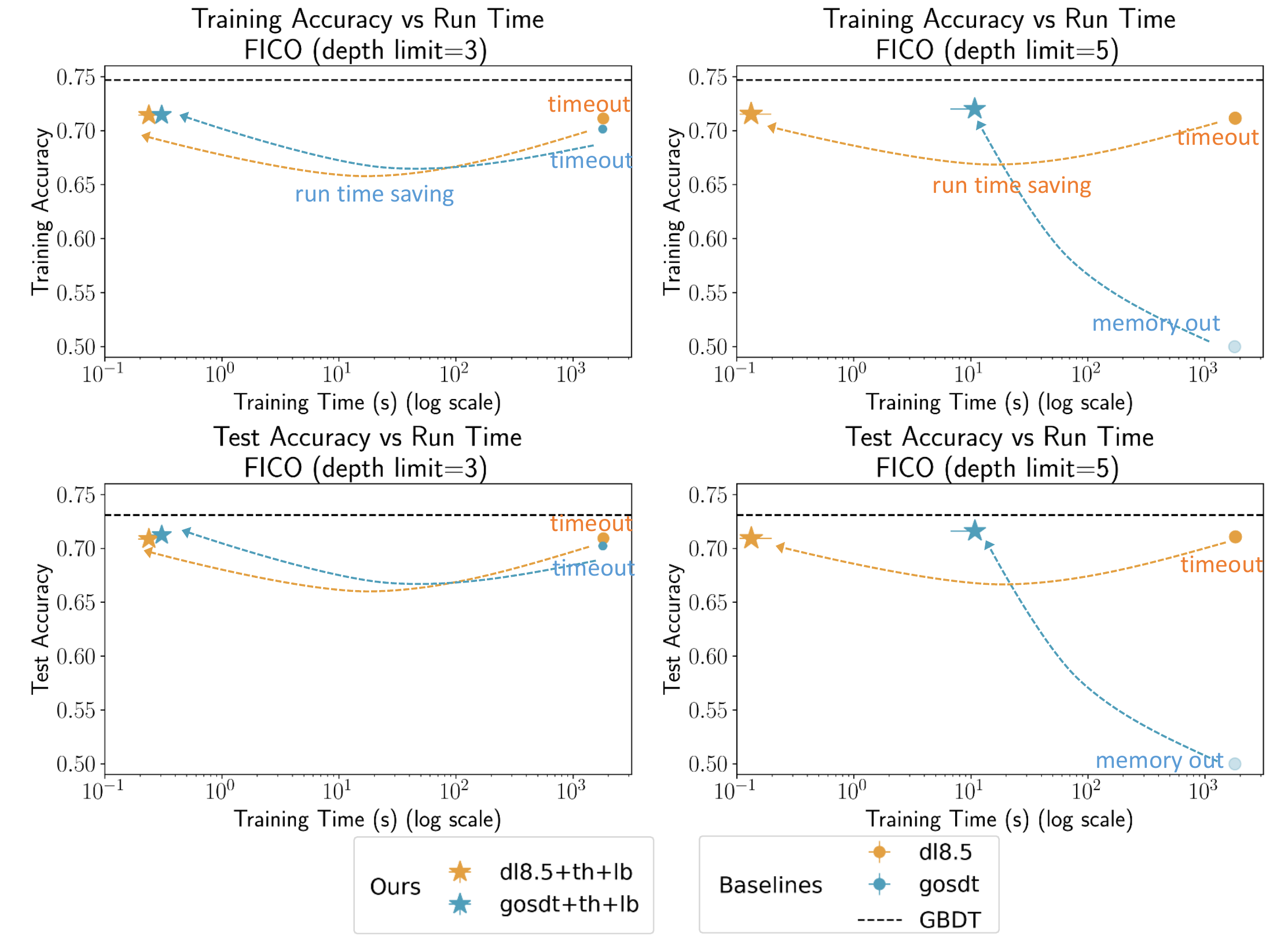}
         \caption{(\textbf{Train-time savings from guessing on the FICO dataset.}) The reference model was trained using 40 decision stumps. $\lambda=0.0005$.}
         \label{fig:acc_vs_runtime_fico}
     \end{subfigure}
     \vskip\baselineskip
    \begin{subfigure}[t]{0.48\textwidth}
         \centering
         \includegraphics[width=\textwidth]{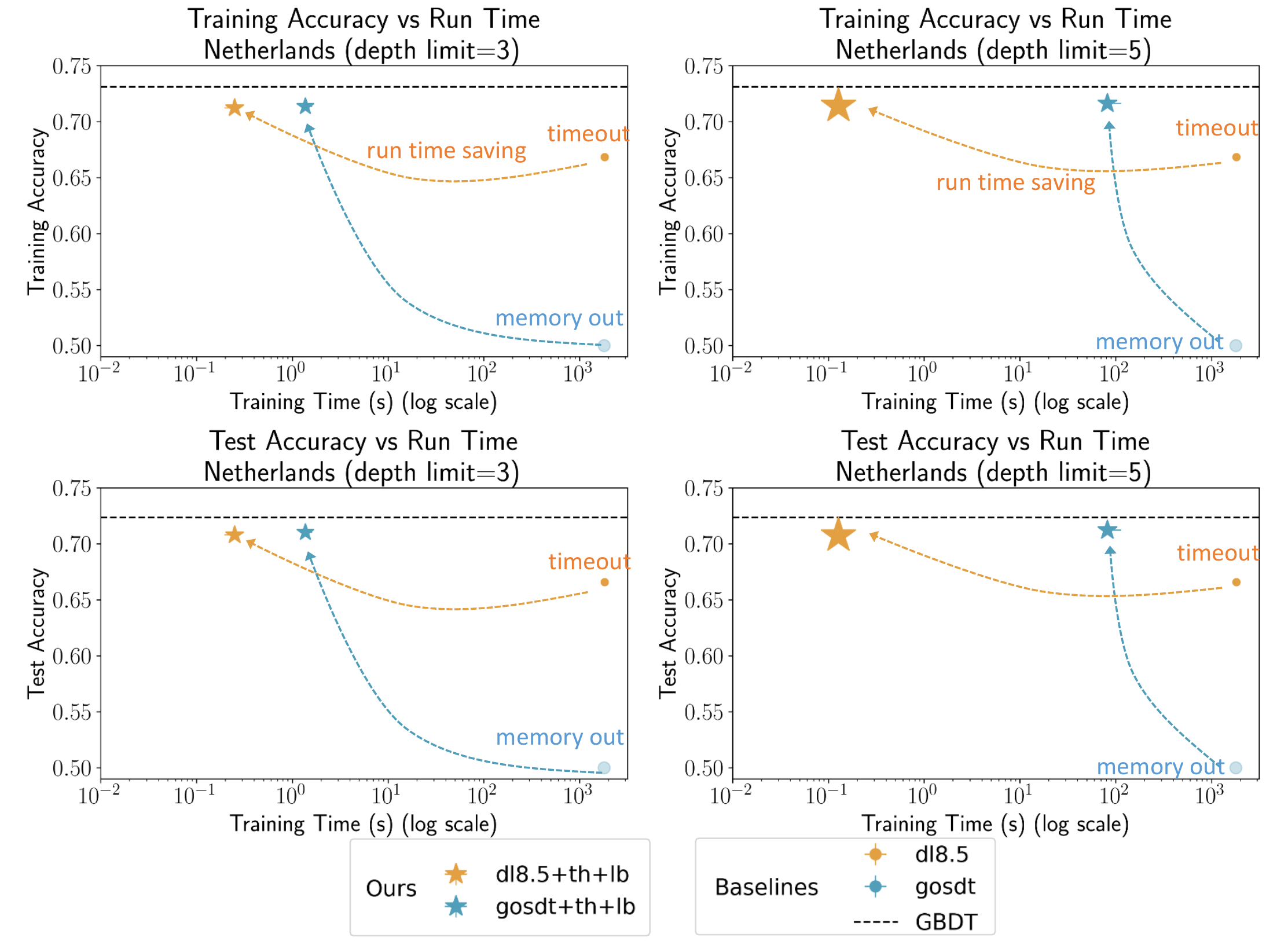}
         \caption{(\textbf{Train-time savings from guessing on the Netherlands dataset.}) The reference model was trained using 30 max-depth 2 weak classifiers. $\lambda=0.001$.}
         \label{fig:acc_vs_runtime_netherlands}
     \end{subfigure}
     \hfill
     \begin{subfigure}[t]{0.48\textwidth}
         \centering
         \includegraphics[width=\textwidth]{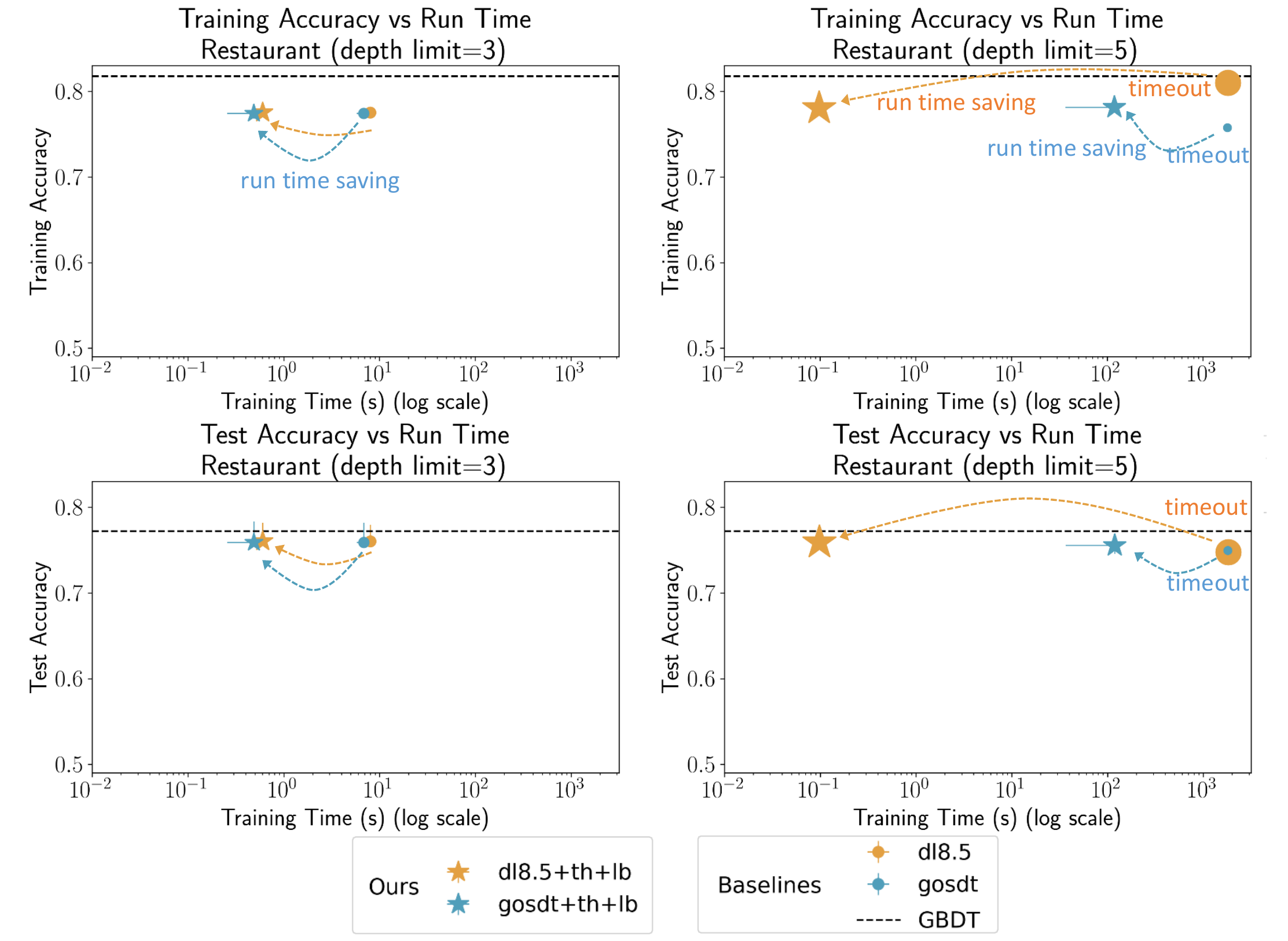}
         \caption{(\textbf{Train-time savings from guessing on the Restaurant dataset.}) The reference model was trained using 50 max-depth 2 weak classifiers. $\lambda=0.001$.}
         \label{fig:acc_vs_runtime_restaurant}
     \end{subfigure}
    \caption{Training accuracy versus run time for GOSDT (blue) and DL8.5 (gold) with and without guessing strategies. The black line shows the training accuracy of a GBDT model (with 100 max-depth3 estimators).}
    \label{fig:acc_vs_runtime}
\end{figure}

\noindent\textbf{Results}: Figure \ref{fig:acc_vs_runtime} shows training and test results on the Netherlands, Broward, FICO, and Restaurant datasets. The difference between the stars and the circles shows that guessing strategies improve both DL8.5 and GOSDT training times, generally by 1-2 orders of magnitude. Moreover, with respect to test accuracy, models trained using guessing strategies have test accuracy that is close to or sometimes even better than that of the black box model. 

\subsection{Sparsity versus Accuracy}
\label{app:acc-v-sparsity}
\noindent\textbf{Calculation}: We use the same calculations as those mentioned in Appendix \ref{app:overall-benefit}. The only difference is that we discarded runs that exceed the 30-minute time limit or 125 GB memory limit. 

\begin{figure}
    \centering
    \begin{subfigure}[t]{0.48\textwidth}
         \centering
         \includegraphics[width=\textwidth]{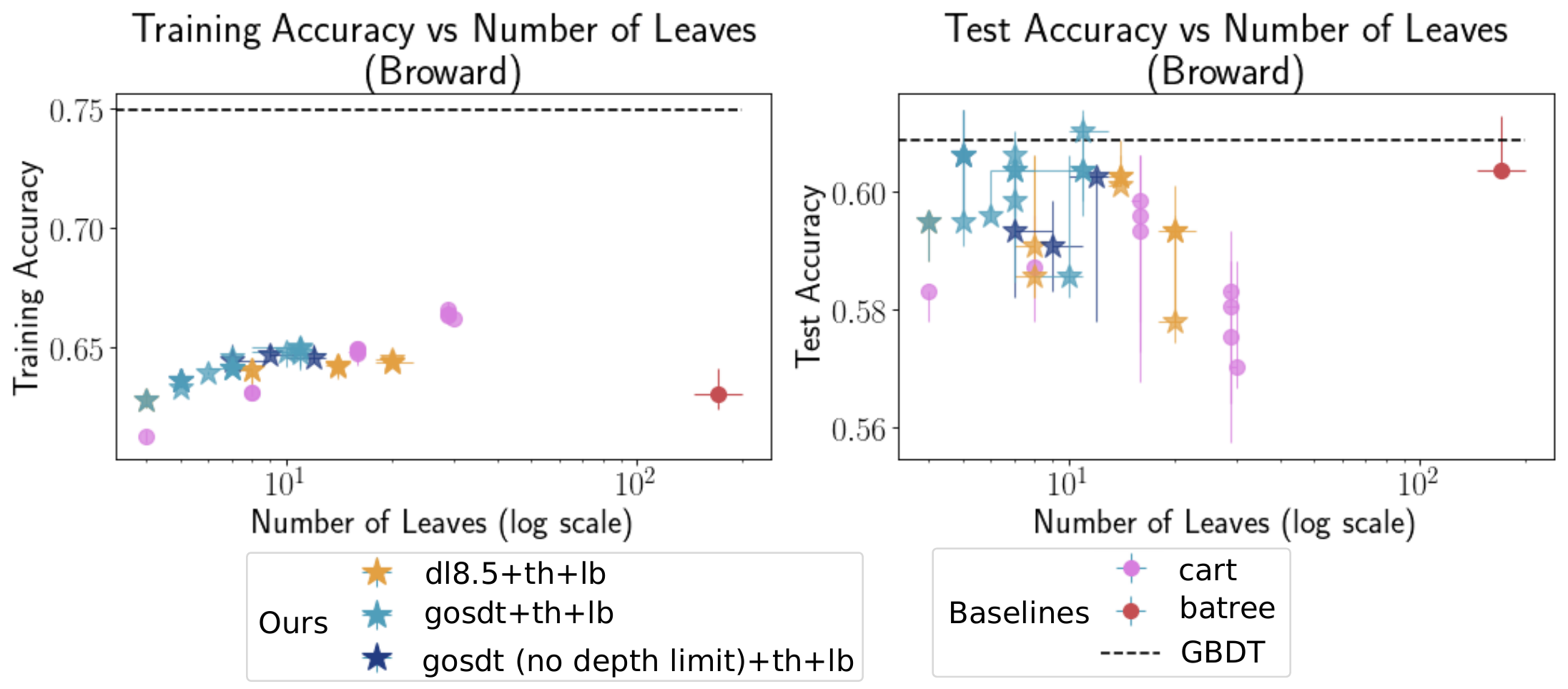}
         \caption{(\textbf{Sparsity vs. accuracy on the Broward dataset.}) The reference model was trained using 40 decision stumps.}
         \label{fig:acc_vs_sparsity_broward}
     \end{subfigure}
     \hfill
     \begin{subfigure}[t]{0.48\textwidth}
         \centering
         \includegraphics[width=\textwidth]{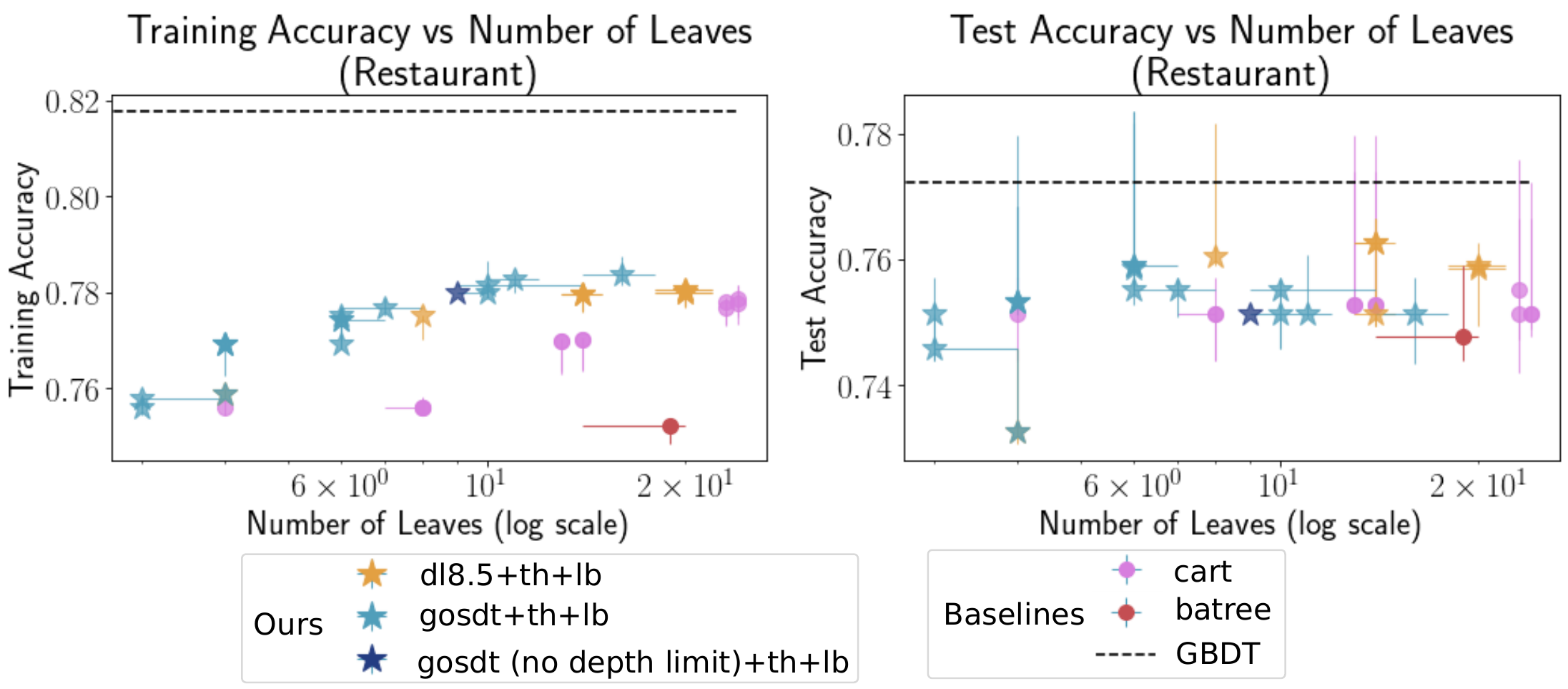}
         \caption{(\textbf{Sparsity vs. accuracy on the Restaurant dataset.}) The reference model was trained using 50 max-depth 2 weak classifiers.}
         \label{fig:acc_vs_sparsity_restaurant}
     \end{subfigure}
    \caption{DL8.5 and GOSDT use guessed thresholds and guessed lower bounds. CART was trained on the original dataset with no guess. Batree was trained using a random forest reference model with 10 max-depth 3 weak classifiers. The black line shows the training accuracy of a GBDT model (with 100 max-depth 3 estimators).}
    \label{fig:acc_vs_sparsity}
\end{figure}

\noindent\textbf{Results}: Figure \ref{fig:acc_vs_sparsity} shows the \textbf{accuracy-sparsity tradeoff} for different decision tree models (with GBDT accuracy indicated by the black line). Our guessed trees are both sparse and accurate, defining a frontier. That means we achieve the highest accuracy for almost every level of sparsity. CART trees sometimes have higher accuracy with more leaves on the training set. But these complicated trees are overfitting, leading to lower test accuracy. Batree usually does not achieve results on the frontier.

\subsection{Value of Threshold Guessing}
\label{app:thresholds}
\noindent\textbf{Calculations}: In this experiment, we summarize the measurements across the set of up to 5 models by plotting the average. If one of 5 folds runs out of time or memory, we mark the result as timeout or memory out on the plot. If the experiment timed out, we analyze the best model found at the time of the timeout; if we run out of memory, we are unable to obtain a model and assign 0.5 to accuracy for visualization purposes. In Tables \ref{tab:threshold_guess_gosdt} and \ref{tab:threshold_guess_dl8.5}, we report the mean and standard deviation. 

\begin{figure}
    \centering
    \includegraphics[scale=0.38]{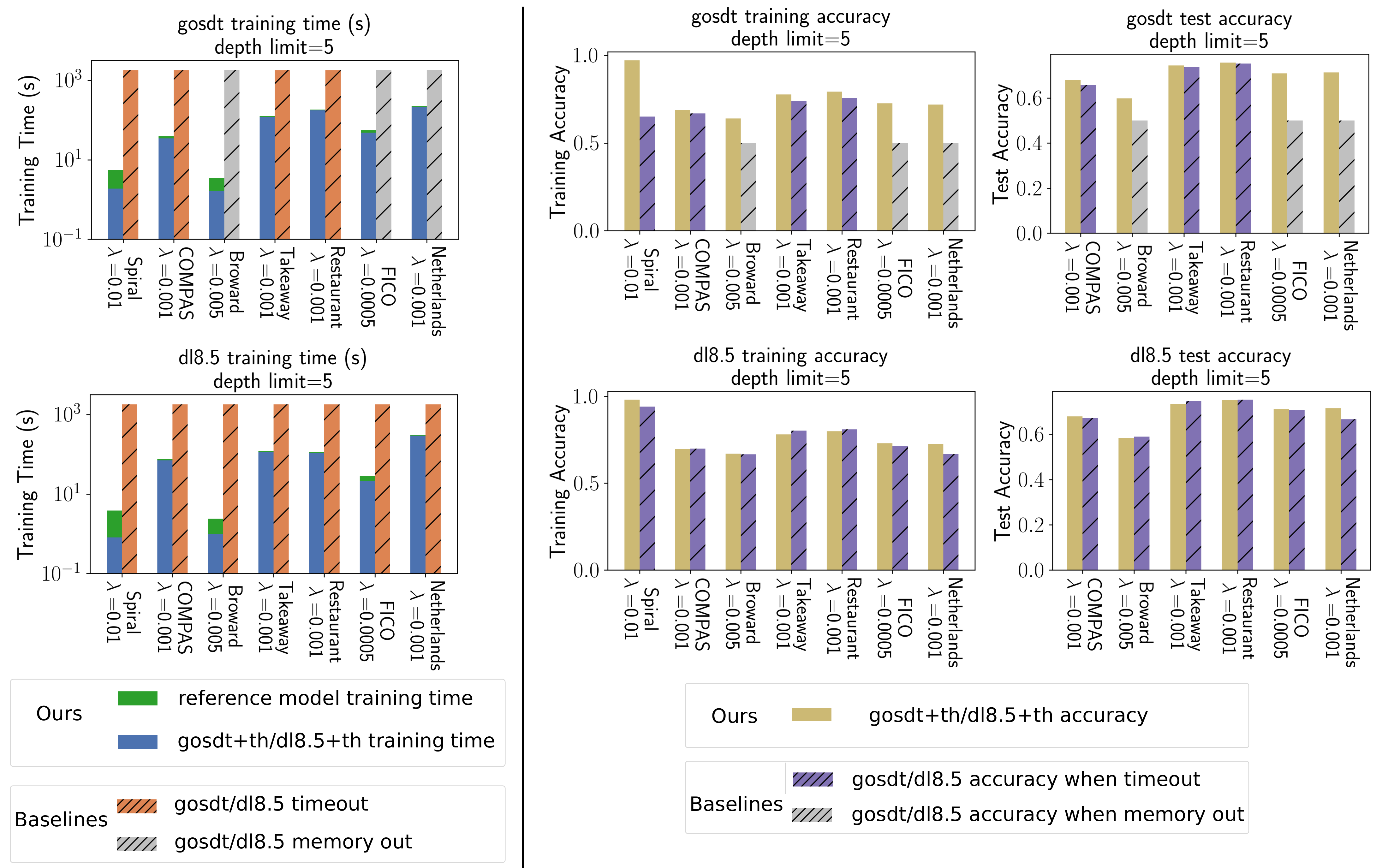}
    \caption{(\textbf{Value of threshold guessing.}) Training time (logscale), training accuracy, and test accuracy for GOSDT and DL8.5 with depth limit 5. Two subfigures on the left compare the training time with (blue) and without (orange) threshold guessing. The reference model training time (in green) is often very small and appears atop the training time with guessing (blue). All baselines (hatched) timed out (orange) or hit the memory limit (grey). Subfigures on the right show accuracy with (gold) and without (purple) threshold guessing. When the algorithm ran out of memory, 0.5 was assigned to accuracy for visualization purposes since no model was produced. The figure shows massive decrease in training time with little change in accuracy. 
    The hyperparameters used train reference models are (n\_est and max\_depth) $=$ (20,3) for COMPAS and Spiral, (40,1) for Broward and FICO, (50,2) for Takeaway and Restaurant, and (30,2) for Netherlands. 
    }
    \label{fig:threshold_time_acc}
\end{figure}

\noindent\textbf{Results}: Figure \ref{fig:threshold_time_acc} compares the training time, training accuracy, and test accuracy for GOSDT and DL8.5 with and without threshold guessing. Two subfigures on the left show that the training time after guessing thresholds (blue bars) is orders of magnitude faster than baselines. Note that all baselines either timed out (orange bars with hatches) or hit the memory limit (grey bars with hatches). Subfigures on the right show accuracy with thresholds guessing (gold) is comparable to or sometimes even better than the baseline (purple), since when the model timed out, we report the best model found, and when the model reaches the memory limit, we are unable to obtain any model.

Tables \ref{tab:threshold_guess_gosdt} and \ref{tab:threshold_guess_dl8.5} list the training time, training accuracy, and test accuracy for GOSDT and DL8.5 using different reference models for threshold guessing. For both GOSDT and DL8.5, using threshold guessing often dramatically reduces the training time, with little change in either training or test accuracy. Different reference models introduce different numbers of thresholds in the reduced datasets, thereby influencing the training time and memory usage. Comparing the performance of a single dataset with different reference models, we find little change in test accuracy, but a more complicated reference model can increase training time.

\begin{table}[]\scriptsize
    \centering
    \begin{tabular}{|m{4.5em}|m{2.5em}|m{4.5em}|m{8.5em}|m{7em}|m{7em}|m{6.5em}|m{7em}|m{6.5em}|}\hline
   \multirow{2}{*}{Dataset} &
   \multirow{2}{*}{$\lambda$} &
   (n\_est,&
   \multicolumn{2}{c|}{Training Time} & 
    \multicolumn{2}{c|}{Training Accuracy} &
    \multicolumn{2}{c|}{Test Accuracy}\\
    \cline{4-9}
    & & max\_depth)&  \textbf{gosdt$+$th} & gosdt & \textbf{gosdt$+$th} & gosdt & \textbf{gosdt$+$th} & gosdt\\\hline
     \multirow{4}{*}{COMPAS} &  \multirow{4}{*}{0.001} & (40,1)  & $0.97\pm 0.676$ & \multirow{4}{*}{algorithm timeout} & $0.684 \pm 0.004$ & \multirow{4}{*}{$0.668 \pm 0.002$}& $0.677 \pm 0.014$ & \multirow{4}{*}{$ 0.657 \pm 0.008$}\\
    \cline{3-4}\cline{6-6}\cline{8-8}
    & & (30,2) & $17.688\pm 9.268$ & & $0.688\pm 0.003$ & & $0.681 \pm 0.013$ & \\
    \cline{3-4}\cline{6-6}\cline{8-8}
    & & (50,2) & $ 65.787 \pm 42.872 $ &  & $ 0.688 \pm 0.003 $ & & $ 0.681 \pm 0.013 $ & \\
    \cline{3-4}\cline{6-6}\cline{8-8}
    & & (20,3) & $ 34.807 \pm 15.298 $ & & $ 0.688 \pm 0.003 $ & & $ 0.68 \pm 0.014 $ & \\\hline
    
     \multirow{4}{*}{Broward} &  \multirow{4}{*}{0.005} & (40,1)  & $ 1.659 \pm 0.969 $ & \multirow{4}{*}{memory out}& $ 0.639 \pm 0.004 $ & \multirow{4}{*}{memory out} & $ 0.598 \pm 0.015 $ & \multirow{4}{*}{memory out}\\
     \cline{3-4}\cline{6-6}\cline{8-8}
    & & (30,2) & $ 104.071 \pm 43.033 $ &  & $ 0.646 \pm 0.008 $ &  &$ 0.594 \pm 0.008 $ & \\
     \cline{3-4}\cline{6-6}\cline{8-8}
    & & (50,2) & algorithm timeout &  & $ 0.645 \pm 0.011$ &  & $ 0.589 \pm 0.004$ & \\
     \cline{3-4}\cline{6-6}\cline{8-8}
    & & (20,3) & algorithm timeout & & $ 0.644 \pm 0.007$ & & $0.594 \pm 0.017$ & \\\hline
    
     \multirow{4}{*}{Netherlands} &  \multirow{4}{*}{0.001} & (40,1)  & $ 0.664 \pm 0.817 $ & \multirow{4}{*}{memory out}& $ 0.703 \pm 0.001 $ & \multirow{4}{*}{memory out} & $ 0.701 \pm 0.004 $ & \multirow{4}{*}{memory out}\\
     \cline{3-4}\cline{6-6}\cline{8-8}
    & & (30,2) & $214.298 \pm 119.642$ &  & $0.719 \pm 0.002$ &  & $0.714 \pm 0.005$ & \\
     \cline{3-4}\cline{6-6}\cline{8-8}
    & & (50,2) &  algorithm timeout &  & $0.716 \pm 0.008$ &  & $0.711 \pm 0.009$ &  \\
     \cline{3-4}\cline{6-6}\cline{8-8}
    & & (20,3) & algorithm timeout &  & $ 0.715 \pm 0.005 $ & & $0.711 \pm 0.01$ & \\\hline
    
    \multirow{4}{*}{FICO} &  \multirow{4}{*}{0.0005} & (40,1)  & $ 47.934 \pm 31.746 $ & \multirow{4}{*}{memory out}& $ 0.726 \pm 0.006 $ & \multirow{4}{*}{memory out} & $ 0.709 \pm 0.01 $ & \multirow{4}{*}{memory out}\\
     \cline{3-4}\cline{6-6}\cline{8-8}
    & & (30,2) & algorithm timeout &  & $0.726 \pm 0.011$ & & $0.71 \pm 0.007$ & \\
     \cline{3-4}\cline{6-6}\cline{8-8}
    & & (50,2) & memory out &  & memory out &  & memory out &  \\
     \cline{3-4}\cline{6-6}\cline{8-8}
    & & (20,3) & memory out &  & memory out & & memory out & \\\hline
    
    \multirow{4}{*}{Takeaway} &  \multirow{4}{*}{0.001} & (40,1)  & $8.8e-5\pm 3.4e-5$ & \multirow{4}{*}{algorithm timeout} & $ 0.738 \pm 0.003 $ & \multirow{4}{*}{$0.739 \pm 0.004$} & $ 0.736 \pm 0.01 $ & \multirow{4}{*}{$ 0.737 \pm 0.014 $}\\
     \cline{3-4}\cline{6-6}\cline{8-8}
    & & (30,2) & $ 12.398 \pm 16.004 $ &  & $ 0.765 \pm 0.02 $ & & $ 0.74 \pm 0.016 $ & \\
     \cline{3-4}\cline{6-6}\cline{8-8}
    & & (50,2) & $ 119.158 \pm 149.966 $ &  & $ 0.777 \pm 0.016 $ &  & $ 0.744 \pm 0.017 $ &  \\
     \cline{3-4}\cline{6-6}\cline{8-8}
    & & (20,3) & $ 350.801 \pm 265.191 $ &  & $ 0.796 \pm 0.007 $ & & $ 0.741 \pm 0.01 $ & \\\hline
    
    \multirow{4}{*}{Restaurant} &  \multirow{4}{*}{0.001} & (40,1)  &$ 0.027 \pm 0.034$ & \multirow{4}{*}{algorithm timeout} & $0.765 \pm 0.01$ & \multirow{4}{*}{$0.757 \pm 0.003$} & $0.764 \pm 0.015$ &  \multirow{4}{*}{$0.752 \pm 0.018$}\\
    \cline{3-4}\cline{6-6}\cline{8-8}
    & & (30,2) & $ 45.988 \pm 24.448 $ &  & $ 0.788 \pm 0.006 $ & & $ 0.756 \pm 0.021 $ & \\
    \cline{3-4}\cline{6-6}\cline{8-8}
    & & (50,2) & $174.871 \pm 121.808$ &  & $ 0.792 \pm 0.007 $ &  & $ 0.757 \pm 0.018 $ &  \\
    \cline{3-4}\cline{6-6}\cline{8-8}
    & & (20,3) & $169.309 \pm 122.635$ &  & $0.794 \pm 0.007$ & & $0.761 \pm 0.015$ & \\\hline
    \end{tabular}
    \caption{(\textbf{Value of threshold guessing for GOSDT.}) Comparison of training time (in seconds), training accuracy, and test accuracy for GOSDT at depth limit 5 with and without threshold guessing. }
    \label{tab:threshold_guess_gosdt}
\end{table}

\begin{table}[]\scriptsize
    \centering
    \begin{tabular}{|m{4.5em}|m{2.5em}|m{4.5em}|m{8.5em}|m{7em}|m{7em}|m{6.5em}|m{7em}|m{6.5em}|}\hline
   \multirow{2}{*}{Dataset} &
   \multirow{2}{*}{$\lambda$} &
   (n\_est,&
   \multicolumn{2}{c|}{Training Time} & 
    \multicolumn{2}{c|}{Training Accuracy} &
    \multicolumn{2}{c|}{Test Accuracy}\\
    \cline{4-9}
    & & max\_depth)&  \textbf{dl8.5$+$th} & dl8.5 & \textbf{dl8.5$+$th} & dl8.5 & \textbf{dl8.5$+$th} & dl8.5\\\hline
    \multirow{4}{*}{COMPAS} & \multirow{4}{*}{0.001} & (40,1) & $1.642 \pm 1.208$ & \multirow{4}{*}{algorithm timeout} & $0.688 \pm 0.003$ & \multirow{4}{*}{$0.699\pm0.004$} & $0.677 \pm 0.015$ & \multirow{4}{*}{$0.671\pm0.017$} \\
    \cline{3-4}\cline{6-6}\cline{8-8}
    & & (30,2) & $26.356 \pm 11.334$ & & $0.696 \pm 0.003$ & & $0.676 \pm 0.016$ & \\
    \cline{3-4}\cline{6-6}\cline{8-8}
    & & (50,2) & $92.019 \pm 62.313$ & & $0.697 \pm 0.003$ & & $0.674 \pm 0.017$ & \\
    \cline{3-4}\cline{6-6}\cline{8-8}
    & & (20,3) & $70.623 \pm 35.068$ & & $0.697 \pm 0.003$ & & $0.679 \pm 0.015$ & \\\hline

    \multirow{4}{*}{Broward} & \multirow{4}{*}{0.005} & (40,1) & $0.991 \pm 0.533$ & \multirow{4}{*}{algorithm timeout } & $0.67 \pm 0.004$ & \multirow{4}{*}{$0.666\pm0.004$} & $0.584 \pm 0.023$ & \multirow{4}{*}{$0.59\pm0.009$} \\
    \cline{3-4}\cline{6-6}\cline{8-8}
    & & (30,2) & $57.767 \pm 30.459$ & & $0.7 \pm 0.003$ & & $0.587 \pm 0.015$ & \\
    \cline{3-4}\cline{6-6}\cline{8-8}
    & & (50,2) & $633.444 \pm 397.964$ & & $0.706 \pm 0.001$ & & $0.572 \pm 0.023$ & \\
    \cline{3-4}\cline{6-6}\cline{8-8}
    & & (20,3) & algorithm timeout & & $0.712 \pm 0.002$ & & $0.588 \pm 0.021$ & \\\hline
    
    \multirow{4}{*}{Netherlands} & \multirow{4}{*}{0.001} & (40,1) & $1.309 \pm 1.616$ & \multirow{4}{*}{algorithm timeout} & $0.704 \pm 0.002$ & \multirow{4}{*}{$0.669\pm0.002$} & $0.702 \pm 0.004$ & \multirow{4}{*}{$0.666\pm0.006$} \\
    \cline{3-4}\cline{6-6}\cline{8-8}
    & & (30,2) & $292.535 \pm 150.88$ & & $0.726 \pm 0.001$ & & $0.716 \pm 0.005$ & \\
    \cline{3-4}\cline{6-6}\cline{8-8}
    & & (50,2) & algorithm timeout & & $0.729 \pm 0.001$ & & $0.718 \pm 0.005$ & \\
    \cline{3-4}\cline{6-6}\cline{8-8}
    & & (20,3) & algorithm timeout & & $0.729 \pm 0.001$ & & $0.717 \pm 0.004$ & \\\hline

    \multirow{4}{*}{FICO} & \multirow{4}{*}{0.0005} & (40,1) & $21.517 \pm 12.273$ & \multirow{4}{*}{algorithm timeout} & $0.73 \pm 0.006$ & \multirow{4}{*}{$0.713\pm0.003$} & $0.711 \pm 0.012$ & \multirow{4}{*}{$0.706\pm0.011$} \\
    \cline{3-4}\cline{6-6}\cline{8-8}
    & & (30,2) & algorithm timeout & & $0.737 \pm 0.002$ & & $0.711 \pm 0.011$ & \\
    \cline{3-4}\cline{6-6}\cline{8-8}
    & & (50,2) & algorithm timeout & & $0.737 \pm 0.004$ & & $0.711 \pm 0.012$ & \\
    \cline{3-4}\cline{6-6}\cline{8-8}
    & & ( 20 3 ) & algorithm timeout & & $0.736 \pm 0.004$ & & $0.706 \pm 0.009$ & \\\hline

    \multirow{4}{*}{Takeaway} & \multirow{4}{*}{0.001} & (40,1) & $0.063 \pm 0.005$ & \multirow{4}{*}{algorithm timeout} & $0.738 \pm 0.002$ & \multirow{4}{*}{$0.803\pm0.004$} & $0.735 \pm 0.01$ & \multirow{4}{*}{$0.747\pm0.015$} \\
    \cline{3-4}\cline{6-6}\cline{8-8}
    & & (30,2) & $7.404 \pm 9.458$ & & $0.768 \pm 0.023$ & & $0.74 \pm 0.012$ & \\
    \cline{3-4}\cline{6-6}\cline{8-8}
    & & (50,2) & $112.648 \pm 132.689$ & & $0.781 \pm 0.019$ & & $0.734 \pm 0.01$ & \\
    \cline{3-4}\cline{6-6}\cline{8-8}
    & & (20,3) & $257.187 \pm 198.759$ & & $0.797 \pm 0.006$ & & $0.743 \pm 0.019$ & \\\hline

    \multirow{4}{*}{Restaurant} & \multirow{4}{*}{0.001} & (40,1) & $0.085 \pm 0.016$ & \multirow{4}{*}{algorithm timeout} & $0.768 \pm 0.012$ & \multirow{4}{*}{$0.809\pm0.004$} & $0.756 \pm 0.017$ & \multirow{4}{*}{$0.752\pm0.016$} \\
    \cline{3-4}\cline{6-6}\cline{8-8}
    & & (30,2) & $21.292 \pm 12.183$ & & $0.797 \pm 0.004$ & & $0.756 \pm 0.016$ & \\
    \cline{3-4}\cline{6-6}\cline{8-8}
    & & (50,2) & $106.846 \pm 77.748$ & & $0.799 \pm 0.006$ & & $0.751 \pm 0.013$ & \\
    \cline{3-4}\cline{6-6}\cline{8-8}
    & & (20,3) & $119.682 \pm 92.536$ & & $0.801 \pm 0.006$ & & $0.752 \pm 0.008$ & \\\hline
    \end{tabular}
    \caption{(\textbf{Value of threshold guessing for DL8.5.}) Comparison of training time (in second), training accuracy, and test accuracy for DL8.5 at depth limit 5 with and without threshold guessing. }
    \label{tab:threshold_guess_dl8.5}
\end{table}

\subsection{Value of Lower Bound Guessing}
\label{app:lowerbounds}
\noindent\textbf{Calculations}: we use the same calculations as described in Appendix \ref{app:thresholds}. 

\noindent\textbf{Results}: Table \ref{tab:lb_guess_gosdt} and \ref{tab:lb_guess_dl8.5} list the training time, training accuracy, and test accuracy for GOSDT and DL8.5 respectively using different reference models for lower bound guessing. For both GOSDT and DL8.5, using lower bound guessing after applying threshold guessing can further reduce the training time with little change in either training or test accuracy. Sometimes the training accuracy decreases by 1\%, but we still obtain test accuracy close to that of the baseline. 

\begin{table}[]\scriptsize
    \centering
    \begin{tabular}{|m{4.5em}|m{2.5em}|m{4.5em}|m{8.5em}|m{8.5em}|m{7em}|m{6.5em}|m{7em}|m{6.5em}|}\hline
   \multirow{2}{*}{Dataset} &
   \multirow{2}{*}{$\lambda$} &
   (n\_est,&
   \multicolumn{2}{c|}{Training Time} & 
    \multicolumn{2}{c|}{Training Accuracy} &
    \multicolumn{2}{c|}{Test Accuracy}\\
    \cline{4-9}
    & & max\_depth)& \textbf{gosdt$+$th$+$lb} & gosdt$+$th & \textbf{gosdt$+$th$+$lb} & gosdt$+$th & \textbf{gosdt$+$th$+$lb} & gosdt$+$th\\\hline

\multirow{4}{*}{COMPAS} & \multirow{4}{*}{0.001} & (40,1) & $1.115 \pm 0.729$ & $0.97\pm0.676$ & $0.684 \pm 0.004$ & $0.684\pm0.004$ & $0.677 \pm 0.014$ & $0.677\pm0.014$  \\
\cline{3-9}
& & (30,2) & $14.019 \pm 7.266$ & $17.688\pm9.268$ & $0.688 \pm 0.003$ & $0.688\pm0.003$ & $0.681 \pm 0.013$ & $0.681\pm0.013$  \\
\cline{3-9}
& & (50,2) & $47.026 \pm 30.745$ & $65.787\pm42.872$ & $0.688 \pm 0.003$ & $0.688\pm0.003$ & $0.681 \pm 0.013$ & $0.681\pm0.013$  \\
\cline{3-9}
& & (20,3) & $25.94 \pm 9.083$ & $34.807\pm15.298$ & $0.688 \pm 0.003$ & $0.688\pm0.003$ & $0.681 \pm 0.013$ & $0.68\pm0.014$  \\\hline

\multirow{4}{*}{Broward} & \multirow{4}{*}{0.005} & (40,1) & $0.341 \pm 0.343$ & $1.659\pm0.969$ & $0.638 \pm 0.004$ & $0.639\pm0.004$ & $0.603 \pm 0.016$ & $0.598\pm0.015$  \\
\cline{3-9}
& & (30,2) & $73.397 \pm 61.751$ & $104.071\pm43.033$ & $0.645 \pm 0.006$ & $0.646\pm0.008$ & $0.593 \pm 0.009$ & $0.594\pm0.008$  \\
\cline{3-9}
& & (50,2) & algorithm timeout & algorithm timeout & $0.647 \pm 0.006$ & $0.645\pm0.011$ & $0.588 \pm 0.006$ & $0.589\pm0.004$  \\
\cline{3-9}
& & (20,3) & algorithm timeout & algorithm timeout & $0.648 \pm 0.009$ & $0.644\pm0.007$ & $0.589 \pm 0.011$ & $0.594\pm0.017$  \\\hline

\multirow{4}{*}{Netherlands} & \multirow{4}{*}{0.001} & (40,1) & $0.445 \pm 0.621$ & $0.664\pm0.817$ & $0.702 \pm 0.001$ & $0.703\pm0.001$ & $0.701 \pm 0.005$ & $0.701\pm0.004$  \\
\cline{3-9}
& & (30,2) & $123.816 \pm 84.661$ & $214.298\pm119.642$ & $0.716 \pm 0.001$ & $0.719\pm0.002$ & $0.713 \pm 0.004$ & $0.714\pm0.005$  \\
\cline{3-9}
& & (50,2) & algorithm timeout & algorithm timeout & $0.715 \pm 0.008$ & $0.716\pm0.008$ & $0.711 \pm 0.008$ & $0.711\pm0.009$  \\
\cline{3-9}
& & (20,3) & algorithm timeout & algorithm timeout & $0.716 \pm 0.006$ & $0.715\pm0.005$ & $0.712 \pm 0.012$ & $0.711\pm0.01$  \\\hline

\multirow{4}{*}{FICO} & \multirow{4}{*}{0.0005} & (40,1) & $9.598 \pm 5.99$ & $47.934\pm31.746$ & $0.72 \pm 0.003$ & $0.726\pm0.006$ & $0.712 \pm 0.009$ & $0.709\pm0.01$  \\
\cline{3-9}
& & (30,2) & algorithm timeout & algorithm timeout & $0.729 \pm 0.003$ & $0.726\pm0.011$ & $0.715 \pm 0.009$ & $0.71\pm0.007$  \\
\cline{3-9}
& & (50,2) & algorithm timeout & memory out & $0.712 \pm 0.012$ & memory out & $0.705 \pm 0.009$ & memory out  \\
\cline{3-9}
& & (20,3) & algorithm timeout & memory out & $0.703 \pm 0.003$ & memory out & $0.697 \pm 0.012$ & memory out  \\\hline

\multirow{4}{*}{Takeaway} & \multirow{4}{*}{0.001} & (40,1) & NA & $0.0\pm0.0$ & NA & $0.738\pm0.003$ & NA & $0.736\pm0.01$  \\
\cline{3-9}
& & (30,2) & $1.82 \pm 2.435$ & $12.398\pm16.004$ & $0.751 \pm 0.007$ & $0.765\pm0.02$ & $0.738 \pm 0.018$ & $0.74\pm0.016$  \\
\cline{3-9}
& & (50,2) & $45.672 \pm 48.383$ & $119.158\pm149.966$ & $0.763 \pm 0.01$ & $0.777\pm0.016$ & $0.745 \pm 0.025$ & $0.744\pm0.017$\\
\cline{3-9}
& & (20,3) & $124.575 \pm 95.884$ & $350.801\pm265.191$ & $0.765 \pm 0.009$ & $0.796\pm0.007$ & $0.746 \pm 0.02$ & $0.741\pm0.01$  \\\hline

\multirow{4}{*}{Restaurant} & \multirow{4}{*}{0.001} & (40,1) & $0.003 \pm 0.002$ & $0.027\pm0.034$ & $0.76 \pm 0.007$ & $0.765\pm0.01$ & $0.753 \pm 0.018$ & $0.764\pm0.015$  \\
\cline{3-9}
& & (30,2) & $16.607 \pm 9.743$ & $45.988\pm24.448$ & $0.774 \pm 0.005$ & $0.788\pm0.006$ & $0.757 \pm 0.022$ & $0.756\pm0.021$  \\
\cline{3-9}
& & (50,2) & $97.332 \pm 70.63$ & $174.871\pm121.808$ & $0.781 \pm 0.007$ & $0.792\pm0.007$ & $0.757 \pm 0.025$ & $0.757\pm0.018$  \\
\cline{3-9}
& & (20,3) & $57.195 \pm 30.067$ & $169.309\pm122.635$ & $0.777 \pm 0.003$ & $0.794\pm0.007$ & $0.762 \pm 0.014$ & $0.761\pm0.015$  \\\hline
    \end{tabular}
    \caption{(\textbf{Value of lower bound guessing for GOSDT.}) Comparison of training time (in second), training accuracy, and test accuracy for GOSDT at depth limit 5 with and without lower bound guessing. NA on the Takeaway dataset is because GOSDT doesn't support a lower bound reference model that predicts only one class.}
    \label{tab:lb_guess_gosdt}
\end{table}

\begin{table}[]\scriptsize
    \centering
    \begin{tabular}{|m{4.5em}|m{2.5em}|m{4.5em}|m{8.5em}|m{8.5em}|m{7em}|m{6.5em}|m{7em}|m{6.5em}|}\hline
   \multirow{2}{*}{Dataset} &
   \multirow{2}{*}{$\lambda$} &
   (n\_est,&
   \multicolumn{2}{c|}{Training Time} & 
    \multicolumn{2}{c|}{Training Accuracy} &
    \multicolumn{2}{c|}{Test Accuracy}\\
    \cline{4-9}
    & & max\_depth)& \textbf{dl8.5$+$th$+$lb} & dl8.5$+$th & \textbf{dl8.5$+$th$+$lb} & dl8.5$+$th & \textbf{dl8.5$+$th$+$lb} & dl8.5$+$th\\\hline

\multirow{4}{*}{COMPAS} & \multirow{4}{*}{0.001} & (40,1) & $0.078 \pm 0.014$ & $1.642\pm1.208$ & $0.684 \pm 0.004$ & $0.688\pm0.003$ & $0.677 \pm 0.015$ & $0.677\pm0.015$\\
\cline{3-9}
& & (30,2) & $0.575 \pm 0.413$ & $26.356\pm11.334$ & $0.689 \pm 0.004$ & $0.696\pm0.003$ & $0.681 \pm 0.011$ & $0.676\pm0.016$\\
\cline{3-9}
& & (50,2) & $0.801 \pm 0.603$ & $92.019\pm62.313$ & $0.69 \pm 0.005$ & $0.697\pm0.003$ & $0.677 \pm 0.015$ & $0.674\pm0.017$\\
\cline{3-9}
& & (20,3) & $0.515 \pm 0.332$ & $70.623\pm35.068$ & $0.69 \pm 0.004$ & $0.697\pm0.003$ & $0.682 \pm 0.014$ & $0.679\pm0.015$\\\hline

\multirow{4}{*}{Broward} & \multirow{4}{*}{0.005} & (40,1) & $0.136 \pm 0.076$ & $0.991\pm0.533$ & $0.645 \pm 0.005$ & $0.67\pm0.004$ & $0.591 \pm 0.012$ & $0.584\pm0.023$  \\
\cline{3-9}
& & (30,2) & $0.233 \pm 0.122$ & $57.767\pm30.459$ & $0.661 \pm 0.005$ & $0.7\pm0.003$ & $0.601 \pm 0.022$ & $0.587\pm0.015$  \\
\cline{3-9}
& & (50,2) & $4.124 \pm 1.828$ & $633.444\pm397.964$ & $0.675 \pm 0.003$ & $0.706\pm0.001$ & $0.592 \pm 0.016$ & $0.572\pm0.023$  \\
\cline{3-9}
& & (20,3) & $71.588 \pm 71.969$ & algorithm timeout & $0.684 \pm 0.009$ & $0.712\pm0.002$ & $0.613 \pm 0.012$ & $0.588\pm0.021$  \\\hline

\multirow{4}{*}{Netherlands} & \multirow{4}{*}{0.001} & (40,1) & $0.099 \pm 0.036$ & $1.309\pm1.616$ & $0.699 \pm 0.004$ & $0.704\pm0.002$ & $0.698 \pm 0.004$ & $0.702\pm0.004$  \\
\cline{3-9}
& & (30,2) & $0.137 \pm 0.021$ & $292.535\pm150.88$ & $0.714 \pm 0.002$ & $0.726\pm0.001$ & $0.709 \pm 0.007$ & $0.716\pm0.005$  \\
\cline{3-9}
& & (50,2) & $0.376 \pm 0.144$ & algorithm timeout & $0.718 \pm 0.001$ & $0.729\pm0.001$ & $0.716 \pm 0.003$ & $0.718\pm0.005$  \\
\cline{3-9}
& & (20,3) & $0.65 \pm 0.652$ & algorithm timeout & $0.721 \pm 0.001$ & $0.729\pm0.001$ & $0.716 \pm 0.004$ & $0.717\pm0.004$  \\\hline

\multirow{4}{*}{FICO} & \multirow{4}{*}{0.0005} & (40,1) & $0.147 \pm 0.049$ & $21.517\pm12.273$ & $0.716 \pm 0.003$ & $0.73\pm0.006$ & $0.709 \pm 0.01$ & $0.711\pm0.012$  \\
\cline{3-9}
& & (30,2) & $5.355 \pm 2.045$ & algorithm timeout & $0.724 \pm 0.003$ & $0.737\pm0.002$ & $0.709 \pm 0.012$ & $0.711\pm0.011$  \\
\cline{3-9}
& & (50,2) & $187.762 \pm 224.422$ & algorithm timeout & $0.728 \pm 0.003$ & $0.737\pm0.004$ & $0.71 \pm 0.015$ & $0.711\pm0.012$  \\
\cline{3-9}
& & (20,3) & $270.019 \pm 220.485$ & algorithm timeout & $0.727 \pm 0.002$ & $0.736\pm0.004$ & $0.707 \pm 0.013$ & $0.706\pm0.009$  \\\hline

\multirow{4}{*}{Takeaway} & \multirow{4}{*}{0.001} & (40,1) & $0.066 \pm 0.01$ & $0.063\pm0.005$ & $0.738 \pm 0.003$ & $0.738\pm0.002$ & $0.738 \pm 0.013$ & $0.735\pm0.01$  \\
\cline{3-9}
& & (30,2) & $0.096 \pm 0.032$ & $7.404\pm9.458$ & $0.751 \pm 0.006$ & $0.768\pm0.023$ & $0.737 \pm 0.024$ & $0.74\pm0.012$  \\
\cline{3-9}
& & (50,2) & $0.088 \pm 0.009$ & $112.648\pm132.689$ & $0.76 \pm 0.007$ & $0.781\pm0.019$ & $0.744 \pm 0.019$ & $0.734\pm0.01$  \\
\cline{3-9}
& & (20,3) & $0.088 \pm 0.012$ & $257.187\pm198.759$ & $0.763 \pm 0.004$ & $0.797\pm0.006$ & $0.739 \pm 0.018$ & $0.743\pm0.019$  \\\hline

\multirow{4}{*}{Restaurant} & \multirow{4}{*}{0.001} & (40,1) & $0.07 \pm 0.009$ & $0.085\pm0.016$ & $0.762 \pm 0.008$ & $0.768\pm0.012$ & $0.762 \pm 0.017$ & $0.756\pm0.017$  \\
\cline{3-9}
& & (30,2) & $0.082 \pm 0.026$ & $21.292\pm12.183$ & $0.771 \pm 0.003$ & $0.797\pm0.004$ & $0.761 \pm 0.017$ & $0.756\pm0.016$  \\
\cline{3-9}
& & (50,2) & $0.106 \pm 0.016$ & $106.846\pm77.748$ & $0.778 \pm 0.004$ & $0.799\pm0.006$ & $0.764 \pm 0.016$ & $0.751\pm0.013$\\
\cline{3-9}
& & (20,3) & $0.581 \pm 0.819$ & $119.682\pm92.536$ & $0.776 \pm 0.003$ & $0.801\pm0.006$ & $0.76 \pm 0.017$ & $0.752\pm0.008$\\\hline
    \end{tabular}
    \caption{(\textbf{Value of lower bound guessing for DL8.5.}) Comparison of training time (in second), training accuracy, and test accuracy for DL8.5 at depth limit 5 with and without lower bound guessing. }
    \label{tab:lb_guess_dl8.5}
\end{table}


\subsection{Trees}\label{app:trees}
\begin{figure}
    \centering
    \begin{subfigure}[t]{0.45\textwidth}
    \centering
\scalebox{0.6}{\begin{forest}
[ $age \le 21.5$ [ $class$ [ $1$ ] ] [ $age \le 33.5$ [ $priors\_count \le 1.5$ [ $class$ [ $0$ ] ] [ $class$ [ $1$ ] ] ] [ $priors\_count \le 6.5$ [ $class$ [ $0$ ] ] [ $class$ [ $1$ ] ] ] ] ]
\end{forest}}
\caption{GOSDT+th+lb (depth limit 3, $\lambda\!=\!0.005$): training time=0.379 (sec), training accuracy=0.678, test accuracy=0.687, \# leaves=5 on fold 2. }
    \end{subfigure}
    \hfill
    \begin{subfigure}[t]{0.45\textwidth}
    \scalebox{0.6}{\begin{forest}
[ $age \le 27.5$ [ $age \le 22.5$ [ $sex\ not\ female$ [ $class$ [ $1$ ] ] [ $class$ [ $0$ ] ] ] [ $priors\_count \le 1.5$ [ $class$ [ $0$ ] ] [ $class$ [ $1$ ] ] ] ] [ $age \le 36.5$ [ $priors\_count \le 2.5$ [ $class$ [ $0$ ] ] [ $class$ [ $1$ ] ] ] [ $priors\_count \le 7.5$ [ $class$ [ $0$ ] ] [ $class$ [ $1$ ] ] ] ] ]
\end{forest}}
\caption{DL8.5+th+lb (depth limit 3, $\lambda\!=\!0.005$): training time=0.498 (sec), training accuracy=0.683, test accuracy=0.695, \# leaves=8 on fold 2 }
    \end{subfigure}
    \caption{GOSDT and DL8.5 trees on the COMPAS dataset with guessed thresholds and guessed lower bounds at depth limit 3. The reference model was trained using 20 max-depth 3 weak classifiers. $\lambda=0.005$.}
    \label{fig:compas_tree_d3_005}
\end{figure}
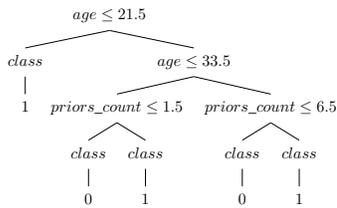
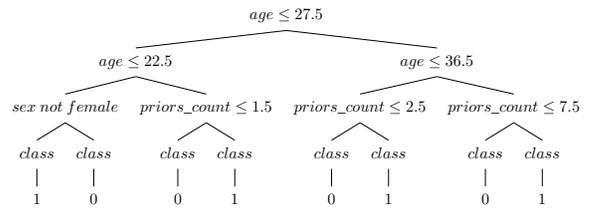

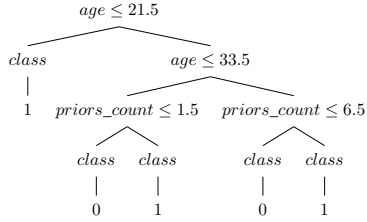
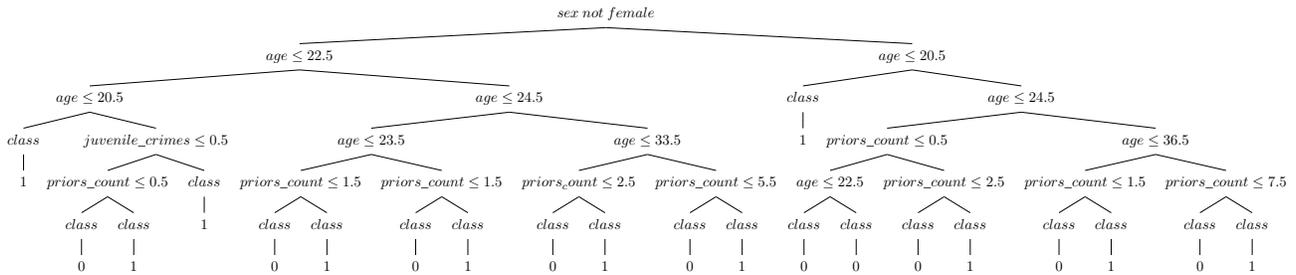
\begin{figure}
    \centering
    \begin{subfigure}[t]{0.98\textwidth}
    \centering
\scalebox{0.65}{\begin{forest}
[ $age \le 21.5$ [ $class$ [ $1$ ] ] [ $age \le 33.5$ [ $priors\_count \le 1.5$ [ $class$ [ $0$ ] ] [ $class$ [ $1$ ] ] ] [ $priors\_count \le 6.5$ [ $class$ [ $0$ ] ] [ $class$ [ $1$ ] ] ] ] ]
\end{forest}}
\caption{GOSDT+th+lb (depth limit 5, $\lambda\!=\!0.005$): training time=22.599 (sec), training accuracy=0.678, test accuracy=0.687, \# leaves=5 on fold 2.}
    \end{subfigure}
    \vskip\baselineskip
    \begin{subfigure}[t]{0.98\textwidth}
    \centering
\scalebox{0.55}{\begin{forest}
[ $sex\ not\ female$ [ $age \le 22.5$ [ $age \le 20.5$ [ $class$ [ $1$ ] ] [ $juvenile\_crimes \le 0.5$ [ $priors\_count \le 0.5$ [ $class$ [ $0$ ] ] [ $class$ [ $1$ ] ] ] [ $class$ [ $1$ ] ] ] ] [ $age \le 24.5$ [ $age \le 23.5$ [ $priors\_count \le 1.5$ [ $class$ [ $0$ ] ] [ $class$ [ $1$ ] ] ] [ $priors\_count \le 1.5$ [ $class$ [ $0$ ] ] [ $class$ [ $1$ ] ] ] ] [ $age \le 33.5$ [ $priors_count \le 2.5$ [ $class$ [ $0$ ] ] [ $class$ [ $1$ ] ] ] [ $priors\_count \le 5.5$ [ $class$ [ $0$ ] ] [ $class$ [ $1$ ] ] ] ] ] ] [ $age \le 20.5$ [ $class$ [ $1$ ] ] [ $age \le 24.5$ [ $priors\_count \le 0.5$ [ $age \le 22.5$ [ $class$ [ $0$ ] ] [ $class$ [ $0$ ] ] ] [ $priors\_count \le 2.5$ [ $class$ [ $0$ ] ] [ $class$ [ $1$ ] ] ] ] [ $age \le 36.5$ [ $priors\_count \le 1.5$ [ $class$ [ $0$ ] ] [ $class$ [ $1$ ] ] ] [ $priors\_count \le 7.5$ [ $class$ [ $0$ ] ] [ $class$ [ $1$ ] ] ] ] ] ] ]
\end{forest}}
\caption{DL8.5+th+lb (depth limit 5, $\lambda\!=\!0.005$): training time=0.549 (sec), training accuracy=0.686, test accuracy=0.694, \# leaves=21 on fold 2.}
    \end{subfigure}
    \caption{GOSDT and DL8.5 trees on the COMPAS dataset with guessed thresholds and guessed lower bounds at depth limit 5. The reference model was trained using 20 max-depth 3 weak classifiers. $\lambda=0.005$.}
    \label{fig:compas_tree_d5_005}
\end{figure}

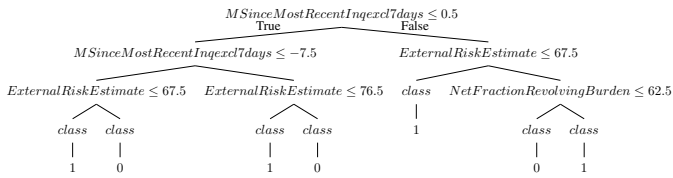
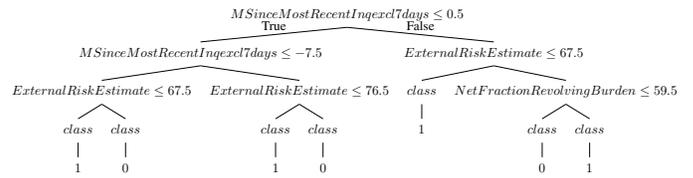
\begin{figure}
\centering
\begin{subfigure}[t]{0.48\textwidth}
\centering
\scalebox{0.5}{\begin{forest}
[ $MSinceMostRecentInqexcl7days \le 0.5$ [ $MSinceMostRecentInqexcl7days \le -7.5$, edge label={node[midway, above] {True}} [ $ExternalRiskEstimate \le 67.5$ [ $class$ [ $1$ ] ] [ $class$ [ $0$ ] ] ] [ $ExternalRiskEstimate \le 76.5$ [ $class$ [ $1$ ] ] [ $class$ [ $0$ ] ] ] ] [ $ExternalRiskEstimate \le 67.5$, edge label={node[midway, above] {False}} [ $class$ [ $1$ ] ] [ $NetFractionRevolvingBurden \le 62.5$ [ $class$ [ $0$ ] ] [ $class$ [ $1$ ] ] ] ] ]
\end{forest}}
\caption{GOSDT+th+lb (depth limit 3, $\lambda\!=\!0.0005$): training time=0.232 (sec), training accuracy=0.714, test accuracy=0.71, \#leaves=7 on fold 4.}
\end{subfigure}
\hfill
\begin{subfigure}[t]{0.48\textwidth}
\centering
\scalebox{0.5}{\begin{forest}
[ $MSinceMostRecentInqexcl7days \le 0.5$ [ $MSinceMostRecentInqexcl7days \le -7.5$, edge label={node[midway, above] {True}} [ $ExternalRiskEstimate \le 67.5$ [ $class$ [ $1$ ] ] [ $class$ [ $0$ ] ] ] [ $ExternalRiskEstimate \le 76.5$ [ $class$ [ $1$ ] ] [ $class$ [ $0$ ] ] ] ] [ $ExternalRiskEstimate \le 67.5$, edge label={node[midway, above] {False}} [ $class$ [ $1$ ] ] [ $NetFractionRevolvingBurden \le 59.5$ [ $class$ [ $0$ ] ] [ $class$ [ $1$ ] ] ] ] ]
\end{forest}}
\caption{DL8.5+th+lb (depth limit 3, $\lambda\!=\!0.0005$): training time=0.279 (sec), training accuracy=0.714, test accuracy=0.709, \#leaves=7 on fold 4 }
\end{subfigure}
\caption{GOSDT and DL8.5 trees on the FICO dataset with guessed thresholds and guessed lower bounds at depth limit 3. The reference model was trained using 40 decision stumps. $\lambda=0.0005$.}
\label{fig:fico_tree_d3}
\end{figure}

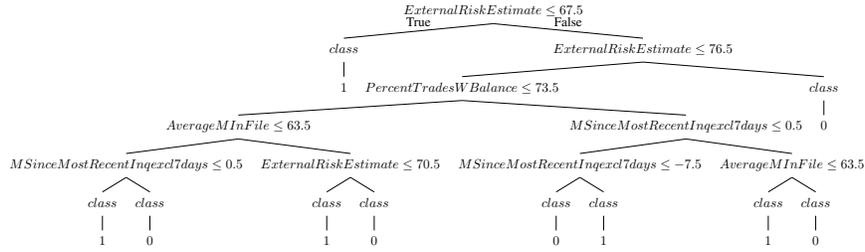
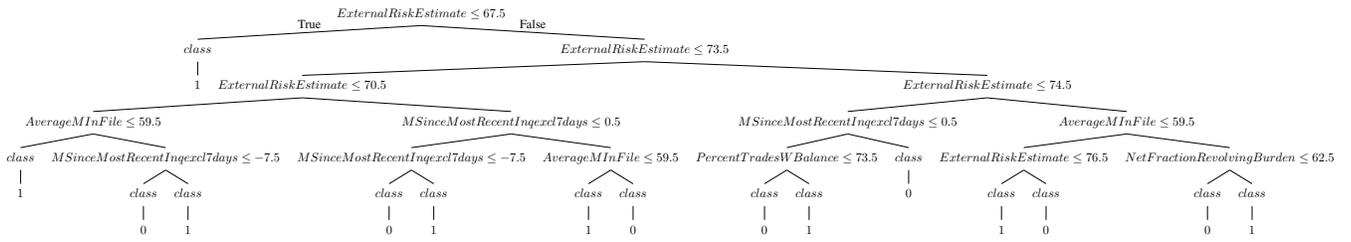
\begin{figure}
\centering
\begin{subfigure}{0.98\textwidth}
\centering
\scalebox{0.5}{\begin{forest}
[ $ExternalRiskEstimate \le 67.5$ [ $class$, edge label={node[midway, above] {True}} [ $1$ ] ] [ $ExternalRiskEstimate \le 76.5$, edge label={node[midway, above] {False}} [ $PercentTradesWBalance \le 73.5$ [ $AverageMInFile \le 63.5$ [ $MSinceMostRecentInqexcl7days \le 0.5$ [ $class$ [ $1$ ] ] [ $class$ [ $0$ ] ] ] [ $ExternalRiskEstimate \le 70.5$ [ $class$ [ $1$ ] ] [ $class$ [ $0$ ] ] ] ] [ $MSinceMostRecentInqexcl7days \le 0.5$ [ $MSinceMostRecentInqexcl7days \le -7.5$ [ $class$ [ $0$ ] ] [ $class$ [ $1$ ] ] ] [ $AverageMInFile \le 63.5$ [ $class$ [ $1$ ] ] [ $class$ [ $0$ ] ] ] ] ] [ $class$ [ $0$ ] ] ] ]
\end{forest}}
\caption{GOSDT+th+lb (depth limit 5, $\lambda\!=\!0.0005$): training time=10.822 (sec), training accuracy=0.72, test accuracy=0.717, \# leaves=10 on fold 4.}
\end{subfigure}
\vskip\baselineskip
\begin{subfigure}{0.98\textwidth}
\centering
\scalebox{0.47}{\begin{forest}
[ $ExternalRiskEstimate \le 67.5$ [ $class$, edge label={node[midway, above] {True}} [ $1$ ] ] [ $ExternalRiskEstimate \le 73.5$, edge label={node[midway, above] {False}} [ $ExternalRiskEstimate \le 70.5$ [ $AverageMInFile \le 59.5$ [ $class$ [ $1$ ] ] [ $MSinceMostRecentInqexcl7days \le -7.5$ [ $class$ [ $0$ ] ] [ $class$ [ $1$ ] ] ] ] [ $MSinceMostRecentInqexcl7days \le 0.5$ [ $MSinceMostRecentInqexcl7days \le -7.5$ [ $class$ [ $0$ ] ] [ $class$ [ $1$ ] ] ] [ $AverageMInFile \le 59.5$ [ $class$ [ $1$ ] ] [ $class$ [ $0$ ] ] ] ] ] [ $ExternalRiskEstimate \le 74.5$ [ $MSinceMostRecentInqexcl7days \le 0.5$ [ $PercentTradesWBalance \le 73.5$ [ $class$ [ $0$ ] ] [ $class$ [ $1$ ] ] ] [ $class$ [ $0$ ] ] ] [ $AverageMInFile \le 59.5$ [ $ExternalRiskEstimate \le 76.5$ [ $class$ [ $1$ ] ] [ $class$ [ $0$ ] ] ] [ $NetFractionRevolvingBurden \le 62.5$ [ $class$ [ $0$ ] ] [ $class$ [ $1$ ] ] ] ] ] ] ]
\end{forest}}
\caption{DL8.5+th+lb (depth limit 5, $\lambda\!=\!0.0005$): training time=0.198 (sec), training accuracy=0.716, test accuracy=0.709, \# leaves=15 on fold 4.}
\end{subfigure}
\caption{GOSDT and DL8.5 trees on the FICO dataset with guessed thresholds and guessed lower bounds at depth limit 5. The reference model was trained using 40 decision stumps. $\lambda=0.0005$.}
\label{fig:fico_tree_d5}
\end{figure}

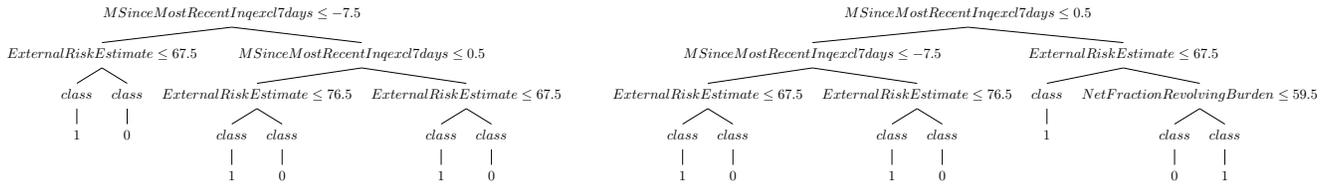
\begin{figure}
    \centering
    \begin{subfigure}[t]{0.4\textwidth}
    \centering
\scalebox{0.53}{\begin{forest}
[ $MSinceMostRecentInqexcl7days \le -7.5$ [ $ExternalRiskEstimate \le 67.5$ [ $class$ [ $1$ ] ] [ $class$ [ $0$ ] ] ] [ $MSinceMostRecentInqexcl7days \le 0.5$ [ $ExternalRiskEstimate \le 76.5$ [ $class$ [ $1$ ] ] [ $class$ [ $0$ ] ] ] [ $ExternalRiskEstimate \le 67.5$ [ $class$ [ $1$ ] ] [ $class$ [ $0$ ] ] ] ] ]
\end{forest}}
\caption{GOSDT+th+lb (depth limit 3, $\lambda$=0.001): training time=0.205 (sec), training accuracy=0.713, test accuracy=0.711, \# leaves=6 on fold 4.}
    \end{subfigure}
    \hfill
    \begin{subfigure}[t]{0.56\textwidth}
    \centering
    \scalebox{0.53}{\begin{forest}
[ $MSinceMostRecentInqexcl7days \le 0.5$ [ $MSinceMostRecentInqexcl7days \le -7.5$ [ $ExternalRiskEstimate \le 67.5$ [ $class$ [ $1$ ] ] [ $class$ [ $0$ ] ] ] [ $ExternalRiskEstimate \le 76.5$ [ $class$ [ $1$ ] ] [ $class$ [ $0$ ] ] ] ] [ $ExternalRiskEstimate \le 67.5$ [ $class$ [ $1$ ] ] [ $NetFractionRevolvingBurden \le 59.5$ [ $class$ [ $0$ ] ] [ $class$ [ $1$ ] ] ] ] ]
\end{forest}}
\caption{DL8.5+th+lb (depth limit 3, $\lambda\!=\!0.001$): training time=0.23 (sec), training accuracy=0.714, test accuracy=0.709, \# leaves=7 on fold 4.}
    \end{subfigure}
    \caption{GOSDT and DL8.5 trees on the FICO dataset with guessed thresholds and guessed lower bounds at depth limit 3. The reference model was trained using 40 decision stumps. $\lambda=0.001$.}
    \label{fig:fico_tree_d3_001}
\end{figure}

\begin{figure}
    \centering
    \begin{subfigure}[t]{0.98\textwidth}
    \centering
\scalebox{0.6}{\begin{forest}
[ $ExternalRiskEstimate \le 70.5$ [ $class$ [ $1$ ] ] [ $ExternalRiskEstimate \le 78.5$ [ $MSinceMostRecentInqexcl7days \le 0.5$ [ $PercentTradesWBalance \le 73.5$ [ $AverageMInFile \le 63.5$ [ $class$ [ $1$ ] ] [ $class$ [ $0$ ] ] ] [ $MSinceMostRecentInqexcl7days \le -7.5$ [ $class$ [ $0$ ] ] [ $class$ [ $1$ ] ] ] ] [ $class$ [ $0$ ] ] ] [ $class$ [ $0$ ] ] ] ]
\end{forest}}
\caption{GOSDT+th+lb (depth limit 5, $\lambda\!=\!0.001$): training time=8.112 (sec), training accuracy=0.717, test accuracy=0.719, \# leaves=7 on fold 4. }
    \end{subfigure}
    \vskip\baselineskip
    \begin{subfigure}[t]{0.98\textwidth}
    \centering
\scalebox{0.48}{\begin{forest}
[ $ExternalRiskEstimate \le 67.5$ [ $class$ [ $1$ ] ] [ $ExternalRiskEstimate \le 73.5$ [ $ExternalRiskEstimate \le 70.5$ [ $AverageMInFile \le 59.5$ [ $class$ [ $1$ ] ] [ $MSinceMostRecentInqexcl7days \le -7.5$ [ $class$ [ $0$ ] ] [ $class$ [ $1$ ] ] ] ] [ $MSinceMostRecentInqexcl7days \le 0.5$ [ $MSinceMostRecentInqexcl7days \le -7.5$ [ $class$ [ $0$ ] ] [ $class$ [ $1$ ] ] ] [ $AverageMInFile \le 59.5$ [ $class$ [ $1$ ] ] [ $class$ [ $0$ ] ] ] ] ] [ $ExternalRiskEstimate \le 74.5$ [ $MSinceMostRecentInqexcl7days \le 0.5$ [ $PercentTradesWBalance \le 73.5$ [ $class$ [ $0$ ] ] [ $class$ [ $1$ ] ] ] [ $class$ [ $0$ ] ] ] [ $AverageMInFile \le 59.5$ [ $ExternalRiskEstimate \le 76.5$ [ $class$ [ $1$ ] ] [ $class$ [ $0$ ] ] ] [ $NetFractionRevolvingBurden \le 62.5$ [ $class$ [ $0$ ] ] [ $class$ [ $1$ ] ] ] ] ] ] ]
\end{forest}}
\caption{DL8.5+th+lb (depth limit 5, $\lambda\!=\!0.001$): training time=0.195 (sec), training accuracy=0.716, test accuracy=0.709, \# leaves=15 on fold 4. }
    \end{subfigure}
    \caption{GOSDT and DL8.5 trees on the FICO dataset with guessed thresholds and guessed lower bounds at depth limit 3. The reference model was trained using 40 decision stumps. $\lambda=0.001$.}
    \label{fig:fico_tree_d5_001}
\end{figure}
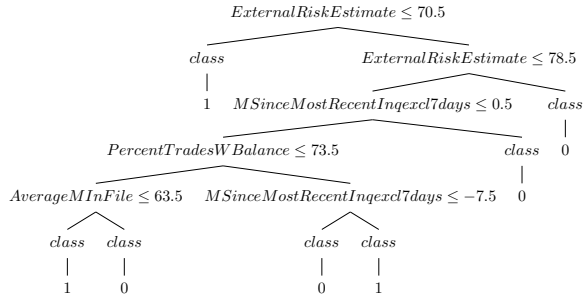
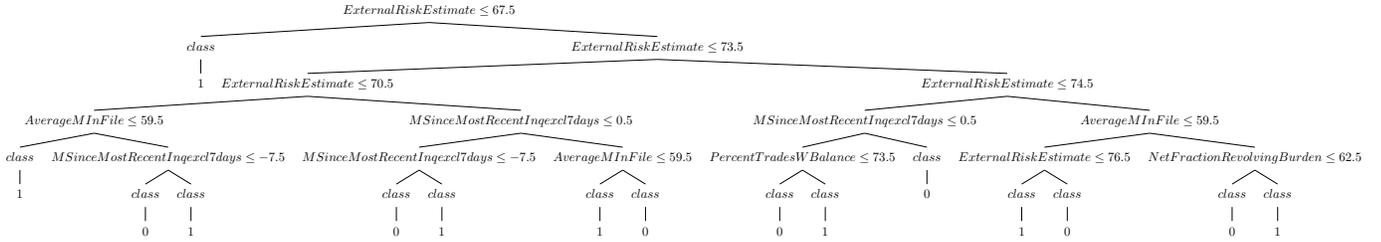

In this section, we show some trees generated by GOSDT and DL8.5 on COMPAS (Figure \ref{fig:compas_tree_d3}, \ref{fig:compas_tree_d5}, \ref{fig:compas_tree_d3_005}, \ref{fig:compas_tree_d5_005}) and FICO (Figure \ref{fig:fico_tree_d3}, \ref{fig:fico_tree_d5}, \ref{fig:fico_tree_d3_001}, \ref{fig:fico_tree_d5_001}) using all guessing strategies. GOSDT and DL8.5 with all guessing strategies tend to achieve similar accuracy and sparsity when depth limit is relatively small (depth limit 3). When the depth limit increases to 5, GOSDT and DL8.5 with all guessing strategies still tend to achieve similar accuracy, but the trees generated by GOSDT with all guessing strategies tend to be sparser than those generated by DL8.5 with all guessing strategies.

\begin{figure}
\centering
\begin{subfigure}{0.497\textwidth}
\centering
    \scalebox{0.6}{\begin{forest}
[ $age \le 27.5$
[ $age \le 21.5$, edge label={node[midway, above] {True}} [ $class$ [ $1$ ] ] [ $priors\_count \le 1.5$ [ $class$ [ $0$ ] ] [ $class$ [ $1$ ] ] ] ] [ $age \le 36.5$,edge label={node[midway, above] {False}} [ $priors\_count \le 2.5$ [ $class$ [ $0$ ] ] [ $class$ [ $1$ ] ] ] [ $priors\_count \le 7.5$ [ $class$ [ $0$ ] ] [ $class$ [ $1$ ] ] ] ] ]
\end{forest}}
\caption{GOSDT+th+lb (depth limit 3, $\lambda\!=\!0.001$): training time=0.513 (sec), training accuracy=0.683, test accuracy=0.689, \#leaves=7 on fold 2.}
\end{subfigure}
\hfill
\begin{subfigure}{0.49\textwidth}
\centering
    \scalebox{0.6}{\begin{forest}
[ $age \le 27.5$
[ $age \le 22.5$, edge label={node[midway, above] {True}} [ $sex\ not \ female$ [ $class$ [ $1$ ] ] [ $class$ [ $0$ ] ] ] [ $priors\_count \le 1.5$ [ $class$ [ $0$ ] ] [ $class$ [ $1$ ] ] ] ] [ $age \le 36.5$ , edge label={node[midway, above] {False}}[ $priors\_count \le 2.5$ [ $class$ [ $0$ ] ] [ $class$ [ $1$ ] ] ] [ $priors\_count \le 7.5$ [ $class$ [ $0$ ] ] [ $class$ [ $1$ ] ] ] ] ]
\end{forest}}
\caption{DL8.5+th+lb (depth limit 3, $\lambda\!=\!0.001$): training time=0.422 (sec), training accuracy=0.683, test accuracy=0.695, \#leaves=8 on fold 2.}
\end{subfigure}
\caption{GOSDT and DL8.5 trees on the COMPAS dataset with guessed thresholds and guessed lower bounds at depth limit 3. The reference model was trained using 20 max-depth 3 weak classifiers. $\lambda=0.001$.}
\label{fig:compas_tree_d3}
\end{figure}
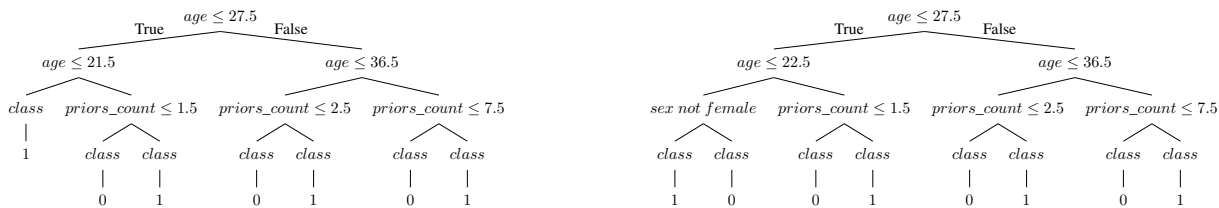

\begin{figure}
    \centering
    \begin{subfigure}[t]{0.98\textwidth}
    \centering
\scalebox{0.6}{\begin{forest}
[ $age \le 19.5$ [ $class$, edge label={node[midway, above] {True}} [ $1$ ] ] [ $priors\_count \le 2.5$, edge label={node[midway, above] {False}} [ $age \le 22.5$ [ $sex\ not \ female$ [ $class$ [ $1$ ] ] [ $class$ [ $0$ ] ] ] [ $age \le 27.5$ [ $priors\_count \le 1.5$ [ $class$ [ $0$ ] ] [ $class$ [ $1$ ] ] ] [ $class$ [ $0$ ] ] ] ] [ $age \le 36.5$ [ $class$ [ $1$ ] ] [ $priors\_count \le 7.5$ [ $class$ [ $0$ ] ] [ $class$ [ $1$ ] ] ] ] ] ]
\end{forest}}
\caption{GOSDT+th+lb (depth limit 5, $\lambda\!=\!0.001$): training time=34.804 (sec), training accuracy=0.685, test accuracy=0.696, \# leaves=9 on fold 2. }
    \end{subfigure}
    \vskip\baselineskip
    \begin{subfigure}[t]{0.98\textwidth}
    \centering
\scalebox{0.6}{\begin{forest}
[ $sex\ not\ female$ [ $age \le 22.5$, edge label={node[midway, above] {True}} [ $age \le 20.5$ [ $class$ [ $1$ ] ] [ $juvenile\_crimes \le 0.5$ [ $priors\_count \le 0.5$ [ $class$ [ $0$ ] ] [ $class$ [ $1$ ] ] ] [ $class$ [ $1$ ] ] ] ] [ $age \le 24.5$ [ $age \le 23.5$ [ $priors\_count \le 1.5$ [ $class$ [ $0$ ] ] [ $class$ [ $1$ ] ] ] [ $priors\_count \le 1.5$ [ $class$ [ $0$ ] ] [ $class$ [ $1$ ] ] ] ] [ $age \le 33.5$ [ $priors\_count \le 2.5$ [ $class$ [ $0$ ] ] [ $class$ [ $1$ ] ] ] [ $priors\_count \le 5.5$ [ $class$ [ $0$ ] ] [ $class$ [ $1$ ] ] ] ] ] ] [ $age \le 19.5$, edge label={node[midway, above] {False}} [ $class$ [ $1$ ] ] [ $age \le 20.5$ [ $class$ [ $1$ ] ] [ $age \le 38.5$ [ $priors\_count \le 2.5$ [ $class$ [ $0$ ] ] [ $class$ [ $1$ ] ] ] [ $priors\_count \le 7.5$ [ $class$ [ $0$ ] ] [ $class$ [ $1$ ] ] ] ] ] ] ]
\end{forest}}
    \caption{DL8.5+th+lb (depth limit 5, $\lambda\!=\!0.001$): training time=0.835 (sec), training accuracy=0.686, test accuracy=0.693, \# leaves=18 on fold 2. }
    \end{subfigure}
    \caption{GOSDT and DL8.5 trees on the COMPAS dataset with guessed thresholds and guessed lower bounds at depth limit 5. The reference model was trained using 20 max-depth 3 weak classifiers. $\lambda=0.001$.}
    \label{fig:compas_tree_d5}
\end{figure}
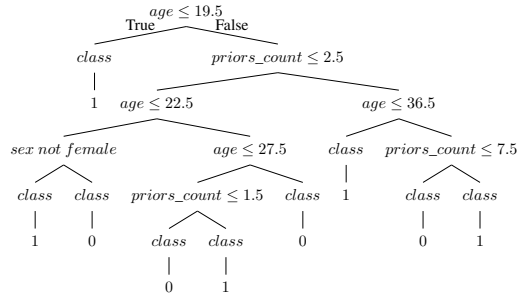
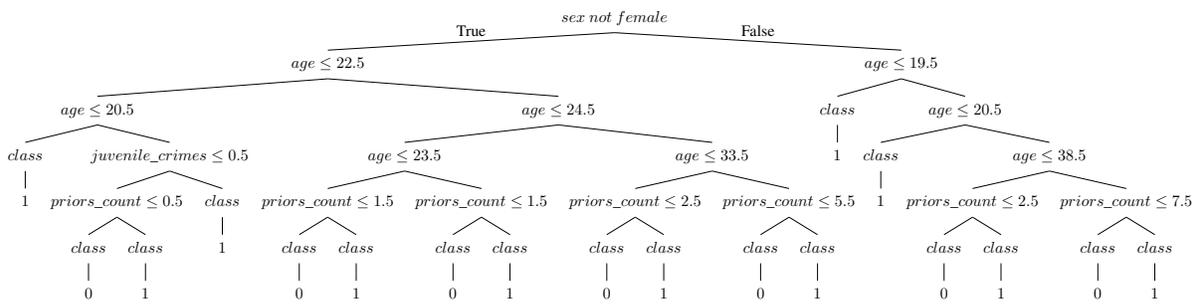

\section{When Guesses are Wrong}
\label{app:bad-guess}

At this point, we have demonstrated the benefit of guessing, but what happens when guesses are wrong? We demonstrate the impact of poor threshold and lower bound guesses by running an experiment using all datasets and the configurations listed in Appendix \ref{app:more-experiments}. On most datasets, we do not lose accuracy when using any of our four different reference model configurations. The spiral dataset provides one example where bad guesses affect accuracy. 

\noindent\textbf{Calculation}: We summarize the training accuracy, training time, and number of leaves for the models. Since we do not use 5-fold cross validation for the spiral dataset, we do not include the variability measure.

\begin{figure}[ht]
    \centering
    \begin{subfigure}{0.49\textwidth}
    \includegraphics[scale=0.23]{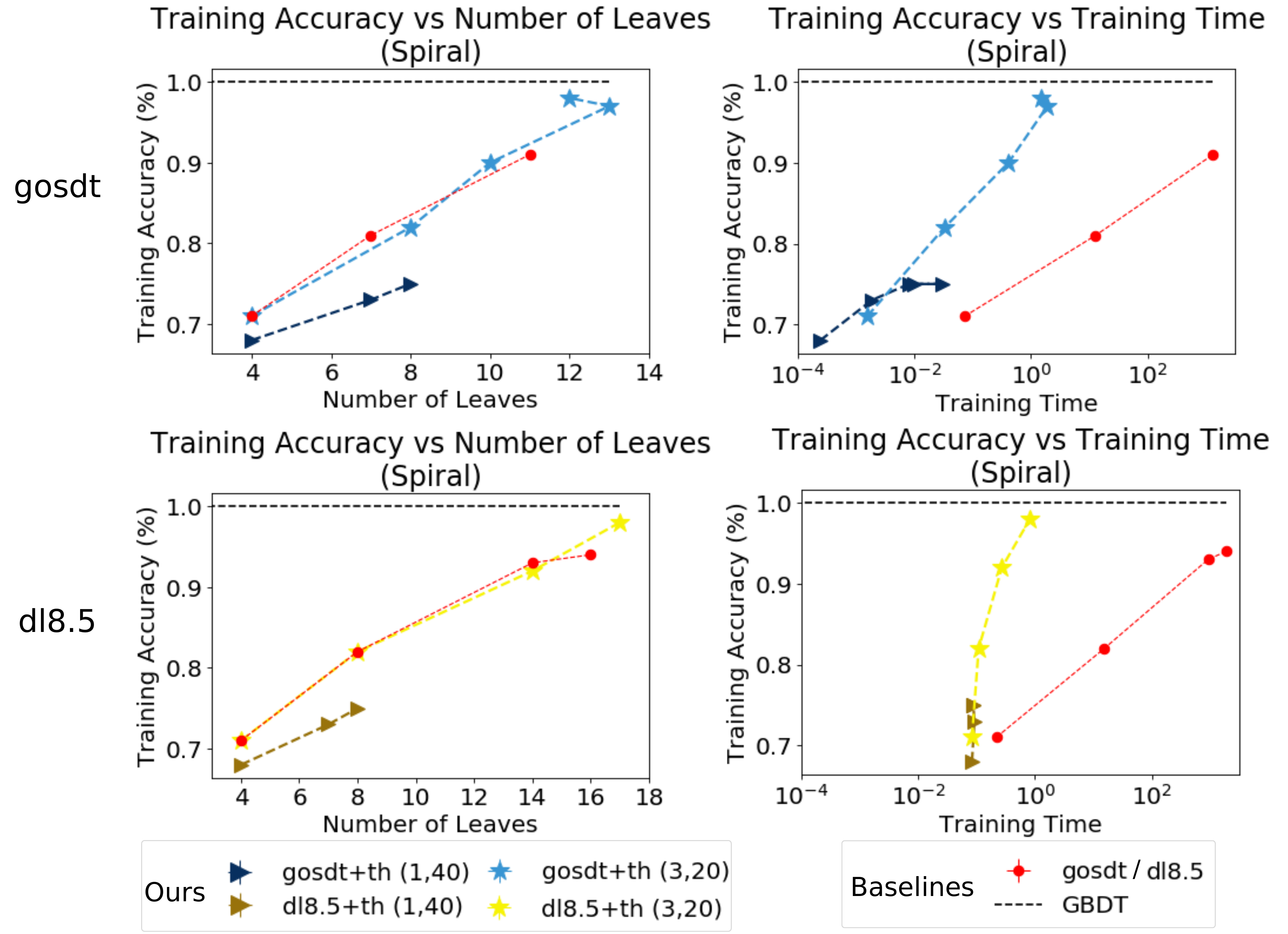}
    \caption{Examples of overly simple threshold guesses.}
    \label{fig:bad-thresh-guess-spiral}
    \end{subfigure}
    \begin{subfigure}{0.49\textwidth}
    \includegraphics[scale=0.23]{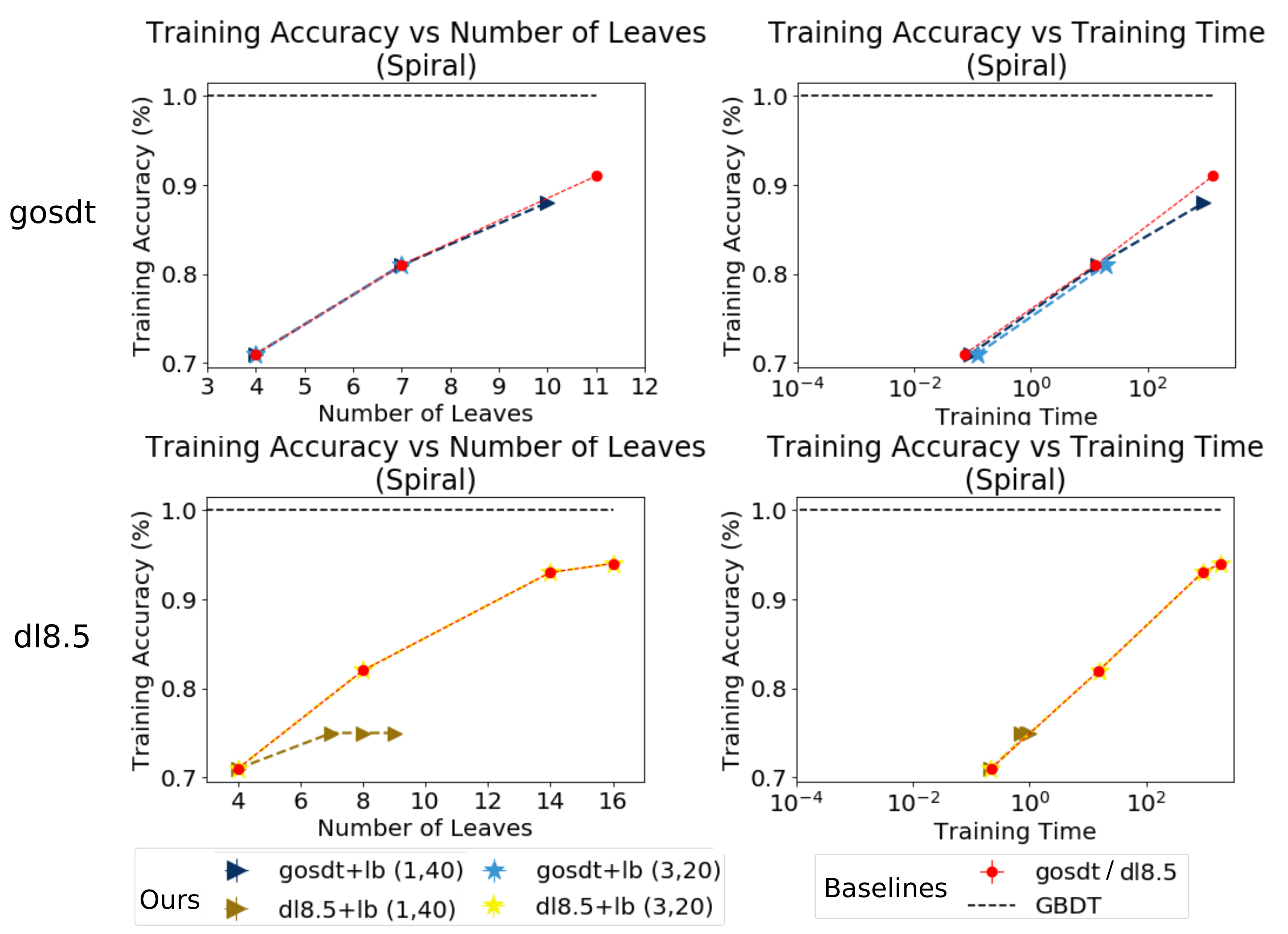}
    \caption{Examples of overly simple lower bound guesses.}
    \label{fig:bad-lb-guess-spiral}
    \end{subfigure}
    \caption{Performance of GOSDT and DL8.5 trained with threshold/lower bound guesses based on two different reference models and without threshold/lower bound guesses (in red dots) on the Spiral dataset. GOSDT and DL8.5 trained with GBDT of 20 max-depth 3 weak classifiers are comparable to the baselines. $\lambda=0.01$. Results of all depth limits are included.}
\end{figure}

\noindent\textbf{Results:} Figure \ref{fig:bad-thresh-guess-spiral} shows examples of guesses that prune too much of the search space. A variety of configurations 
lead to orders of magnitude reduction in training time.
However, GOSDT and DL8.5 trained with a reference model of 40 decision stumps 
can lead to worse sparsity/accuracy tradeoffs than GOSDT and DL8.5 without guessing. 
Similar patterns hold for lower bound guesses (see Figure \ref{fig:bad-lb-guess-spiral} bottom left).
But GOSDT and DL8.5 trained using a reference model with 20 max-depth 3 weak classifiers are comparable to the baselines.

This phenomenon is easily explained by noting that the reference model with 40 decision stumps has poor performance to begin with, e.g. 0.75 accuracy. The decision trees found with this reference model, although low in accuracy, remain as good as the performance of the reference model (in this case, even better!). 
Though we provide worst case bounds in terms of how far the decision tree will differ from the reference model, we do not guarantee that we find a strong tree unless the reference model is strong. Any practitioner using our approach should make sure that the reference model is accurate enough not to misguide the search. 

\section{Threshold Guess by Explainable Boosting Machine (EBM)}\label{app:ebm}
Section \ref{sec:thresh_guess} introduces how we guess thresholds using GBDT as a reference model. In this appendix, we propose another threshold guessing strategy using the explainable boosting machine (EBM)\footnote{Code is from https://github.com/interpretml/interpret}. EBM \citep{lou2012intelligible, caruana2015intelligible} is a generalized additive model (GAM) which is learned using a shallow bagged ensemble of trees on a single feature in each step of stochastic gradient boosting. This GAM model generally has accuracy comparable to the state-of-the-art black-box models like random forests and boosted trees while still maintaining high interpretability. Since EBM can visualize shape functions of each feature, we can easily see the contribution of a single feature to the prediction. Some visible jumps in these shape functions are very informative \citep{caruana2015intelligible}. We can use the thresholds where the shape function jumps as our guess: First, calculate the difference of scores between two neighboring thresholds for each feature. Second, for each feature $x_{\cdot,j}$, select thresholds where the difference is larger than $\textrm{tolerance} \times \frac{\max_j\textrm{variable importance}(x_{\cdot, j})}{\textrm{variable importance}(x_{\cdot, j})}$, where the variable importance is calculated as the average of the absolute predicted value of each feature for the training dataset. That is, if a feature is less important overall, the thresholds should have larger jumps in order to be selected. There is no restriction for tolerance, and in this experiment, we set it to be either 0.1 or 0.08. 

\noindent\textbf{Collection and Setup}: We ran GOSDT with a GBDT reference model trained using 40 decision stumps and GOSDT with a EBM reference model trained using tolerance 0.1 and 0.08 on the COMPAS and FICO datasets. We ran this particular experiment on a 2.7Ghz (768GB RAM 48 cores) Intel Xeon Gold 6226 processor. We set a 30-minute time limit and 200 GB memory limit. 

\noindent\textbf{Calculations}: We use the same calculations as those mentioned in Appendix \ref{app:acc-v-sparsity}.

\begin{figure}[ht]
    \centering
    \includegraphics[scale=0.35]{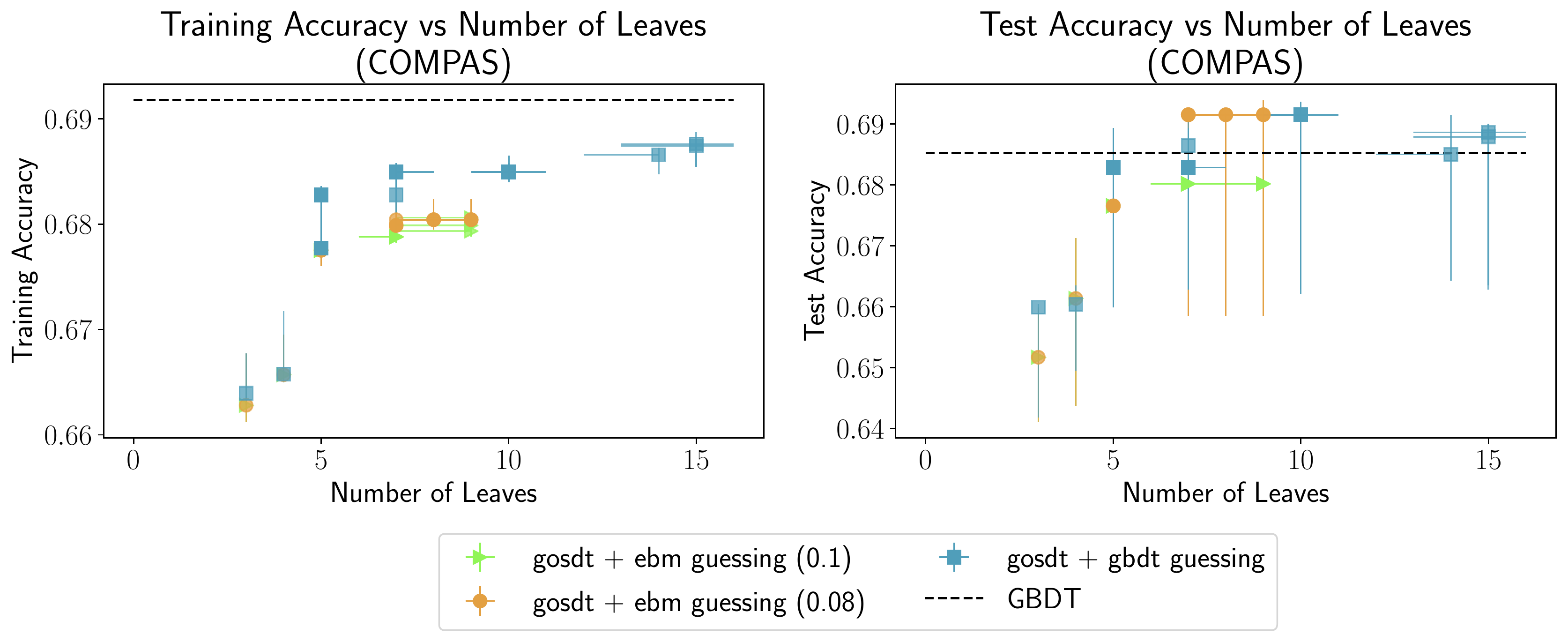}
    \includegraphics[scale=0.35]{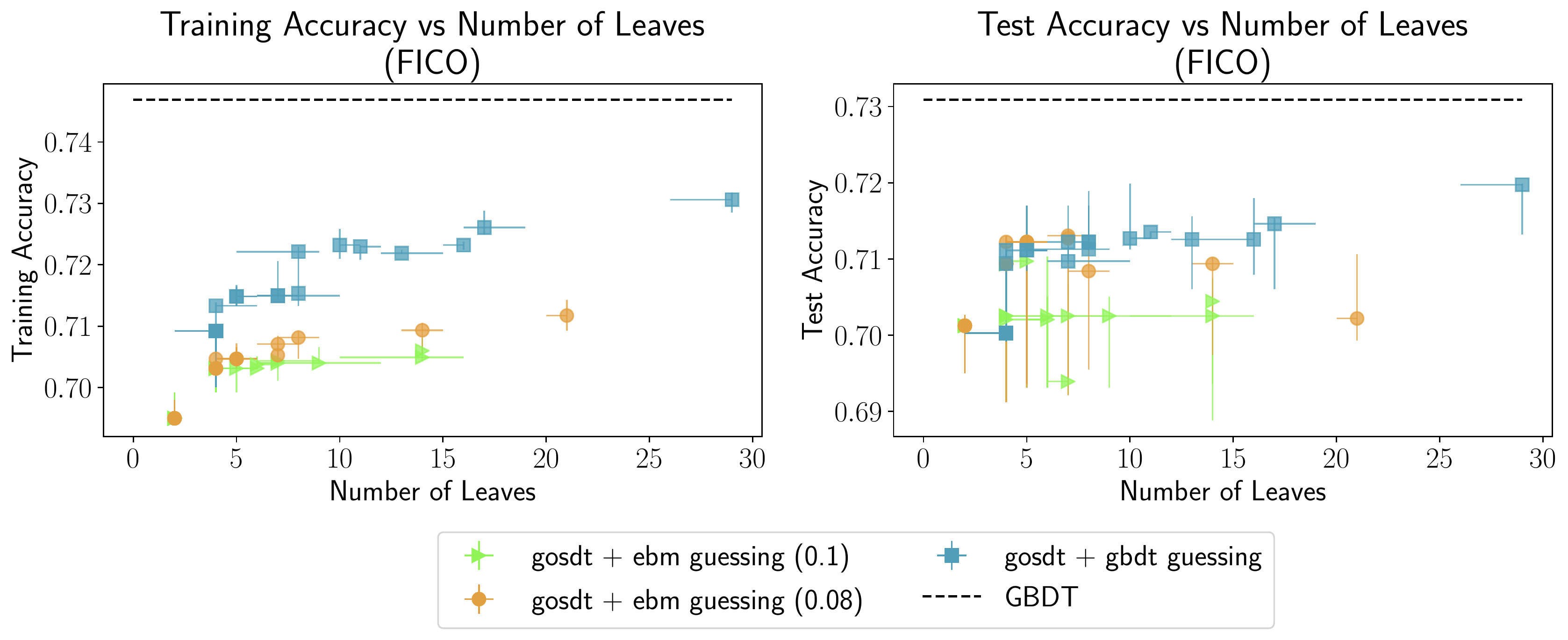}
    \caption{Sparsity versus accuracy for GOSDT with EBM threshold guessing and GBDT threshold guessing. Two different tolerances (0.1, 0.08) were used for EBM threshold guesses. 40 decision stumps were used to train GBDT reference model. The black dash line is a GBDT with 100 max-depth 3 weak classifiers. }
    \label{fig:ebm_guess}
\end{figure}

\noindent\textbf{Results}: Figure \ref{fig:ebm_guess} shows the sparsity accuracy trade-offs for GOSDT with EBM threshold guessing and GBDT threshold guessing. Both guessing strategies can help GOSDT achieve accuracy comparable to GBDT with 100 max-depth 3 weak classifiers (black dash line). Two subfigures on the left show that GBDT-based threshold guessing is better than EBM-based threshold guessing when considering training accuracy. 

\section{FICO Case Study}\label{app:fico}
A decision tree is a special type of decision set. In this appendix, we transform some trees we trained using GOSDT on the FICO dataset to decision sets and compare these transformed decision sets with boolean rules learned by \citet{dash2018boolean}\footnote{Code is from https://github.com/Trusted-AI/AIX360}. For simplicity, we call our algorithm \textit{CGrules}. 

\noindent\textbf{Collection and Setup}: Details about training GOSDT with all guessing strategies were introduced in Appendix \ref{app:more-experiments}. To train CGrules, we used two different configurations: 1) All parameters were chosen to be their default values, including the way of binarizing features. We denote CGrules with this configuration as \textit{CGrules-default}. 2) All parameters were chosen to be in the default setting except two model complexity parameters which penalize the number of clauses in the rule and the number of conditions in each clause. We set these two parameters to 1e-7. We denote CGrules with this configuration as \textit{CGrules-cp}. We ran this particular experiment on a 2.7Ghz (768GB RAM 48 cores) Intel Xeon Gold 6226 processor. 

\noindent\textbf{Results}: 
Table \ref{tab:compare_cgrules} shows decision rules learned by CGrules and GOSDT with all three guessing strategies. The model from CGrules trained using the default setting is very sparse, consisting of only one clause with two conditions. Compared to the rules trained by CGrules-cp, a depth-2 GOSDT tree is simpler but achieves higher accuracy. A depth-5 GOSDT tree, though more complicated than the others, has the highest accuracy.

\begin{table}[]\scriptsize
    \centering
    \begin{tabular}{|m{7em}|m{42em}|m{5em}|m{5em}|}\hline
    Algorithm & Decision Rules & Train Acc & Test Acc\\\hline
    CGrules-default & $\begin{aligned}
    & \textrm{if ExternalRiskEstimate} > 71 \textrm{ and AverageMInFile} > 47, \textrm{then predict good}\\
    & \textrm{else predict bad}
    \end{aligned}$ & 0.701 & 0.709\\ \hline
    CGrules-cp & $\begin{aligned}
    & \textrm{if ExternalRiskEstimate} > 74 \textrm{ and NetFractionRevolvingBurden} \leq 47\textrm{, then predict good}\\
    &\textrm{or if ExternalRiskEstimate} > 65 \textrm{ and AverageMInFile} > 27 \textrm{ and NumSatisfactoryTrades} > 5 \textrm{ and}\\
    &\quad \textrm{MSinceMostRecentInqexcl7days} \leq -8 \textrm{ and NetFractionRevolvingBurden} \leq 77\textrm{, then predict good}\\
    &\textrm{else predict bad}
    \end{aligned}$ & 0.704 & 0.704\\\hline
    GOSDT+th+lb (depth\_limit=2, $\lambda=0.001$) & $\begin{aligned}
    & \textrm{if AverageMInFile} \le 59.5 \textrm{ and ExternalRiskEstimate} >78.5, \textrm{then predict good}\\
    & \textrm{or if AverageMInFile} > 59.5 \textrm{ and ExternalRiskEstimate} > 70.5, 
     \textrm{then predict good}\\
    & \textrm{else predict bad}
    \end{aligned}$ & 0.708 & 0.707\\\hline
    GOSDT+th+lb (depth\_limit=5, $\lambda=0.001$) & $\begin{aligned}
    & \textrm{if ExternalRiskEstimate} > 78.5, \textrm{then predict good}\\
    & \textrm{or if ExternalRiskEstimate} > 70.5 \textrm{ and ExternalRiskEstimate} \leq 78.5 \textrm{ and} \\
    &\quad \textrm{MSinceMostRecentInqexcl7days}> 0.5, \textrm{then predict good}\\
     & \textrm{or if ExternalRiskEstimate} > 70.5 \textrm{ and ExternalRiskEstimate} \leq 78.5 \textrm{ and} \\
    &\quad \textrm{MSinceMostRecentInqexcl7days}\leq -7.5 \textrm{ and PercentTradesWBalance} > 73.5, \textrm{then predict good}\\
    & \textrm{or if ExternalRiskEstimate} > 70.5 \textrm{ and ExternalRiskEstimate} \leq 78.5 \textrm{ and} \\
    &\quad \textrm{MSinceMostRecentInqexcl7days}\leq 0.5 \textrm{ and PercentTradesWBalance} \leq 73.5 \textrm{ and}\\
    &\quad\textrm{AverageMInFile}>63.5, \textrm{then predict good}\\
    & \textrm{else predict bad}
    \end{aligned}$ & 0.717 & 0.719\\\hline
    \end{tabular}
    \caption{Comparison of CGrules and GOSDT trees with all three guessing strategies. All decision rules are shown for the $5^{th}$ fold. The reference model for GOSDT was trained using 40 decision stumps.}
    \label{tab:compare_cgrules}
\end{table}

\section{Formal Description of Modifications to Branch-and-Bound Algorithms to Use Lower Bound Guessing With a Reference Model}\label{app:lb-algs}
Below is the full description of the DL8.5 algorithm, exactly as provided in \cite{aglin2020learning}, except for the modifications shown in \textcolor{red}{red} to support lower bound guessing. We use the same notation as in the original DL8.5 paper, additionally defining \textcolor{red}{$T$} as the reference model for lower bound guessing. The resulting algorithm supports branch and bound decision tree optimization with lower bound guessing. 
We refer to \cite{aglin2020learning} for detailed explanation of the notation used in this algorithm. 

\definecolor{comment}{rgb}{0.2,0.4,0.2}
\def\comment#1{\textcolor{comment}{\textit{ // #1 }}}
\begin{algorithm}[H]
\caption{DL8.5$(\textit{maxdepth}, \textit{minsup})$} \label{alg:dl85}
\begin{tabbing}
xxx \= xx \= xx \= xx \= xx \= xx \kill
1: \> \textbf{struct} \textit{{\BestTree \{\initub: float; tree: Tree; error: float\}}} \\
2: \> $\textit{cache} \gets \textit{HashSet}< \textit{Itemset, BestTree} >$ \\
3: \> $\textit{bestSolution} \gets \textrm{DL8.5 \textendash{} Recurse}(\emptyset, +\infty)$ \\
4: \> \textbf{return} \textit{bestSolution.tree} \\
5: \> \textbf{Procedure} $\textrm{DL8.5 \textendash{} Recurse}(\textit{$I$, init\_ub})$ \\
6: \> \> \textbf{if} $\textit{leaf\_error}($I$) = 0$ \textit{or} $|I| = \textit{maxdepth}$ \textit{or time-out is reached} \textbf{then} \\
7: \> \> \> \textbf{return} \textit{BestTree(init\_ub, make\_leaf($I$), leaf\_error($I$))}\\
8: \> \> $\textit{solution} \gets \textit{cache.get($I$)}$\\
9: \> \> \textbf{if} \textit{solution was found and} $((\textit{solution.tree} \neq \textrm{NO\_TREE}) \textit{ or } (\initub \leq \textit{solution}.\initub))$ \textbf{then}\\
10: \> \> \> \textbf{return} \textit{solution} \\
11: \> \> $(\tau, b, ub) \gets (\textrm{NO\_TREE}, +\infty, \initub)$ \\
12: \> \> \textbf{for} \textit{all attributes i in a well-chosen order} \textbf{do} \\
13: \> \> \> \textbf{if} $|\textit{cover}(I \cup \{ i \} )| \geq \textit{minsup}$ \textit{and} $|\textit{cover}(I \cup \{ \neg i \})| \geq \textit{minsup}$ \textbf{then}\\\hspace*{200pt}\quad\quad\quad\textrm{(support is sufficient along both branches)}\\
14: \> \> \> \> $\textit{sol}_1 \gets \textrm{DL8.5 \textendash{} Recurse}(I \cup \{ \neg i \}, ub)$ \quad\textrm{(recurse on the left branch)}\\
15: \> \> \> \textbf{if} $\textit{sol}_1.\textit{tree} = \textrm{NO\_TREE}$ \textbf{then}\quad\quad\quad\textrm{(this is a leaf node)}\\
16: \> \> \> \> \textbf{continue}\\
17: \> \> \> \textbf{if} $\textit{sol}_1.\textit{error} \leq ub$ \textbf{then} \\
18: \> \> \> \> $\textit{sol}_2 \gets \textrm{DL8.5 - Recurse}(I \cup \{ i \}, ub - \textit{sol}_1.\textit{error})$ \quad\textrm{(recurse on the right branch)} \\
19: \> \> \> \> \textbf{if} $\textit{sol}_2.\textit{tree} = \textrm{NO\_TREE}$ \textbf{then}\quad\quad\quad\textrm{(this is a leaf node)}\\
20: \> \> \> \> \> \textbf{continue} \\
21: \> \> \> \> $\textit{feature}\_\textit{error} \gets \textit{sol}_1.\textit{error} + \textit{sol}_2.\textit{error}$ \\
22: \> \> \> \> \textbf{if} $\textit{feature\_error} \leq ub$ \textbf{then} \quad\quad\quad\textrm{(we have a new current best)}\\
23: \> \> \> \> \> $\tau \gets \textit{make\_tree}(i, \textit{sol}_1.\textit{tree}, \textit{sol}_2.\textit{tree})$ \\
24: \> \> \> \> \> $b \gets \textit{feature}\_\textit{error}$ \\
25: \> \> \> \> \> $ub \gets b - 1$ \quad\quad\textrm{(we must get at least one more point correct in order to change the bound later)}\\
 \> \> \> \> \textcolor{red}{(make DL8.5 notation consistent with our notation) let $s_I$ be the indices of the dataset captured by itemset $I$,} \\
 \> \> \> \> \textcolor{red}{that is, $s_I := cover(I)$, and then let $lb_\textrm{guess}(s_I) = \frac{1}{N}\sum_{j\in s_I}\mathds{1}[y_j \neq \hat{y}_j^T]$} \\
26: \> \> \> \> \redsout{\textbf{if} $\textit{feature\_error} = 0$ \textbf{then} }\\ 
 \> \> \> \> \textbf{if} $\textit{feature\_error} \textcolor{red}{\leq lb_\textrm{guess}(s_I)}$ \textbf{then} \\
27: \> \> \> \> \> \textbf{break} \\
28: \> $\textit{solution} \gets \BestTree(\initub, \tau, b)$ \\
29: \> $\textit{cache.store($I$, solution)}$ \\
30: \> \textbf{return} \textit{solution}  \\
\end{tabbing}
\end{algorithm}
For DL8.5, the modifications we actually need to make are are somewhat simpler than might be expected from the description in Section \ref{sec:lbguess}, since DL8.5 does not use a per-leaf penalty $\lambda = 0$, and normally sets the lower bound for each subproblem to 0 (as in line 26, which only decides that splitting on a certain feature is better than splitting on any other possible feature when the error is 0). We adjust to calculate a lower bound using our guessing approach (with $\lambda = 0$), and choose for our split the first feature that does as well or better than the guessed lower bound.

Below are several functions from the GOSDT algorithm, exactly as provided in \cite{lin2020generalized}, except for the modifications shown in red to support lower bound guessing. We use the same notation as in the original GOSDT paper, additionally defining \textcolor{red}{$T$} as the reference model for lower bound guessing. The resulting algorithm supports branch and bound decision tree optimization with lower bound guessing. 
We refer to \citet{lin2020generalized} for detailed explanation of the notation used in this algorithm, and for the full description of the algorithm (we do not include functions that do not need modifications to support lower-bound guessing). 
\setcounter{algorithm}{0}
\begin{algorithm}[htb!]
\caption{GOSDT$(R, \x, \y, \lambda, \textcolor{red}{T})$ \label{alg:gosdt_summary_app}}
\begin{tabbing}
xxx \= xx \= xx \= xx \= xx \= xx \kill
1: \> \textbf{input:} $R$, $Z$, $z^-$, $z^+$, $\lambda$,  \textcolor{red}{$T$} \comment{risk, unique sample set, negative sample set, positive sample set, regularizer, \textcolor{red}{reference model}} \\
2: \> $Q = \emptyset$ \comment{priority queue}\\
3: \> $G=\emptyset$ \comment{dependency graph}\\
4: \> $s_0 \leftarrow \{1,...,1\}$\comment{bit-vector of 1's of length $U$} \\
5: \> $p_0 \leftarrow$ FIND\_OR\_CREATE\_NODE($G, s_0$)\comment{node for root}\\
6: \> $Q.{\rm push}(s_0)$\comment{add to priority queue}\\
7: \> \textbf{while} $p_0.lb \neq p_0.ub$ \textbf{do}\\
8: \> \> $s\leftarrow Q.{\rm pop}()$\comment{index of problem to work on}\\
9: \> \> $p\leftarrow G.{\rm find}(s)$\comment{find problem to work on}\\
10: \> \> \textbf{if} $p.lb=p.ub$ \textbf{then}\\
11: \> \> \> \textbf{continue}\comment{problem already solved} \\
12: \> \> $(lb', ub') \leftarrow (\infty, \infty)$\comment{very loose starting bounds}\\
13: \> \> \textbf{for} each feature $j \in [1,M]$ \textbf{do}\\
14: \> \> \> $s_l, s_r \leftarrow \text{split}(s,j,Z)$ \comment{create children if they don't exist} \\
15: \> \> \> $p_l^j\leftarrow$FIND\_OR\_CREATE\_NODE$(G,s_l)$\\
16: \> \> \> $p_r^j\leftarrow$FIND\_OR\_CREATE\_NODE$(G,s_r)$\\
\>\>\comment{create bounds as if $j$ were chosen for splitting}
\\
17: \> \> \> $lb' \leftarrow \min(lb', p_l^j.lb + p_r^j.lb)$ \\
18: \> \> \> $ub' \leftarrow \min(ub', p_l^j.ub + p_r^j.ub)$ \\
\> \> \comment{signal the parents if an update occurred} \\
19: \> \> \textbf{if} $p.lb \neq lb'$ \textbf{or} $p.ub \neq ub'$  \textbf{then} \\
20: \> \> \> $(p.lb, p.ub) \leftarrow (lb', ub')$ \\
21: \> \> \> \textbf{for} $p_{\pi} \in G.{\rm parent}(p)$ \textbf{do} \\
\> \> \> \> \comment{propagate information upwards}\\
22: \> \> \> \> $Q.{\rm push}(p_{\pi}.{\rm id}, {\rm priority}=1)$\\
23: \> \> \redsout{\textbf{if} $p.lb = p.ub$ \textbf{then}} \\
 \> \> \textbf{if} $p.lb \textcolor{red}{\geq} p.ub$ \textbf{then} \\
23.5: \> \> \> \textcolor{red}{$p.lb \gets p.ub$} \\
24: \> \> \> \textbf{continue} \comment{problem solved just now} \\
\> \> \comment{loop, enqueue all children that are dependencies} \\

25: \> \> \textbf{for} each feature $j \in [1,M]$ \textbf{do} \\
\> \> \comment{fetch $p_l^j$ and $p_r^j$ in case of update from other thread} \\
26: \> \> \> repeat line 14-16\\
27: \> \> \> $lb' \leftarrow p_l^j.lb + p_r^j.lb$ \\
28: \> \> \> $ub' \leftarrow p_l^j.ub + p_r^j.ub$ \\
29: \> \> \> \textbf{if} $lb' < ub'$ \textbf{and} $lb' \le p.ub$ \textbf{then} \\
30: \> \> \> \> $Q.{\rm push}(s_{l}, {\rm priority}=0)$ \\
31: \> \> \> \> $Q.{\rm push}(s_{r}, {\rm priority}=0)$ \\
32: \> \textbf{return}\\
---------------------------------------------------------------------------\\
33: \> \textbf{subroutine} FIND\_OR\_CREATE\_NODE(G,s)\\
34: \> \> if $G.{\rm find}(s) = {\rm NULL}$ \comment{$p$ not yet in dependency graph}\\
35: \> \> \> $p.id \leftarrow s$ \comment{identify $p$ by $s$}\\
36: \> \> \> $p.lb \leftarrow {\rm get\_lower\_bound}(s,Z,z^-,z^+)$\\
37: \> \> \> $p.ub \leftarrow {\rm get\_upper\_bound}(s,Z,z^-,z^+)$\\
38: \> \> \> \textbf{if} fails\_bounds$(p)$ \textbf{then}\\
39: \> \> \> \> $p.lb=p.ub$ \comment{no more splitting allowed}\\
40: \> \> \> G.insert(p) \comment{put $p$ in dependency graph}\\
41: \> \> \textbf{return} G.find(s)
\end{tabbing}
\end{algorithm}
\begin{algorithm}
\caption{get\_lower\_bound$(s, Z, z^-, z^+)$ $\rightarrow$ $lb$\label{alg:lowerbound})}
\begin{minipage}{1.0\linewidth}
\begin{tabbing}
xxx \= xxx \= xxx \kill
\textbf{input:} $s, Z, z^-, z^+$ \comment{support, unique sample set, negative sample set, positive sample set} \\
\textbf{output:} $lb$ \comment{Risk lower bound} \\
\redsout{\comment{Compute the risk contributed if applying a class to every equivalance class independently} }\\
\redsout{\textbf{for} each equivalence class $u \in [1,U]$ \textbf{define}}\\
\> \redsout{\comment{Values provided in $Z$}}\\
\> \redsout{$z_u^- = \frac{1}{N}\sum_{i=1}^{N}\mathds{1}[y_i = 0 \land x_i = z_u]$} \\
\> \redsout{$z_u^+ = \frac{1}{N}\sum_{i=1}^{N}\mathds{1}[y_i = 1 \land x_i = z_u]$} \\
\> \redsout{\comment{Risk of assigning a class to equivalence class $u$}} \\
\> \redsout{$z_u^{\min} = \min(z_u^-, z_u^+)$} \\
\redsout{\comment{Add all risks for each class $u$}}\\
\comment{\textcolor{red}{Compute the risk contributed if we make exactly the same errors as the reference model T}} \\
\comment{Add a single $\lambda$ which is a lower bound of the complexity penalty}\\
\redsout{$lb \leftarrow \lambda + \sum_u s_u z_u^{\min}$}\\
$lb \leftarrow \lambda + \textcolor{red}{\frac{1}{N}\sum_{i \in s}\mathds{1}[y_i \neq \hat{y}_i^T]}$\\
\textbf{return} $lb$
\end{tabbing}
\end{minipage}
\end{algorithm}

The modifications to support lower bound guessing are quite simple for GOSDT as well -- we simply replace the previous lower bound calculation with our guessed calculation (and account for the potential of the lower bound exceeding the actual solution by changing an = sign to a $\geq$ sign). The complexity of the modifications listed in Section \ref{alg:lbg} result from accounting for the effects this relatively small code change can have on the algorithm's behaviour. Because branch and bound approaches to decision tree optimization can work on subproblems in a variety of different orders and can cache intermediate results, changing the lower bound calculation requires us to account for the effects of these adjusted lower bounds when searching through subproblems in almost any order.

The coded implementation of GOSDT with lower bound guessing we provide differs slightly from this algorithm, because the public code base for GOSDT continued to evolve after the submission of the GOSDT paper, and now differs slightly from the algorithm described herein. We refer to the codebase itself for details on these differences, most of which are minor. We do highlight one significant difference -- GOSDT's new codebase features a notion of ``scope'' for each subproblem, which prevents the algorithm from exploring one half of a split if the other half has too high a lower bound. We originally allowed scope to be used alongside lower-bound guessing (doing so is consistent with how we describe lower bound guessing, and does not change the validity of our proofs on bounding the maximal inaccuracy). However, we found that when we allowed scope to be used alongside lower-bound guessing, it led to difficulty with verifying the correctness of our implementation. The right child of a split might be excluded from the search space because the left child of that split had a high guessed lower bound, but if we later found that the left child's lower bound guess was too high and decreased the lower bound, GOSDT would not re-enqueue the right child. That is not a violation of our lower bound approach (we simply used guessed lower bounds to prune the search space, and it can be shown that Theorem \ref{thm:glb} still holds with this strategy), but GOSDT's normal error-checking functions would occasionally report that GOSDT failed to converge when this happened. Rather than suppressing these errors, we chose not to base scope off of the guessed lower bound, but rather to base it off of the equivalent points bound (see Theorem B.9 in \citealt{lin2020generalized}), which is provably never an overestimate of the lower bound.  


\section{GOSDT with all guessing strategies versus optimal classification trees} \label{app:oct}
In this appendix, we compare GOSDT with all three guessing strategies to the product \textit{optimal decision trees} provided by interpretable AI with an academic license. We do not know how this algorithm works, and it is not publicly available without a license. The website suggests that the technique is related to OCT \citep{bertsimas2017optimal}, thus we label it OCT in our plots; however, the OCT algorithm of \citet{bertsimas2017optimal} is extremely slow, so we do not know whether a similar approach is used in interpretable.ai's software.

\noindent\textbf{Collection and Setup:} We ran GOSDT with a GBDT reference model trained using 40 decision stumps and OCT on the COMPAS and FICO datasets. The corresponding depth limit and regularizers for these two datasets are in Table \ref{tab:tree_config}.  We ran this particular experiment on a 2.7Ghz (768GB RAM 48 cores) Intel Xeon Gold 6226 processor. We set a 30-minute time limit and 200 GB memory limit. We set complexity parameters $cp$ and depth in interpretable.ai's software to the same values of regularizers and depth limits in Table \ref{tab:tree_config} and left all other parameters in default.

\noindent\textbf{Calculations:} We use the same calculations as those mentioned in Appendix \ref{app:acc-v-sparsity}. 

\begin{figure}
    \centering
    \includegraphics[scale=0.35]{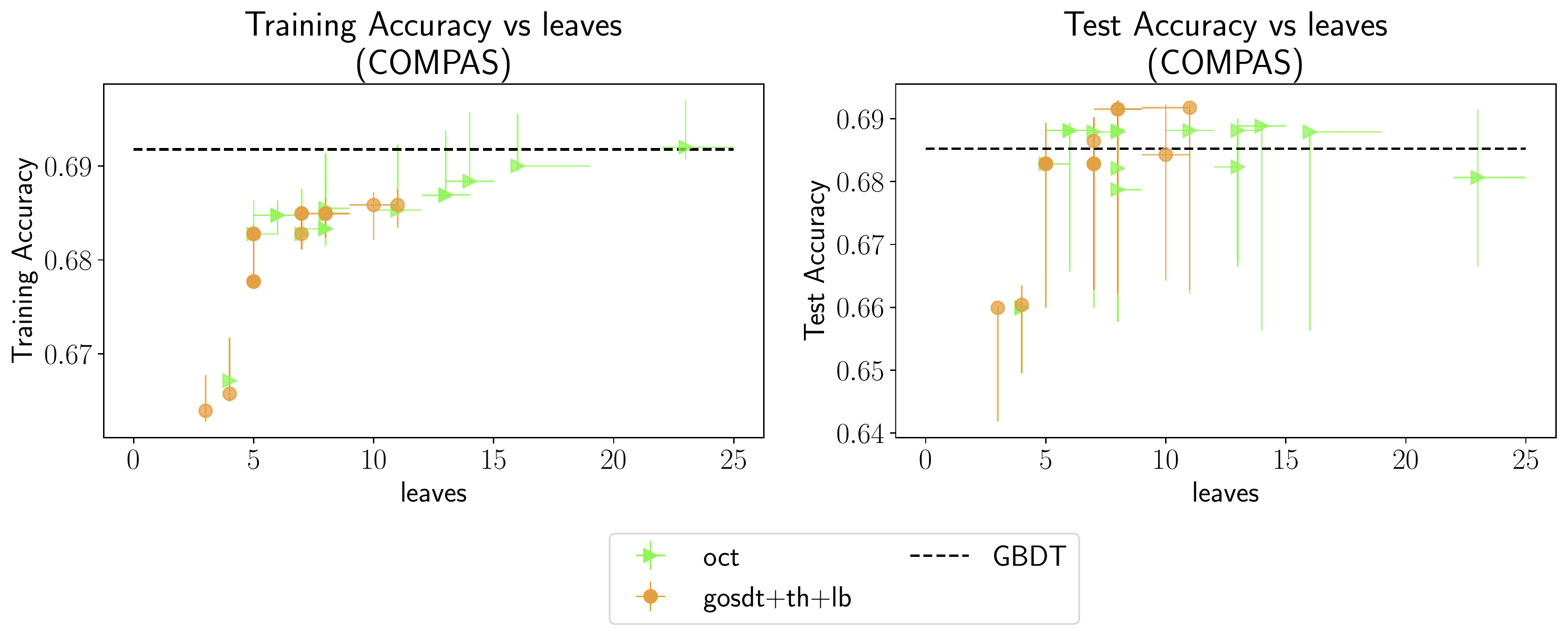}
    \includegraphics[scale=0.35]{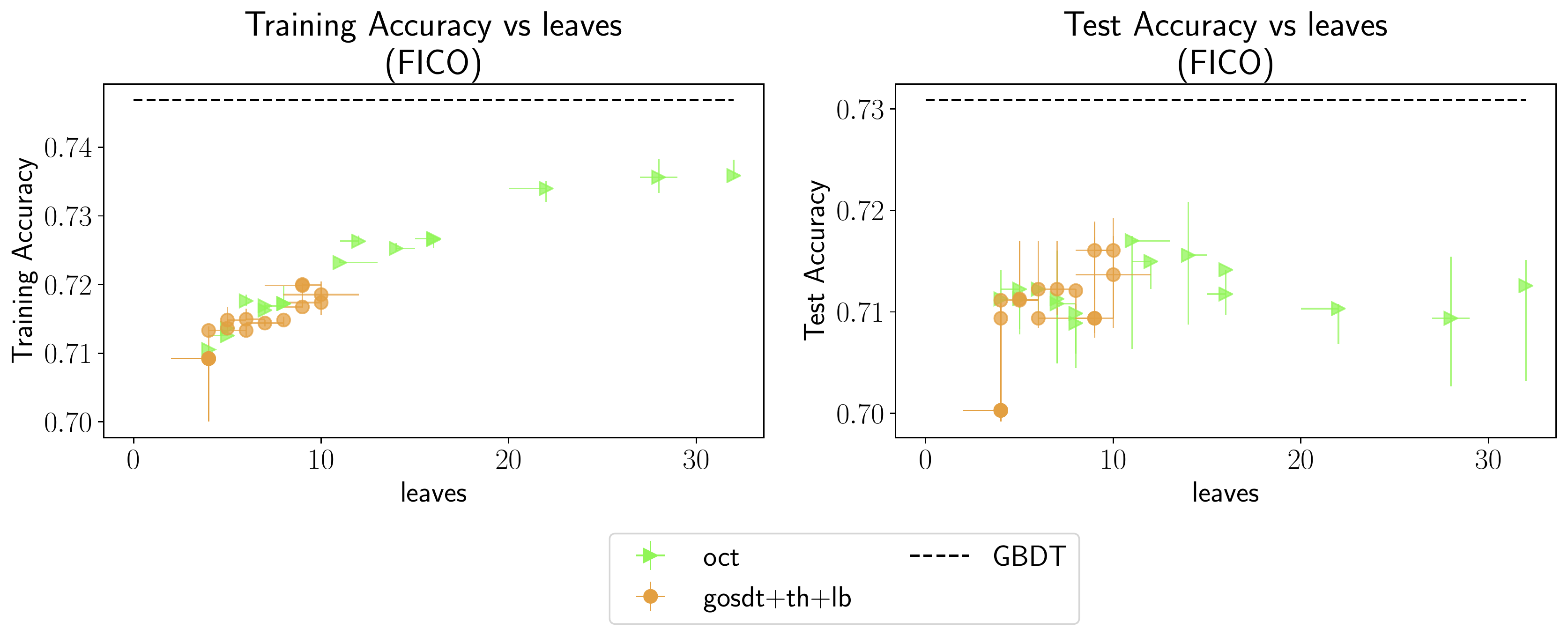}
    \caption{Sparsity versus accuracy for GOSDT with all three guessings and OCT. The black dash line is a GBDT with 100 max-depth 3 weak classifiers. }
    \label{fig:vs_oct_sparsity}
\end{figure}

\begin{figure}
    \centering
    \includegraphics[scale=0.35]{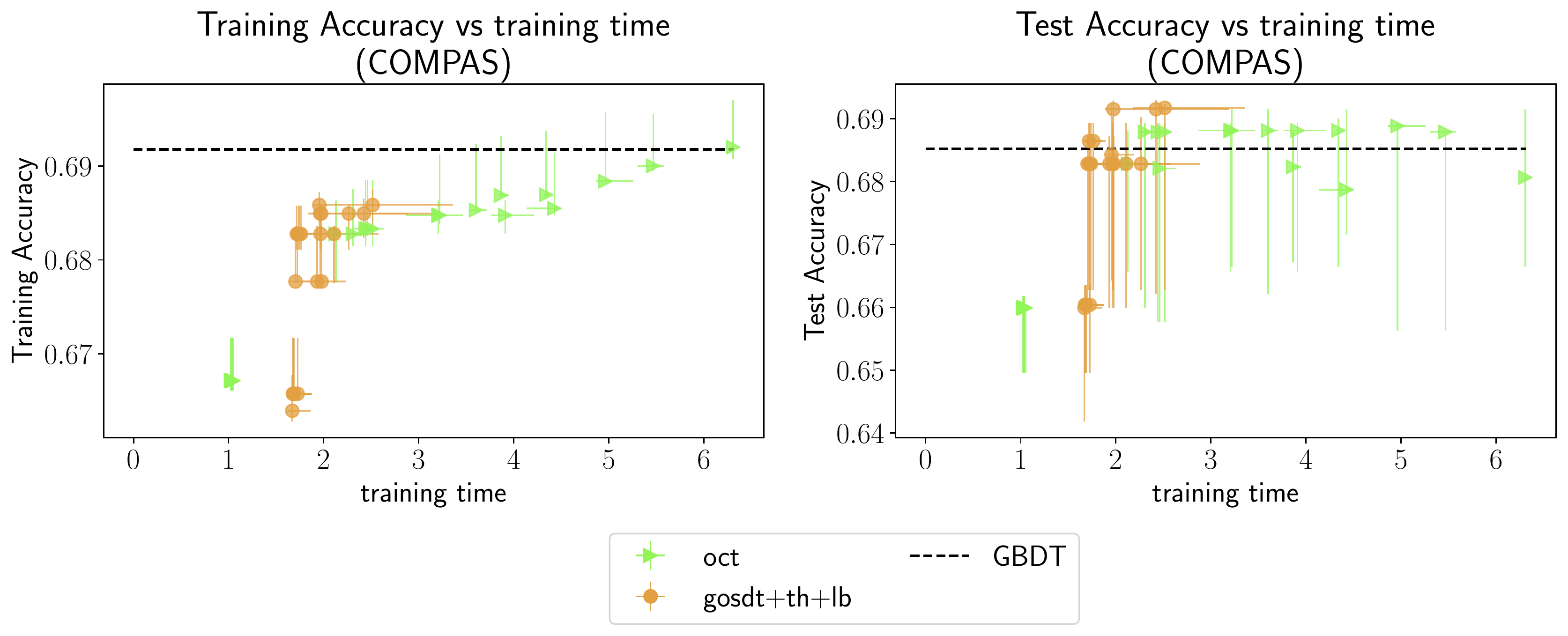}
    \includegraphics[scale=0.35]{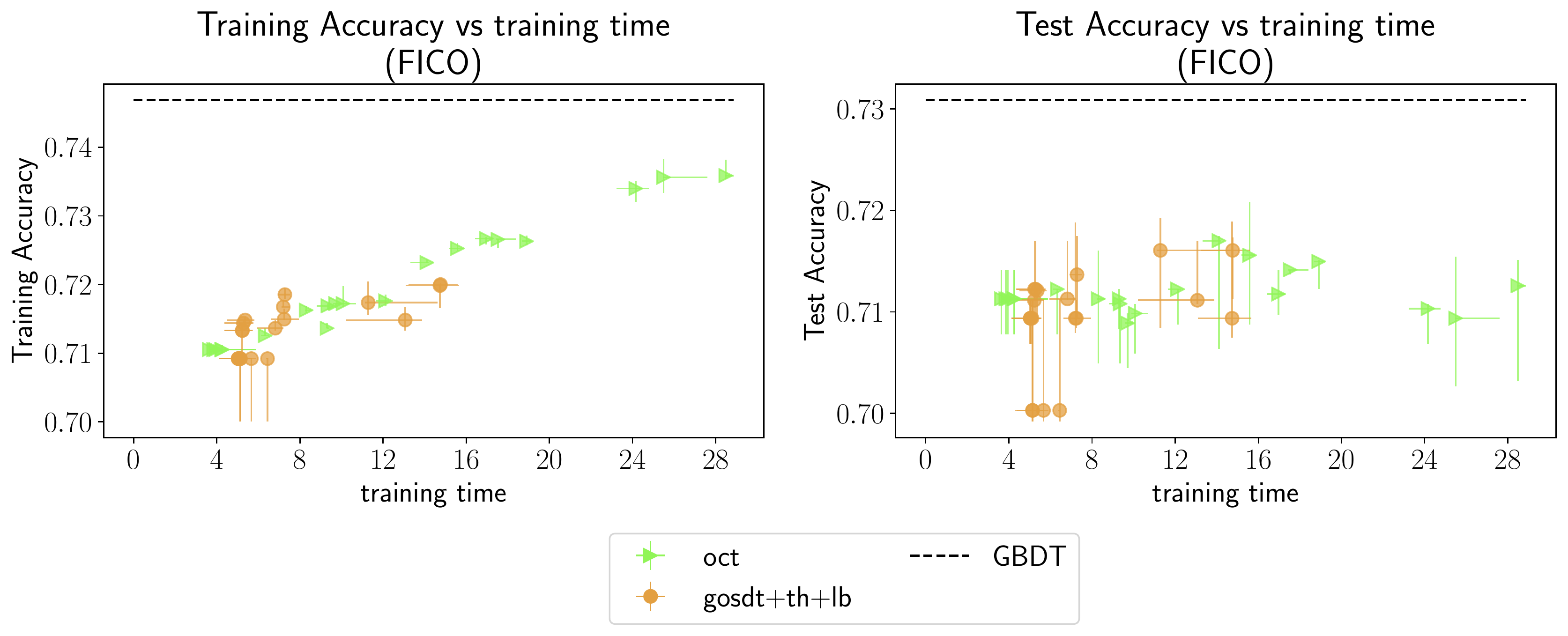}
    \caption{Time versus accuracy for GOSDT with all three guessings and OCT. The black dash line is a GBDT with 100 max-depth 3 weak classifiers. }
    \label{fig:vs_oct_time}
\end{figure}

\noindent\textbf{Results:} Figure \ref{fig:vs_oct_sparsity} shows the sparsity and accuracy trade-offs for GOSDT+guessing and OCT. The test performance on both datasets is similar, though it appears that the larger OCT models tended to overfit (training accuracy improved while test accuracy stayed the same), whereas the GOSDT+guessing models did not have this problem. This is possibly because GOSDT+guessing used a reference model that did not overfit, so when GOSDT+guessing matched its performance, it also did not overfit. The major difference between the performance of the algorithms was that \textit{on both datasets, GOSDT+guessing was faster to converge}. 
Figure \ref{fig:vs_oct_time} shows the time in seconds versus accuracy for GOSDT and OCT. The brown dots to the left illustrate that GOSDT was generally faster than OCT to get to the same level of accuracy.

Thus, GOSDT+guessing yields several advantages over OCT. (1) It is publicly available. (2) It yields faster solutions. (3) Its solutions come with guarantees on performance, whereas OCT (as far as we can tell) does not.

\end{document}